\documentclass[journal]{IEEEtran}
\DeclareUnicodeCharacter{2113}{\ensuremath{\ell}}
\DeclareUnicodeCharacter{2032}{\ensuremath{'}}
\DeclareUnicodeCharacter{0097}{\textemdash}
\usepackage[utf8]{inputenc}

\usepackage{amsmath,amsfonts}
\usepackage{algorithmic}
\usepackage{algorithm}
\usepackage{array}
\usepackage[caption=false,font=normalsize,labelfont=sf,textfont=sf]{subfig}
\usepackage{textcomp}
\usepackage{stfloats}
\usepackage{url}
\usepackage{verbatim}
\usepackage{graphicx}
\usepackage{cite}
\usepackage{color}
\usepackage{multirow}
\usepackage{booktabs}

\usepackage{tikz}
\usetikzlibrary{arrows.meta}

\begin{document}

\title{Graph Attention-Driven Bayesian Deep Unrolling\\for Dual-Peak Single-Photon Lidar Imaging}

\author{Kyungmin~Choi,
        JaKeoung~Koo,
        Stephen~McLaughlin,~\IEEEmembership{Fellow,~IEEE}, % <-this % stops a space
        Abderrahim~Halimi,~\IEEEmembership{Senior Member,~IEEE}
\thanks{This work was supported by the National Research Foundation of Korea (NRF) grant, which was funded by the Korea government (MSIT) (No. RS-2023-00240740) and was supported by the UK Royal Academy of Engineering under the Research Fellowship Scheme (RF/201718/17128) and EPSRC Grants EP/T00097X/1,EP/S026428/1. \textit{(Corresponding author: JaKeoung Koo.)}}%
\thanks{Kyungmin Choi and JaKeoung Koo are with the School of Computing, Gachon University, Seongnam, 13120, South Korea (e-mail: jiwoo33333@gachon.ac.kr; jakeoung@gachon.ac.kr)}
\thanks{Stephen McLauglin and Abderrahim Halimi are with the School of Engineering and Physical Sciences, Heriot-Watt University, Edinburgh, EH14 4AS, United Kingdom (e-mail: s.mclaughlin@hw.ac.uk; a.halimi@hw.ac.uk).}}

% \author{ChoiIEEE Publication Technology,~\IEEEmembership{Staff,~IEEE,}
% \thanks{This paper was supported by.}% <-this % stops a space

% for arxiv : below 2line

% for arxiv: below 1 line

\maketitle

\begin{abstract}

Single-photon Lidar imaging offers a significant advantage in 3D imaging due to its high resolution and long-range capabilities, however it is challenging to apply in noisy environments with multiple targets per pixel. To tackle these challenges, several methods have been proposed. Statistical methods demonstrate interpretability on the inferred parameters, but they are often limited in their ability to handle complex scenes. Deep learning-based methods have shown superior performance in terms of accuracy and robustness, but they lack interpretability or they are limited to a single-peak per pixel. In this paper, we propose a deep unrolling algorithm for dual-peak single-photon Lidar imaging. We introduce a hierarchical Bayesian model for multiple targets and propose a neural network that unrolls the underlying statistical method. To support multiple targets, we adopt a dual depth maps representation and exploit geometric deep learning to extract features from the point cloud. The proposed method takes advantages of statistical methods and learning-based methods in terms of accuracy and quantifying uncertainty. The experimental results on synthetic and real data demonstrate the competitive performance when compared to existing methods, while also providing uncertainty information.

%are built on the assumption of a single peak per pixel. In this paper, we propose a novel Bayesian deep unrolling algorithm for dual-peak single-photon Lidar imaging. We introduce a hierarchical Bayesian model that provides a basis for the deep learning model. The proposed network architecture unrolls the underlying statistical method, which is capable of handling multiple peaks in single-photon Lidar imaging. The network consists of multiple stages, each of which consists of two blocks: the squeeze block and the expansion block. The squeeze block estimates the true depth from a multiscale point cloud, while the expansion block refines the multiscale point cloud. To learn meaningful features from the point cloud, we construct a graph by applying the k-nearest neighbors algorithm and use Graph Attention Networks to extract features and compute attention weights. The network is trained end-to-end in a supervised manner. We evaluate the proposed method on synthetic and real data and show superior performance compared to existing methods.

\end{abstract}


\section{Introduction}

\IEEEPARstart{S}{ingle-photon} Lidar imaging is a versatile 3D imaging technique deployed in various applications such as autonomous driving due to its high resolution and long range capababilities~\cite{wallace2020full,rapp2020advances}. Single-photon Lidar systems work by sending laser pulses and recording the time-of-flight (ToF) of returning photons, using the time-correlated single-photon counting (TCSPC) technique~\cite{buller2007ranging}. Collecting the recorded photon counts, the system builds a histogram of photon counts with respect to ToFs. 
% With the presence of multiple targets, the histogram can contain multiple peaks \textcolor{blue}{due to factors, such as the partially transmissive targets (e.g., camouflage), divergence of the laser beam, and low resolution of the sensors.} 
The histogram can contain multiple peaks within a single pixel when partially transmissive targets (e.g. camouflage) are present, or when recorded photons originate from multiple targets due to the divergence of the laser beam and the low resolution of the sensors.
Locating such peaks is crucial for 3D reconstruction of the scene. However, the reconstruction process faces several challenges, including low photon counts in the histogram and the presence of background photons from ambient light.

Several methods have been proposed for 3D reconstruction from single-photon Lidar data. Existing studies can be categorized into two groups: statistical methods and deep learning-based methods. The former rely on statistical models~\cite{kirmani2014first,shin2015photonefficient} with some priors such as sparsity of data~\cite{halimi2016restoration,chen2020learning} or spatial smoothness in image representation~\cite{tachella2019bayesian}. With such prior models, the solution can be obtained by different strategies such as Stochastic Simulation~\cite{hernandez-marin2008multilayered,halimi2017object,tachella2019bayesian}, optimization~\cite{Pawlikowska_OE2017,rapp2017few,halimi2020robust,Tobin_SR21}, expectation-maximization~\cite{altmann2018range,legros2020expectationmaximization}, or Plug-and-Play methods~\cite{venkatakrishnan2013plugandplay,tachella2019realtime}. These methods provide interpretable results and often provide uncertainty information on the inferred parameters, but they are often limited to handling complex scenes and require user-defined hyper parameters. %Some methods~\cite{tachella2019realtime,belmekki2022fast} support reconstructing multiple targets. 

Deep learning-based methods offer an alternative data-driven approach. These methods train neural networks on large SPAD histogram data or compressed representation to infer scene parameters. With the significant advancements in network architecture design of the computer vision field~\cite{{ronneberger2015unet,zhou2020unet++,Chen_2018_ECCV,chen2020realworld,wang2018nonlocal,Yue2018CompactGN}}, many of these architectures have been directly or conceptually adopted in various studies\cite{{lindell2018singlephoton,sun2020spadnet,peng2020photonefficient,Zang:21,Zhao:22,peng2023penonlocal}} for single-photon Lidar imaging. Because these methods infer the scene by learning the mapping directly from large histogram data to depth profiles, they divide the histogram data into small patches and process them. Although these methods show good performance, they often lack interpretability of the inferred parameters, which limits in practical usage. An interpretable network BU3D in~\cite{koo2022bayesiana} was proposed, which combines a Bayesian algorithm~\cite{halimi2021robust} with deep learning via an unrolling approach. The algorithm unrolling is a framework~\cite{gregor2010learning,monga2021algorithm,yang2016deep,zhang2020deep} to bridge model-based and learning-based approaches. It transforms each stage of iterative methods into neural network layers, enabling the integration of domain knowledge directly into the network architecture. BU3D~\cite{koo2022bayesiana} provides a balance between accuracy and interpretability, providing uncertainty on the results. However, existing deep learning-based methods rely on image-based representation and build on the assumption of a single-peak per pixel that is not always valid in practice. 

\IEEEpubidadjcol % should be put in the second column of the first page
%
% In this paper, we propose an algorithm unrolling method for dual-peak single-photon Lidar imaging. We extend an existing Bayesian model~\cite{halimi2021robust} and its iterative algorithm for dual-peak imaging. Each step of the iterative method is unrolled into neural network layers. Unlike existing deep learning approaches that adopt image-based representations, our method employs a point-cloud representation for supporting dual peaks and utilizes graph neural networks (GNNs) for feature extraction. By combining the strengths of statistical modeling with the adaptability of deep learning, our architecture is both efficient and interpretable, providing uncertainty information on the inferred point clouds. An early version of the proposed work was presented in~\cite{koo2024bayesian}. We extend this conference paper to provide a new Bayesian model considering multiple peaks, improve network architecture, and add uncertainty estimation. %We also provide comprehensive experimental results, showing competitive resuls on synthetic and real data.
%^ original TCI
%
In this paper, we propose an algorithm unrolling method for dual-peak single-photon Lidar imaging. We extend an existing Bayesian model~\cite{halimi2021robust} and its iterative algorithm for dual-peak imaging. Each step of the iterative method is unrolled into neural network layers. Unlike existing deep learning approaches that adopt image-based representations, our method employs a point-cloud representation for supporting dual peaks and utilizes graph neural networks (GNNs) for feature extraction. By combining the strengths of statistical modeling with the adaptability of deep learning, our architecture is both efficient and interpretable, providing uncertainty information on the inferred point clouds. An early version of the proposed work was presented in~\cite{koo2024bayesian}. We extend this conference paper to provide a new Bayesian model considering multiple peaks, add uncertainty estimation, and improve the network architecture by incorporating an explicit hard attention mechanism and modifying the graph attention layer design.

Processing large histogram data requires a large amount of memory and high computational cost. To address this issue, we adopt a multiscale approach~\cite{halimi2021robust}, where we downsample the histogram data and estimate the dual peaks from each downsampled histogram. The estimated dual peaks are used as initial multiscale point clouds, which is the input to the proposed network. In this regard, we compressed large histogram data~\cite{sheehan2021sketchinga,gutierrez-barragan2023learned} into a multiscale point cloud representation, which is more efficient to process and requires less memory. The proposed network layers sequentially estimate the true depth from the multiscale point cloud and refine the multiscale point cloud. The components of layers rely on soft attention and hard attention inspired by median filtering in the Bayesian algorithm~\cite{halimi2021robust}. The proposed method is evaluated on synthetic and real data and shows superior performance compared to existing methods with uncertainty information, a key advantage of the proposed method. Our code is available at \url{https://github.com/daedalus-KM/PointcloudUnrolling}.

% In summary, the contributions of this paper are as follows:
% \begin{itemize}
%     \item 
% \end{itemize}

The paper is organized as follows. Section~\ref{sec:2} introduces an observation model for multi-peak single-photon Lidar imaging. In Section~\ref{sec:3}, an iterative Bayesian model is proposed. This iterative method is unrolled into the deep learning model in Section~\ref{sec:4}. In Section~\ref{sec:5}, we evaluate the proposed method on synthetic and real data. Section~\ref{sec:6} concludes the paper.

\section{Multiscale multi-peak observation model} \label{sec:2}

In this section, we introduce the observation model for multi-peak single-photon Lidar imaging, which will be included in the Bayesian model in Section~\ref{sec:3}. The single-photon Lidar system measures the range of a scene by illuminating the scene and recording the time it takes for returning photons to reach the sensor. The returned photon counts $y_{n,t}$ are assumed to follow a Poisson distribution $y_{n,t} \sim \mathcal{P}(s_{n,t})$, where $s_{n,t}$ is the average photon in the $n$-th pixel and in the $t$-th time bin~\cite{shin2015photonefficient,rapp2017few}. Assuming the presence of $K$ targets per pixel, we consider the observation model for $s_{n,t}$ as follows:
\begin{equation} \label{eq:snt}
s_{n, t} = \sum_{k=1}^K r_{n,k} \, g (t-d_{n,k}) + b_{n},
\end{equation}
where $r_{n,k}$ and $d_{n,k}$ are the reflectivity and the depth value of the $k$-th target, respectively, $b_n$ is the background photons from ambient light or detector noises and $g$ is the system impulse response function (IRF). 
We assume that the background photons are absent for simplicity.
% We assume that $\sum_{k=1}^K r_{n,k}=1$ and the background photons are absent for simplicity. 
This IRF is commonly approximated by the Gaussian function $\mathcal N(t\, ; \mu, \sigma^2)$ with the mean $\mu$ and the variance $\sigma^2$, satisfying $\sum_{t=1}^T g\left(t-d_{n,k}\right)=1$~\cite{halimi2016restoration,halimi2021robust}. With the assumption of independent observations with respect to $n$ and $t$, the likelihood function of the observation model is given by
\begin{equation} \label{eq:p(Y)}
p\,(\boldsymbol{Y} \mid \boldsymbol{d}, \boldsymbol{r})=\prod_{n=1}^{N} \prod_{t=1}^{T} \frac{s_{n, t}^{y_{n, t}}}{y_{n, t} !} \exp ^{-s_{n, t}},
\end{equation}
where $\boldsymbol{Y} = \{y_{n,t}\}$, $\boldsymbol{d} = \{d_{n,k}\}$ and $\boldsymbol{r} = \{r_{n,k}\}$.
This likelihood function~\eqref{eq:p(Y)} for the $n$-th pixel can be rewritten as
\begin{equation}
\begin{aligned}
    p \left(\boldsymbol{y}_n \mid \boldsymbol{r}_n, \boldsymbol{d}_n \right)  \propto & \prod_t \left(\sum_{k} r_{n, k} \, g (t-d_{n,k}) \right)^{ y_{n,t}} \frac{1}{y_{n, t}!}\\ 
    & \times \prod_{k} \exp \left(-r_{n, k}  \right).
\end{aligned}
\end{equation}
Applying the Jensen's inequality on the log-likelihood, the lower bound of the likelihood function is approximated as
\begin{equation}
\begin{aligned}
p\left(\boldsymbol{y}_n \mid \boldsymbol r_n, \boldsymbol d_n \right) \geq & \prod_k \left[ \left(r_{n, k}\right)^{\bar{r}_{n}} \exp(-r_{n,k}) \right]  Q\left(\boldsymbol y_{n}\right) \\
& \times \prod_{t,k} g(t-d_{n,k})^{y_{n,t}},
\end{aligned}
\end{equation}
where $Q$ is a function of $\boldsymbol{y}_{n}$ and $\bar{r}_n := \frac{1}{K} \sum_{k=1}^K r_{n, k}$ approximating the scaled total photon counts in the $n$-th pixel $\frac{1}{K} \sum_{t=1}^T y_{n,t}$. The approximated lower bound of the likelihood function can then be rewritten as
\begin{equation}
\begin{aligned}
p \, ( \boldsymbol y_{n} \mid  \boldsymbol r_{n}, \boldsymbol d_{n} ) \geq \prod_{k} & \left[ \mathcal{G}\left(r_{n,k} ; 1+\bar{r}_n, 1\right) \mathcal{N} (d_{n,k} ; d_{n,k}^{\mathrm{ML}}, \bar{\sigma}^{2}) \right] \\[-7pt]
 & \times \overline Q \left( \boldsymbol{y}_{n}\right), 
\end{aligned} 
\label{eq:observation}
\end{equation}
where $\mathcal{G}$ is a gamma distribution, $\bar{\sigma}^{2}= \sigma^2/ \bar{r}_n$, $\overline Q$ is a normalization factor and $d^{\mathrm{ML}}_{n,k}$ is the maximum likelihood (ML) estimation of the depth on the approximated lower bound. This ML estimation, assuming that each photon count belongs to a single target, is given by
\begin{equation}
d^{\mathrm{ML}}_{n,k} = \arg\max_d \sum_{t} h_{n,t,k} y_{n,t} \log \, ( g(t-d)).
\end{equation}
where $h_{n,t,k}$ is an indicator function for the $k$-th target, where $h_{n,t,k}=1$ if $y_{n,t}$ belongs to the $k$-th target and $h_{n,t,k}=0$ otherwise. 

\textbf{Multiscale model.} To address low-photon or noisy data, we employ a multiscale approach~\cite{rapp2017few,halimi2021robust,lindell2018singlephoton,peng2020photonefficient}. By utilizing the fact that the low-pass filtering of histograms by summing neighboring pixels still follows a Poisson distribution,  we consider $L$ downsampled histograms $\boldsymbol y_n^{(\ell)}$ with $\ell \in \{ 2, \cdots, L\}$ corresponding to different kernel sizes and approximate the lower bound of the likelihood for each histogram as
\begin{equation} \label{eq:multiscale}
\begin{aligned}
    p \, ( \boldsymbol y_{n}^{(\ell)} \mid  \boldsymbol r_{n}^{(\ell)}, \boldsymbol d_{n}^{(\ell)} ) \geq 
\prod_{k} & \mathcal{G}\left(r_{n,k}^{(\ell)} ; 1+\bar{r}^{(\ell)}_n, 1\right) \mathcal{N} (d_{n,k}^{(\ell)} ; d_{n,k}^{\mathrm{ML} (\ell) }, \bar{\sigma}^{2})  \\[-7pt]
    & \times \overline Q \left( \boldsymbol{y}_{n}^{(\ell)}\right).
\end{aligned}
\end{equation}

\section{Underlying Bayesian model} \label{sec:3}

In this section, we introduce a hierarchical Bayesian model that provides a basis for the deep learning model in Section~\ref{sec:4}. To reconstruct a scene in challenging scenarios, we impose prior information on the parameters of interest and consider the posterior distribution. The proposed hierarchical model is similar to~\cite{halimi2021robust,koo2022bayesiana}, but our model extends to the situation of multiple peaks per pixel. Although the proposed model is capable of handling both depth and reflectivity, we focus only on depth estimation.

\subsection{Prior and posterior distributions}

From the multiscale observation model~\eqref{eq:multiscale}, we aim to estimate the true depth, a latent variable, denoted by $\{x_{n,k}\}$ for each pixel $n$ and each target $k$. To achieve this, we impose spatial smoothness within homogeneous regions on the latent variable, while preserving boundaries of the targets. Extending the approach in~\cite{halimi2021robust} to multiple targets, we define some guidance weights, to relate multiscale depths with the latent variable, as $\boldsymbol W=\{ w_{\nu_n, n,k}^{(\ell)}\}$. These are pre-defined values assigned between neighbouring pixels $\nu_{n}$ of the $n$-th pixel at the scale $\ell$ and the target $k$. Large values of $w_{n^{\prime},n, k}^{(\ell)}$ suggest the latent variable $x_{n,k}$ to be similar to $d_{n^{\prime},k}^{(\ell)}$. With these guidance weights, the latent variable $x_{n,k}$ is designed to follow the conditional Laplace distribution:
\begin{equation} \label{eq:prior_x}
\begin{array}{c}
x_{n,k} \mid d_{\nu_{n,k}}^{(1, \cdots, L)}, w_{\nu_n, n, k}^{(1, \cdots, L)}, \epsilon_{n,k} \sim \\
\prod_{n^{\prime} \in \nu_{n}}\left[\prod_{\ell=1}^{L} \mathcal{L}\left(x_{n,k} ; d_{n^{\prime},k}^{(\ell)}, \frac{\epsilon_{n,k}}{w_{n^{\prime}, n, k}^{(\ell)}}\right)\right]
\end{array}
\end{equation}
where $\mathcal L(x\cdot\, ; \mu, \epsilon)=1/(2\epsilon) \exp(-|x-\mu|/\epsilon)$ is the Laplace distribution with the mean and the scale parameter, $\epsilon_{n,k}$ is proportional to the variance of the latent variable $x_{n,k}$. The parameter $\epsilon_{n,k}$ is constrained to be positive, and is assigned a conjugate inverse gamma distribution as
\begin{equation} \label{eq:prior_eps}
\boldsymbol \epsilon \sim \prod_{n=1}^N \prod_{k=1}^K \mathcal{I} \mathcal{G}\left(\epsilon_{n,k} ; \alpha_d, \beta_d \right)
\end{equation}
where $\boldsymbol \epsilon=\{\epsilon_{n,k}\}$, and $\alpha_d,\beta_d$ are positive hyperparameters.

Let $\mathbf{D} = [\boldsymbol{d}^{(1)}, \boldsymbol{d}^{(2)}, \ldots, \boldsymbol{d}^{(L)}]$.
We have derived the likelihood function $p\,(\boldsymbol{Y} \mid \boldsymbol{D})$ in~\eqref{eq:multiscale} and the prior distribution $p(\boldsymbol x, \boldsymbol D \mid \boldsymbol \epsilon, \boldsymbol W)$ in~\eqref{eq:prior_x} and $p(\boldsymbol \epsilon)$ in~\eqref{eq:prior_eps}, which leads to the posterior distribution proportional to
\begin{equation} \label{eq:posterior}
\begin{aligned}
p\,(\boldsymbol{x}, \boldsymbol{\epsilon}, \boldsymbol{D} \mid \boldsymbol{Y}, \boldsymbol{W}) \propto \,& p\,(\boldsymbol{Y} \mid \boldsymbol{D}) \, p\,(\boldsymbol{x}, \boldsymbol{D} \mid \boldsymbol{\epsilon}, \boldsymbol{W}) \, p \,(\boldsymbol{\epsilon}).
\end{aligned}
\end{equation}
Based on the posterior distribution, we seek to find the maximum a posteriori (MAP) estimate.

\subsection{Iterative algorithm}

To solve the MAP estimates, a coordinate descent algorithm is adopted to minimize the negative log-posterior of~\eqref{eq:posterior}. As summarized in Algorithm~\ref{alg1}, this iterative algorithm updates alternatively the latent variable $x_{n,k}$, the multiscale depths $\boldsymbol d^{(1,\cdots,L)}$, and the depth uncertainty $\boldsymbol \epsilon$, while keeping the other variables fixed during each update step. The details of each update step are provided in the following.

\textbf{Squeeze step.} The update of the latent variable $x_{n,k}$ is performed by weighted median filtering:
%
% \begin{equation} \label{eq:xnargmin} 
% x_{n,k}  \leftarrow \underset{x}{\operatorname{argmin}}  \, \mathcal{C}(x) = \sum_{\ell, n^{\prime} \in \nu_{n}} w_{n^{\prime}, n,k}^{(\ell)}\left|x-d_{n^{\prime},k}^{(\ell)}\right|.
% \end{equation}
% ^ original TCI
%

\begin{equation} \label{eq:xnargmin} 
x_{n,k}  \leftarrow \underset{x}{\operatorname{argmin}}  \, \mathcal{C}(x), \, \text{where }
\mathcal{C}(x) = \sum_{\ell, n^{\prime} \in \nu_{n}} w_{n^{\prime}, n,k}^{(\ell)}\left|x-d_{n^{\prime},k}^{(\ell)}\right|.
\end{equation}

This operation will be replaced by attention mechanisms in the proposed deep learning model in Section~\ref{sec:4}.
\textbf{Expansion step.} The multiscale depths $\boldsymbol d^{(1,\cdots,L)}$ appear in the likelihood~\eqref{eq:observation} and the prior~\eqref{eq:prior_x}. Considering minimizing the negative logarithm of the posterior distribution, the update of the multiscale depths is given by
\begin{equation} \label{eq:dnargmin}
d_{n,k}^{(\ell)}  \leftarrow \underset{d}{\operatorname{argmin}} \frac{\left[d-d_{n,k}^{\mathrm{ML}(\ell)}\right]^{2}}{2 \bar{\sigma}^{2(\ell)}}+\sum_{n^{\prime} \in \nu_{n}} \frac{w_{n, n^{\prime},k}^{(\ell)}\left|d-x_{n^{\prime}}\right|}{\epsilon_{n^{\prime},k}}.
\end{equation}
This operator is known as a generalized soft-thresholding and can be solved analytically~\cite{parikh2014proximal}. It will be replaced by the expansion block in the proposed network in Section~\ref{sec:4}. 

\textbf{Uncertainty estimation.} The depth uncertainty variable $\boldsymbol \epsilon$ is estimated by its conditional distribution:
\begin{equation} \label{eq:epsilon_nk}
\epsilon_{n,k} \mid \boldsymbol{x}, \boldsymbol{D}, \boldsymbol{W} \sim \mathcal{I} \mathcal{G}\left[L \bar{N}+\alpha_{d}, \, \mathcal{C}\left(x_{n,k}\right)+\beta_{d}\right],
\end{equation}
where $\mathcal C(x)$ is given in Eq.~\eqref{eq:xnargmin}, $\bar N=|\nu_n|$ is the number of neighbors considered. The mode of this distribution is given by
%iffalse
\begin{equation} \label{eq:epsilon}
\hat{\epsilon}_{n,k} \leftarrow (\mathcal{C}\left(x_{n,k}\right)+\beta_{d}) / (L  \bar{N}+\alpha_{d}+1).
\end{equation}
This equation will be used to estimate the uncertainty of the depth map in the proposed deep learning model in Section~\ref{sec:4}.

\begin{algorithm}[ht]
\caption{Underlying statistical method} \label{alg1}
\begin{algorithmic}[1]
    \STATE \underline{Input}: Histogram data $\boldsymbol Y$, the number of scales $L$
    \STATE Construct low-pass filtered histograms $\boldsymbol Y^{(1,...,L)}$
    \STATE Compute multiscale multi-peak depths $\boldsymbol{d}^{\mathrm{ML}(1,...,L)}$
    \STATE Compute the guidance weights $\boldsymbol W$
    \STATE \textbf{Iterate} until convergence
        \STATE \phantom{ab} Squeeze step: Update $\boldsymbol x$ by \eqref{eq:xnargmin} %, \eqref{eq:mnln}
        \STATE \phantom{ab} Expansion step: Update $\boldsymbol d^{(1,\cdots, L)}$ by \eqref{eq:dnargmin} %, \eqref{eq:rnargmin}
    \STATE Update the uncertainty information $\boldsymbol \epsilon$ by \eqref{eq:epsilon} %, 
	\STATE \underline{Output}: $\boldsymbol x, \boldsymbol \epsilon$  
\end{algorithmic}
\end{algorithm} 

\begin{algorithm}[ht]
    \caption{Proposed unrolling method} \label{alg2}
    \begin{algorithmic}[1]
        \STATE \underline{Input}: Histogram data $\boldsymbol Y$, the number of scales $L$
        \STATE Construct downsampled histograms $\boldsymbol Y^{(1,...,L)}$
        \STATE Compute multiscale dual-peak depths $\boldsymbol{d}^{\mathrm{ML}(1,...,L)}$
        \STATE For $s=1$ to $S-1$
            \STATE \phantom{ab} Squeeze step: Update $\boldsymbol x$
            \STATE \phantom{ab} Expansion step: Update $\boldsymbol d^{(1,\cdots, L)}$
        
        \STATE Squeeze step: Update $\boldsymbol x$
        % \STATE Squeeze step: Update $\boldsymbol x$
        \STATE Compute the uncertainty information $\boldsymbol \epsilon$ by \eqref{eq:epsilon} %, 
        \STATE \underline{Output}: $\boldsymbol x, \boldsymbol \epsilon$  
    \end{algorithmic}
\end{algorithm} 

% -----------------------------------------
% Architecture
\def\fh{290pt}
\begin{figure*}[h]\center
\includegraphics[totalheight=\fh,trim={0 0.0cm 0 0},clip]{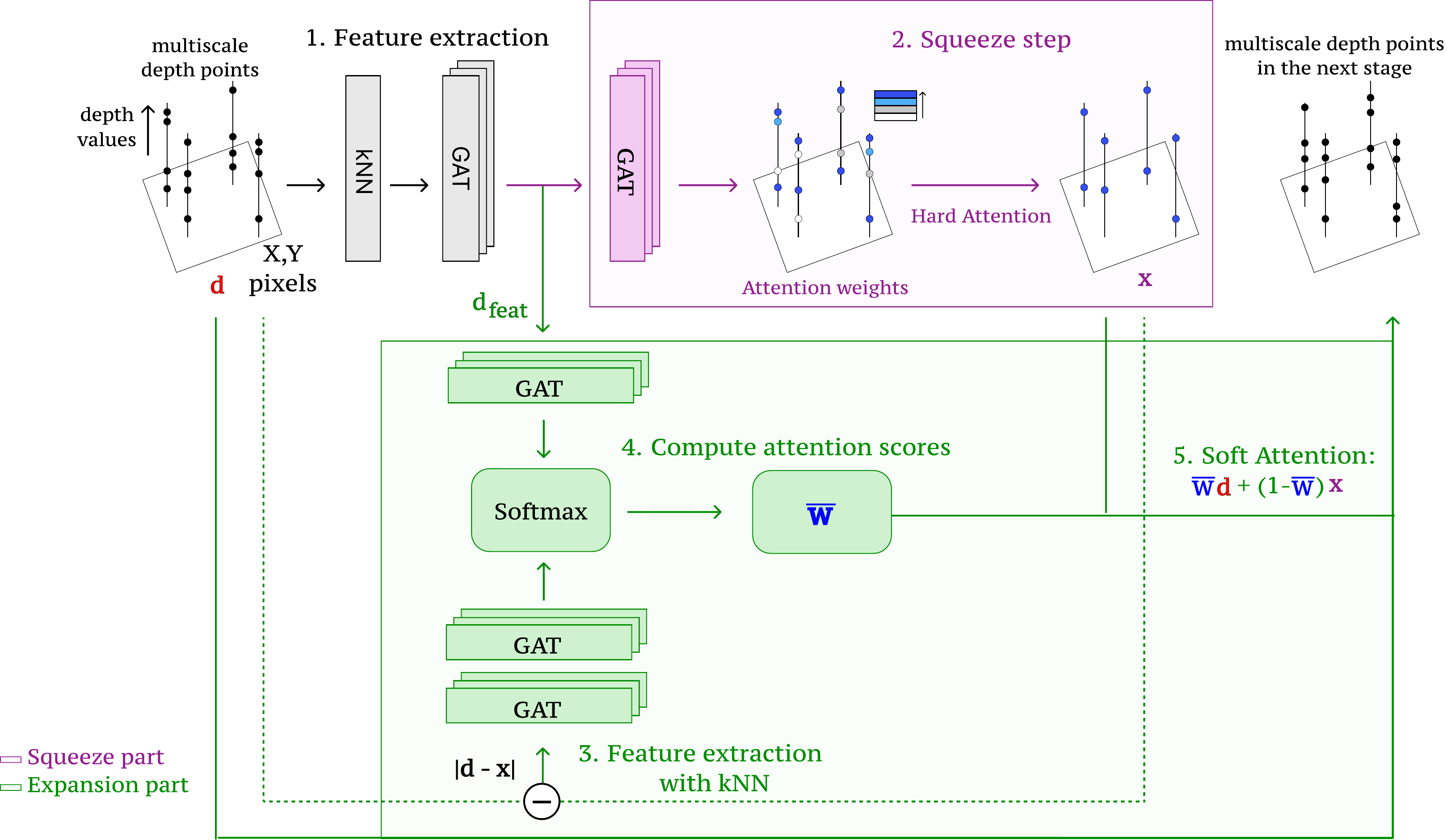}
\caption{Overview of the proposed network architecture for one stage. GAT stands for Graph Attention Networks and kNN stands for k-Nearest Neighbor.}
\label{fig:architecture}
\end{figure*}

\section{Proposed unrolling method} \label{sec:4}

We propose a neural network that unrolls the underlying statistical method introduced in Section~\ref{sec:3}. Although the statistical method is capable of handling multiple peaks, we will focus on the dual-peak case. By leveraging a point cloud representation, we effectively capture the dual-surface per pixel. The overall procedure of the proposed method is summarized in Algorithm~\ref{alg2}.
%Through this representation, we transform the weighted median filtering in Bayesian algorithm as a hard-attention mechanism implemented via Graph Attention Networks (GATs).

\subsection{Network}

As illustrated in Fig.~\ref{fig:architecture}, the architecture takes as input an initial multiscale point cloud, denoted by $\boldsymbol d$, instead of the large volume histogram data. From the multiscale point cloud, the network aims to estimate true depth map $\boldsymbol x$. The network consists of $S$ stages, where each stage consists of two blocks: the squeeze block and the expansion block. The squeeze block estimates the true depth $\boldsymbol x$ from a multiscale point cloud $\boldsymbol d$. The expansion block refines the multiscale point cloud by applying a weighted average between the initial multiscale point cloud and the squeezed point cloud. In the final stage, the network only estimates the squeezed point cloud without refining the multiscale point cloud. The network is trained end-to-end in a supervised manner.

To learn meaningful features from the multiscale point cloud, we construct a graph by applying the k-nearest neighbors (kNN) algorithm with $k=6$. On the constructed graph, we use Graph Attention Networks (GATs)~\cite{velickovic2018graph} with a single head to extract features from the point cloud and compute attention weights. The GAT generalizes the convolutional operation on the graph, allowing the network to learn the importance of each neighboring point.

%To process the point cloud with convolutional operations, we construct a graph by applying k-nearest neighbors (kNN) algorithm with $k=6$. On the constructed graph, we use Graph Attention Networks, known as GAT~\cite{velickovic2018graph}. The GAT enables to learn meaningful features from the point cloud and compute attention weights, which are used to perform hard attention on the point cloud. The network is trained to minimize the $L_1$-norm loss function between the estimated depth map and the ground truth depth map.

\textbf{Sequeeze block.} We extract features from the constructed graph through three layers of GATs. These extracted features, denoted as $\boldsymbol {d_{feat}}$, are then employed to compute attention weights via three additional layers of GATs. Using the computed attention weights, we apply hard attention to the initial multiscale point cloud $\boldsymbol d$, selecting the scale with the highest attention weight for each of the first and second surfaces individually, thereby identifying the most important scales for each surface. For hard attention, we use a differentiable argmax operation known as Gumbel-Softmax proposed in~\cite{maddison2017concrete,jang2017categorical}. This selective process yields the squeezed point cloud $\boldsymbol x$.

\textbf{Expansion block.}  From the squeezed point cloud $\boldsymbol x$, we compute the residual $|\boldsymbol d - \boldsymbol x|$ and extract features from it. These extracted features are used to compute new attention weights. These attention weights serve to perform a weighted average between multiscale point cloud $\boldsymbol d$ and the squeezed point cloud $\boldsymbol x$, refining the multiscale point cloud. 

% -----------------------------------------

\subsection{Estimation of Initial Multiscale Dual Peaks}

To estimate the initial multiscale point cloud from the original histogram data, we apply the cross-correlation to the histogram by the IRF. To the resulting histogram, we apply the convolution spatially with 4 uniform filters whose sizes are $\{1 \times 1,\, 3 \times 3,\, 7 \times 7,\, 13 \times 13 \}$. This spatial convolution yields low-pass filtered histograms $\boldsymbol Y^{(1,2,3,4)}$. To support reconstructing two surfaces per pixel, we estimate dual peaks from each histogram $\boldsymbol Y^{(1,2,3,4)}$, as the input to our network is an initial multiscale point cloud rather than the large volume histogram. 

For each multi-scale histogram $\boldsymbol Y^{(\ell)}$, we estimate the first peak of the histogram. We then remove the histogram counts within the vicinity of the estimated peak, corresponding to the width of the IRF. From the adjusted histogram, we estimate the second peak of the histogram.  After having two peak estimates, we consider the smaller depth estimate as the first depth and the larger estimate as the second depth. Consequently, the initial multiscale point cloud $\boldsymbol d_{n}$ is constructed for each pixel $n$, where the input depth $\boldsymbol d_{n} \in \mathbb R^{10}$ contains the $x$ and $y$ coordinates along with the 8 depth values, derived from the 4 scales, each yielding two depth estimates.

\subsection{Uncertainty estimation}

Following the underlying Bayesian model, we estimate the uncertainty of the depth map produced by the proposed network. Motivated by the mode of the depth variance in~\eqref{eq:epsilon}, we define the uncertainty as follows:
\begin{equation} \label{eq:epsilon_ours} \small
\epsilon_{n,k} = \frac{1}{S-1} \sum_{s=1}^{S-1} \frac{\mathcal C^s_{n,k} + \beta_d }{L + 2 + \alpha_d}, \,\, \mathcal C^s_{n,k} = \sum_{\ell=1}^L \overline{\overline{w}}_{n,k}^{s,(\ell)} |d^{s,(\ell)}_{n,k} - x^S_{n,k} |,
\end{equation}
where $S$ is the number of stages in the network, $\overline{\overline{\boldsymbol w}}^s_{:,k}$ is the normalized version of $1 -  \overline{\boldsymbol w}^k$ by softmax operation in terms of scales $\ell$. %, ensuring $\sum_{\ell=1}^L \overline{\overline{w}}_n^{k,(\ell)}=1$
The hyperparameters $\alpha_d,\beta_d$ are set to zeros in the experiments.
%, ensuring $\sum_{\ell=1}^L \overline{\overline{w}}_n^{k,(\ell)}=1$

\subsection{Training procedure}

We generate synthetic histogram data with dual peaks for training, using the scenes with depth and reflectivity from the Sintel dataset~\cite{butler2012naturalistic}. Each histogram data contains two identical surfaces, where the first surface lies between 1 and 300 time bins and the duplicated surface lies between 400 and 700 time bins. To simulate various challenging scenarios, we consider different levels of the Photons Per Pixel, $\text{PPP}=\frac{1}{N}\sum_{n=1}^N \left(r_n + b_{n} T\right)$ and Signal-to-Background Ratio, $\text{SBR}=(\sum_{n=1}^N r_n) / (\sum_{n=1}^N  b_{n} T)$. We consider the pairs of PPP and SBR as $\{(64, 64), (64, 4), (4, 64), (4, 4)\}$. 

We train the network with the Adam optimizer with the learning rate of $5 \times 10^{-5}$, the batch size of 24 and 100 epochs. The loss function is the $L_1$-norm function between the estimated depth map and the ground truth depth map.

% \textbf{Loss function.} We use the $L_1$-norm loss function to measure the difference between the estimated depth map and the ground truth depth map. The loss function is defined as
% %
% \begin{equation}
%     \mathcal{L} = \frac{1}{N} \sum_{n=1}^N \left\| \boldsymbol d_n - \boldsymbol d_n^* \right\|_1,
% \end{equation}
% %
% where $\boldsymbol d_n$ and $\boldsymbol d_n^*$ are the estimated and the ground truth depth maps, respectively.

\section{Experiments} \label{sec:5}

In this section, we evaluate the proposed method on synthetic and real data with dual peaks. 

\subsection{Results on synthetic data}

\begin{figure}[t]
    \centering
    \includegraphics[totalheight=100pt]{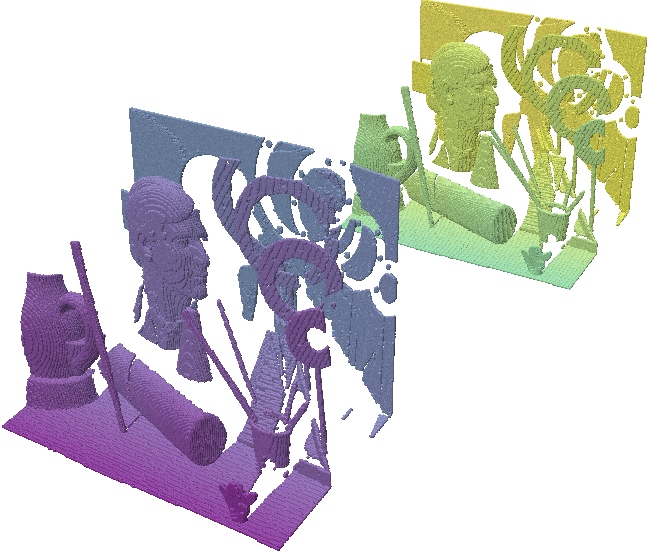}
    \includegraphics[totalheight=100pt]{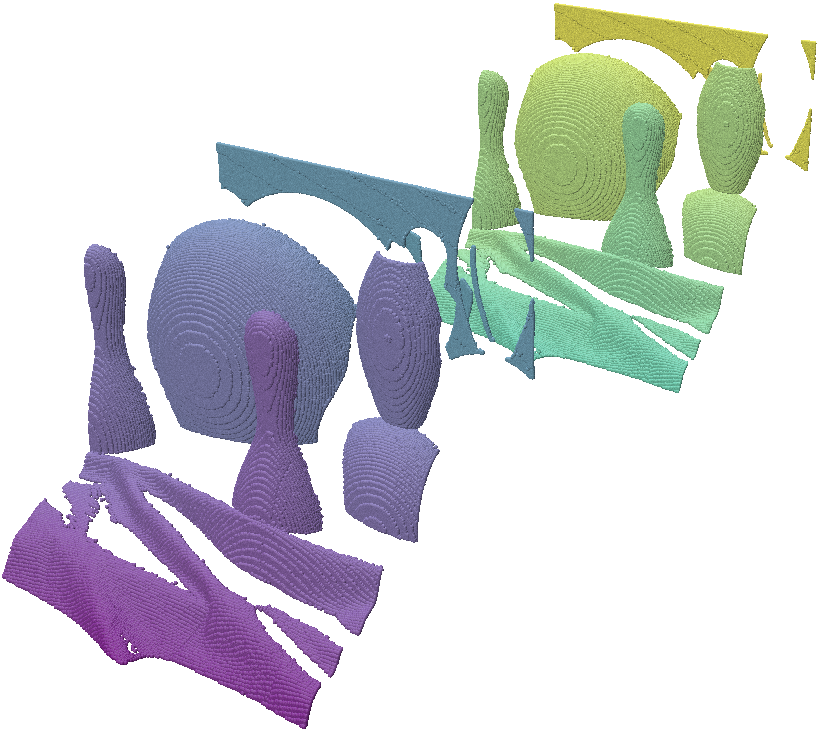}
    \caption{Ground truth point clouds for the Art scene (Left) and the Bowling scene (Right) with dual peaks.}
    \label{fig:gt}
\end{figure}

\textbf{Test dataset.} To evaluate the proposed method, we simulate the SPAD histogram data with two peaks. We used two scenes of Art and Bowling from the Middlebury dataset~\cite{hirschmuller2007evaluation} for reflectivity and depth profiles, with missing depth values carefully interpolated using median filtering. Fig.~\ref{fig:gt} shows the ground truth point clouds for the Art and Bowling scenes. Each scene contains two identical surfaces per pixel, where the first surface lies between 1 and 300 time bins and the duplicated surface lies between 400 and 700 time bins. The total time bins is 1024 which corresponds to 3.072 meter.

% \textbf{Comparison methods.} We compare the proposed method with other single-photon Lidar imaging methods by Lindell et al.~\cite{lindell2018singlephoton}, Peng et al.~\cite{peng2020photonefficient}, and BU3D~\cite{koo2022bayesiana}. As these state-of-the-art methods consider one surface per pixel, they are not directly applicable to the dual-peak reconstruction. We run these methods twice to estimate the depth of the first and the second surfaces separately, assuming the interval of each surface. In the first run, the histogram data is modified to eliminate the second surface, copying uniform background photons onto the time bins between 400 and 700. In the second run, the histogram is modified to eliminate the first surface, copying uniform background photons onto the time bins between 0 and 300. Note that, when running the proposed method, we do not modify the histogram data and directly estimate the dual peaks.
% ^ original TCI
\textbf{Comparison methods.}
We compare the proposed method with other state-of-the-art single-photon Lidar imaging methods by RT3D~\cite{tachella2019realtime}, Lindell et al.\cite{lindell2018singlephoton}, Peng et al.\cite{peng2020photonefficient}, and BU3D~\cite{koo2022bayesiana}. While RT3D supports multi-surface reconstruction, other three methods consider one surface per pixel and are thus not directly applicable to the dual-peak reconstruction.
% As these state-of-the-art methods consider one surface per pixel, they are not directly applicable to the dual-peak reconstruction. 
We run these methods~\cite{lindell2018singlephoton,peng2020photonefficient,koo2022bayesiana} twice to estimate the depth of the first and the second surfaces separately, assuming the interval of each surface. In the first run, the histogram data is modified to eliminate the second surface, copying uniform background photons onto the time bins between 400 and 700. In the second run, the histogram is modified to eliminate the first surface, copying uniform background photons onto the time bins between 0 and 300. Note that, when running the proposed method, we do not modify the histogram data and directly estimate the dual peaks.

% #GT Left-Right

% -------------------------------------
% Point clouds for art
\begin{figure*}[tp]
    \centering
    \def\fh{75pt} %RT3D없을땐 90
    \def\ppp{16.0} \def\sbr{4.0}
    \def\pppa{4.0} \def\sbra{4.0}
    \def\pppb{4.0} \def\sbrb{1.0}
    \def\scene{Art}
    
    \begin{tabular}{c@{ }c@{ }c@{ }c@{ }c@{ }c}
    
    \rotatebox[origin=l]{90}{\small\parbox{4cm}{$\,$PPP $=\ppp$, SBR$=\sbr$}} &
    \includegraphics[totalheight=\fh]{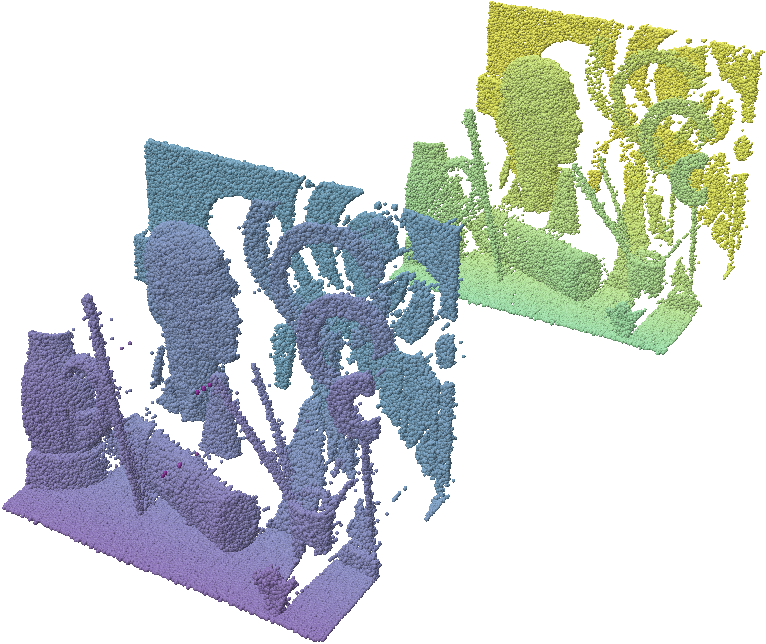} & %RT3D
    \includegraphics[totalheight=\fh]{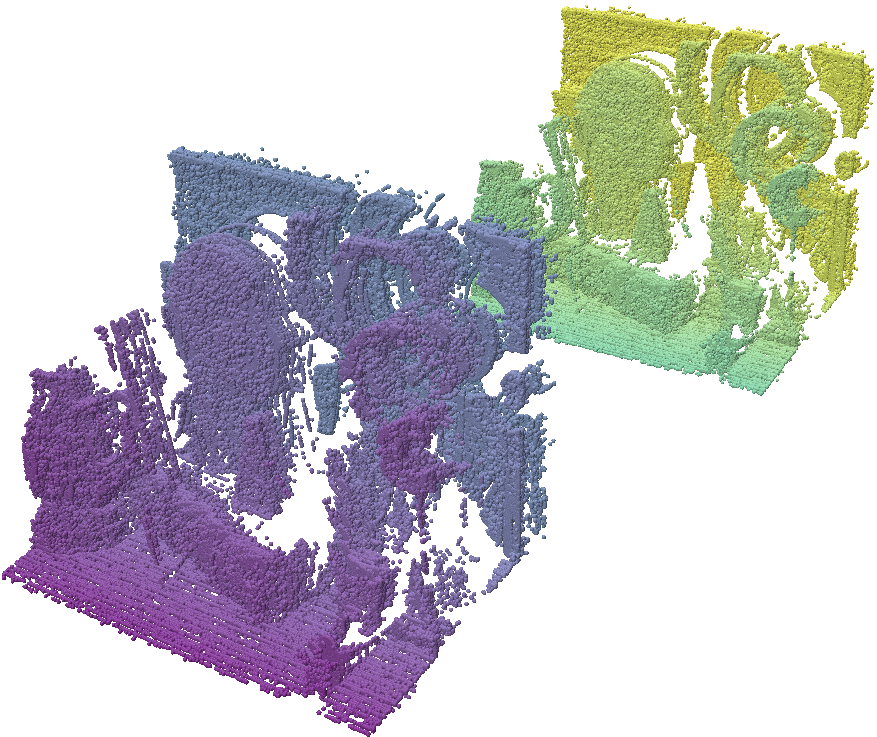} & %lindell
    \includegraphics[totalheight=\fh]{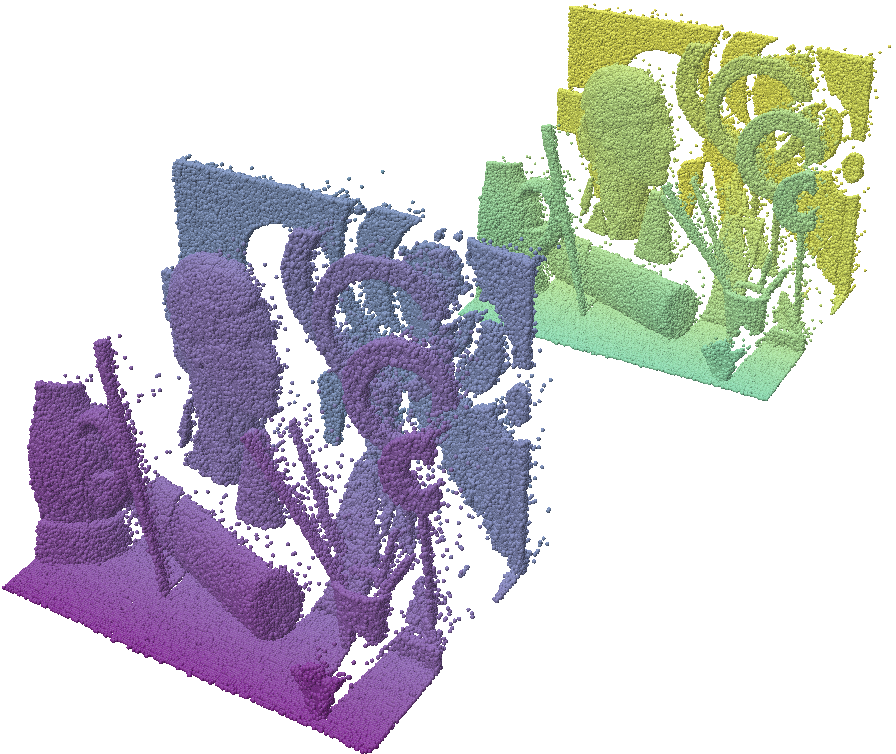} & %Peng
    \includegraphics[totalheight=\fh]{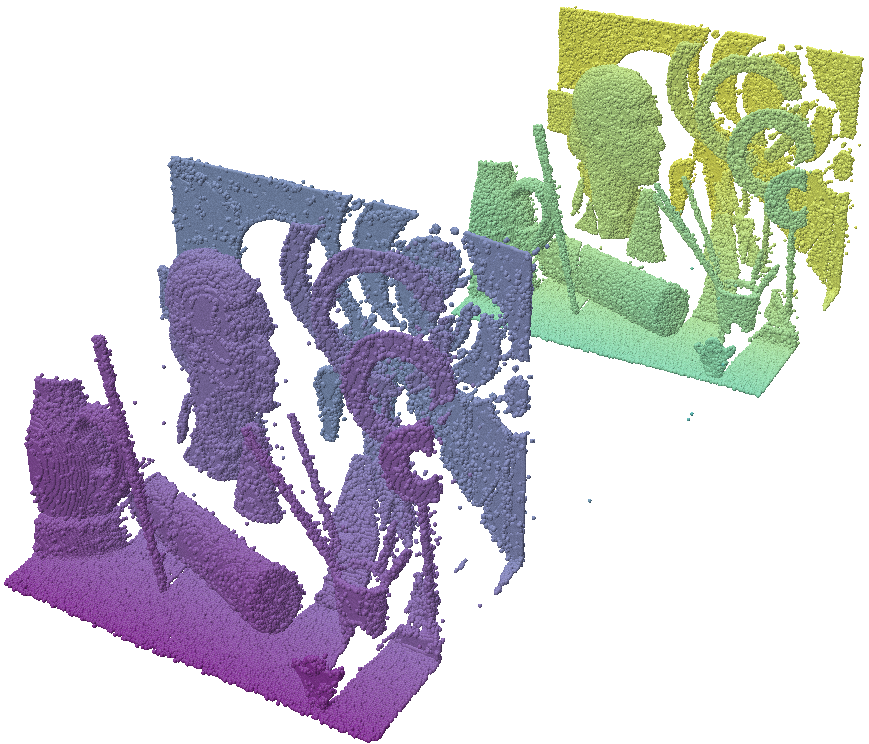} & % BU3D
    \includegraphics[totalheight=\fh]{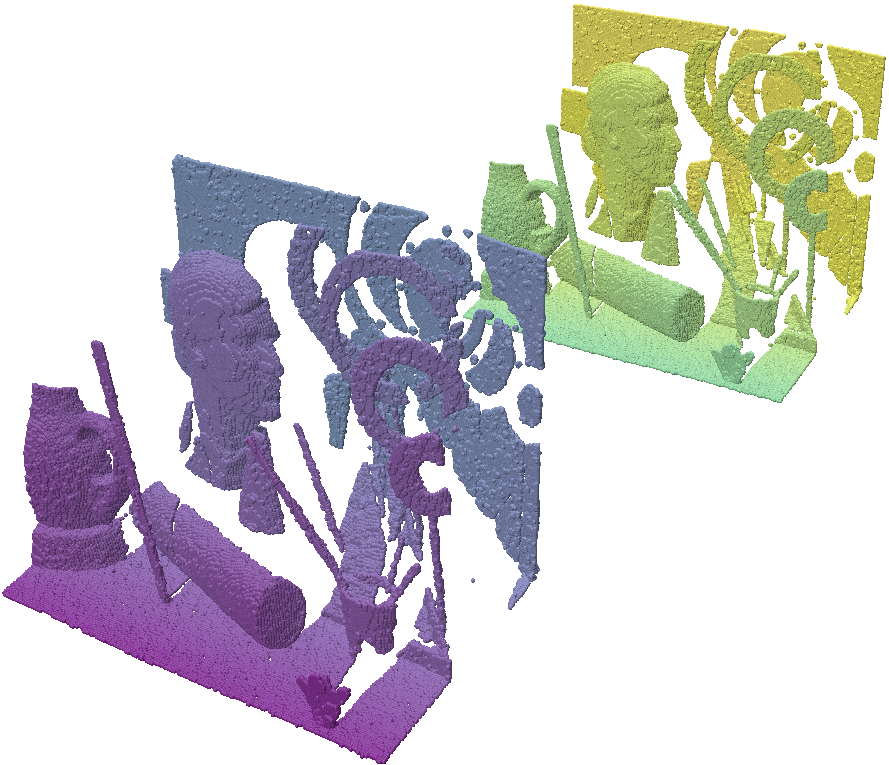} \\[-15pt]
    
    \rotatebox[origin=l]{90}{\small\parbox{4cm}{$\,$PPP $=\pppa$, SBR$=\sbra$}} &
    \includegraphics[totalheight=\fh]{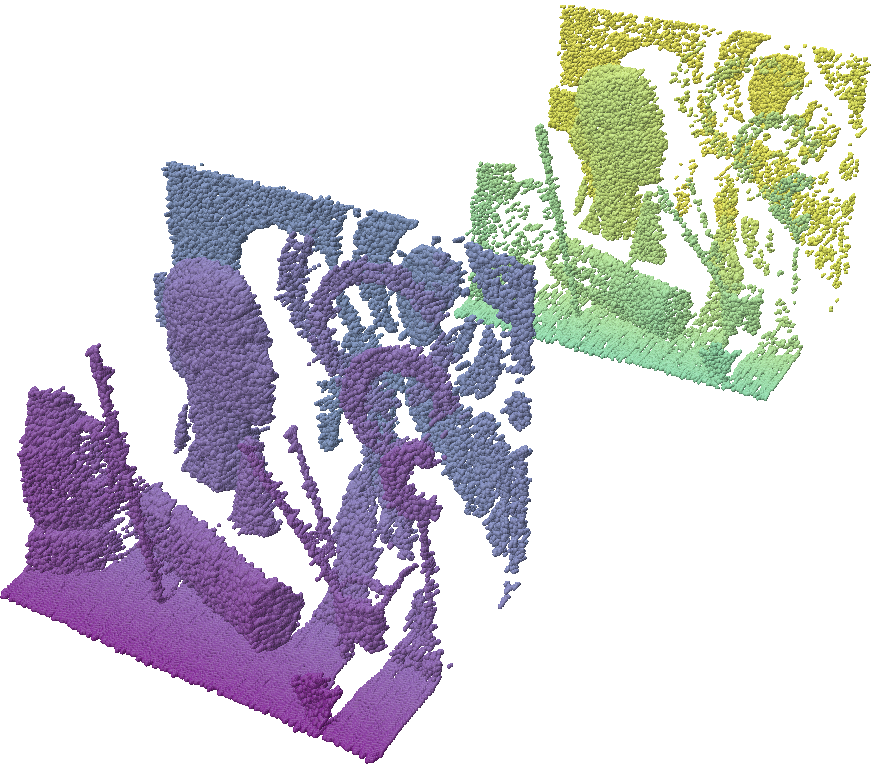} & %RT3D
    \includegraphics[totalheight=\fh]{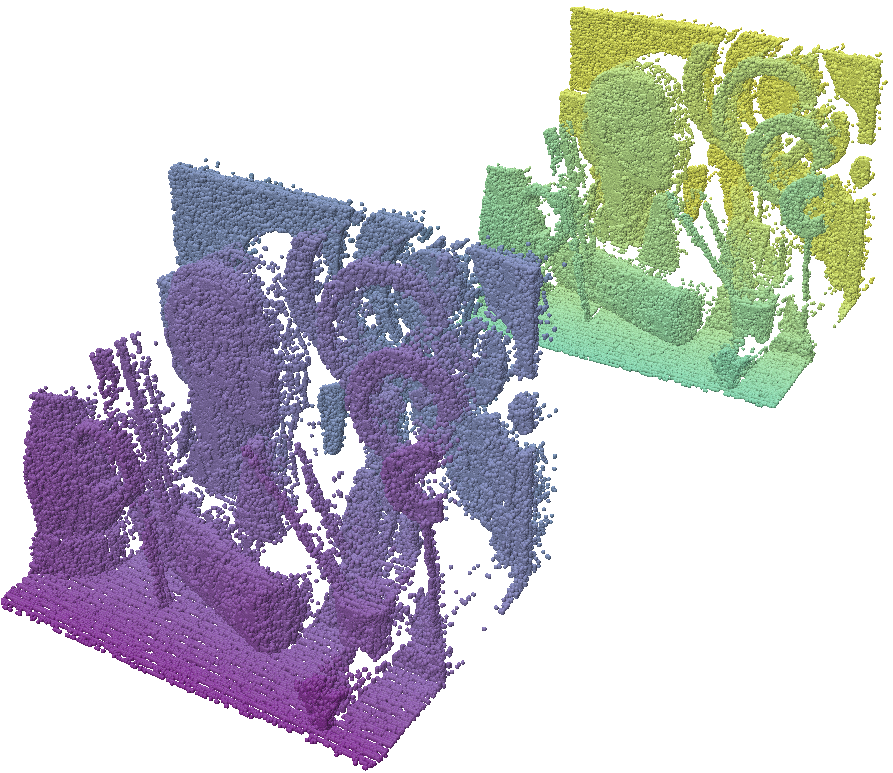} & %Lindell
    \includegraphics[totalheight=\fh]{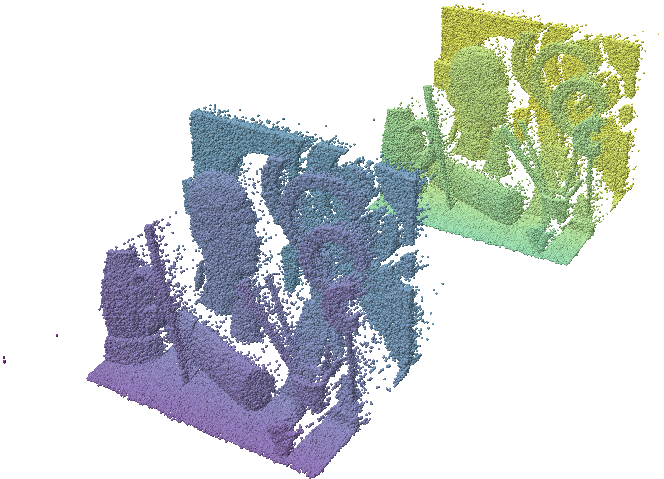} & %peng
    \includegraphics[totalheight=\fh]{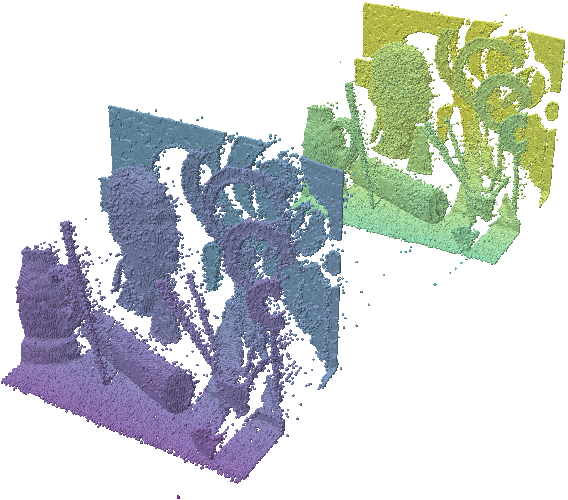} & %BU3D
    \includegraphics[totalheight=\fh] {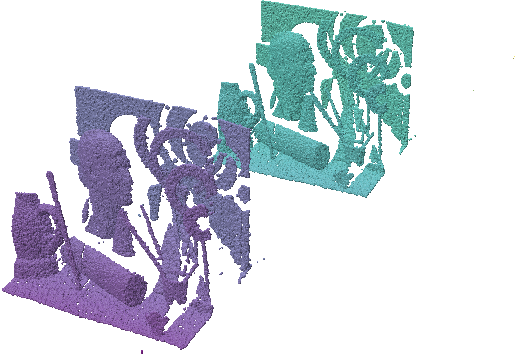}\\[-15pt]

    \rotatebox[origin=l]{90}{\small\parbox{4cm}{$\,$PPP $=\pppb$, SBR$=\sbrb$}} &
    \includegraphics[totalheight=\fh]{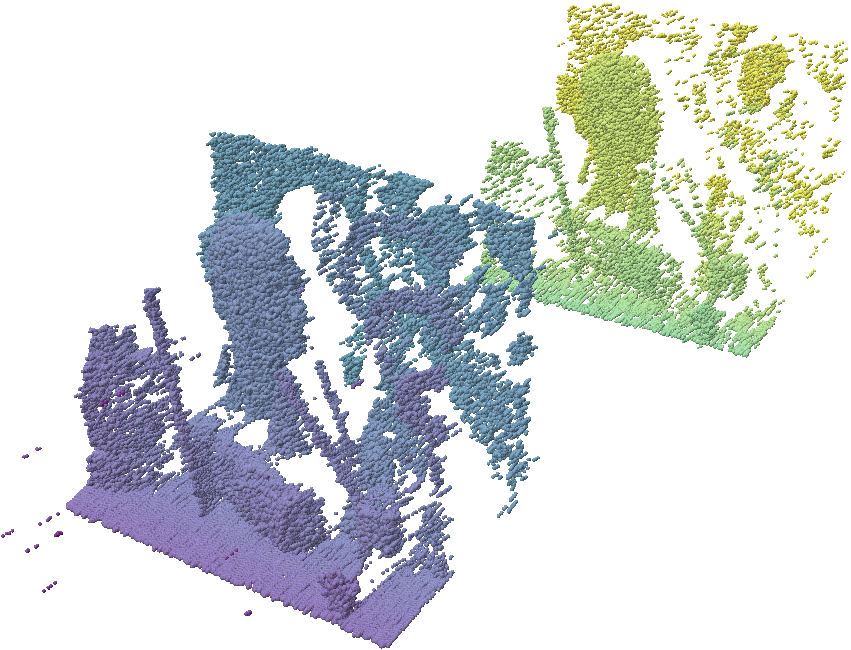} & %RT3D
    \includegraphics[totalheight=\fh]{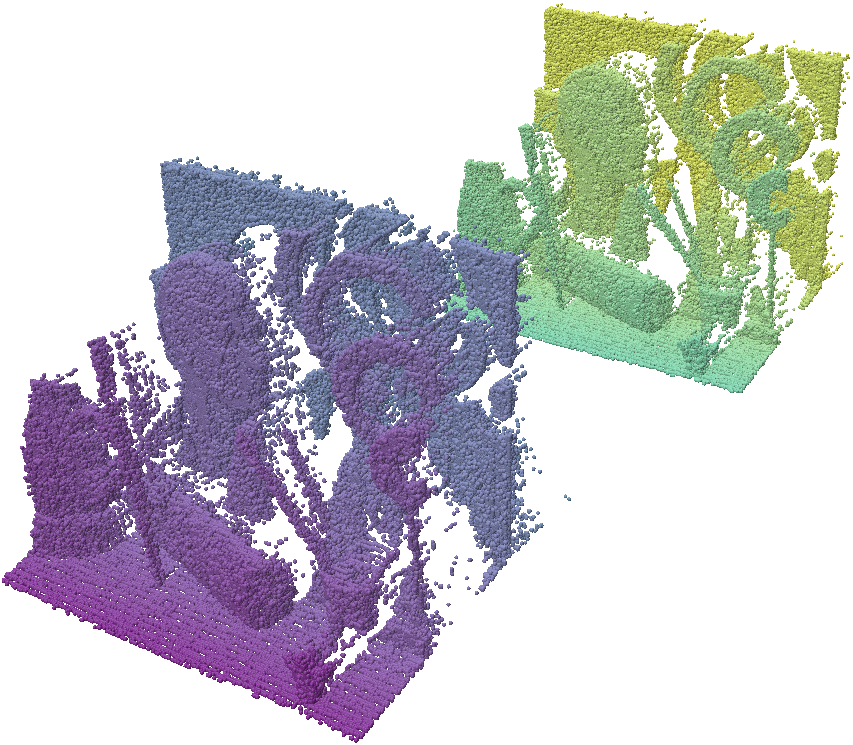} & %Lindell
    \includegraphics[totalheight=\fh]{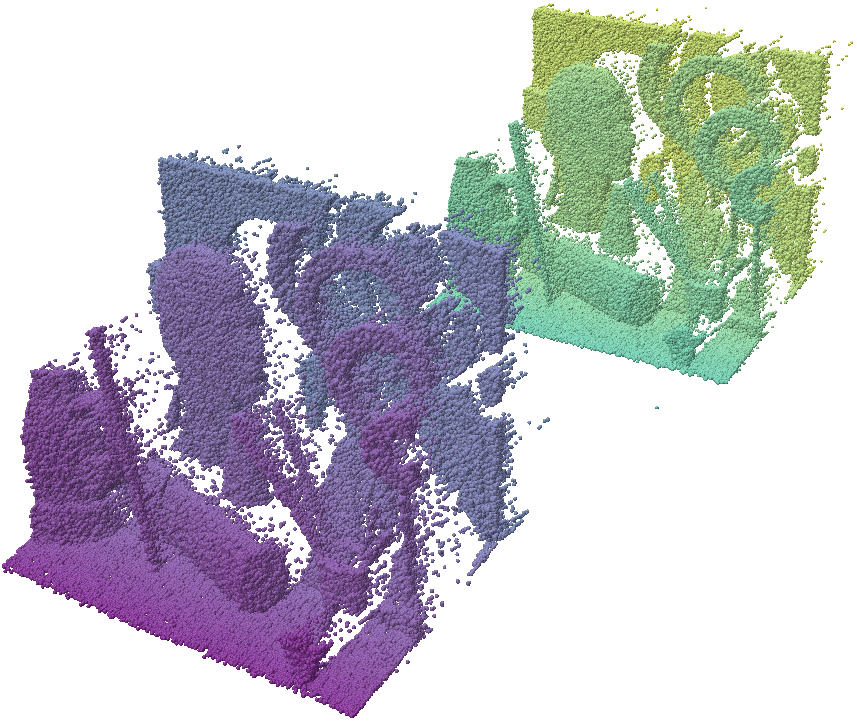} & %peng
    \includegraphics[totalheight=\fh]{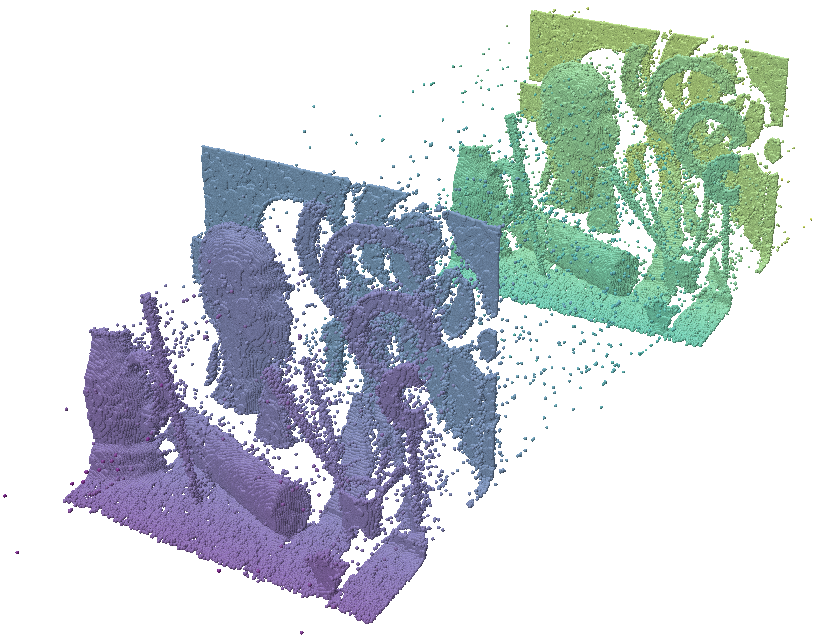} & %BU3D
    \includegraphics[totalheight=\fh] {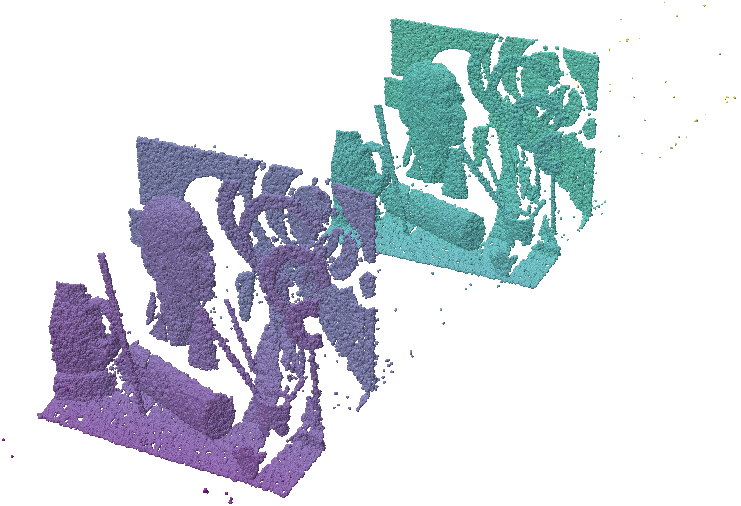}\\%noise 
    
    & {RT3D~\cite{tachella2019realtime}} & Lindell~\cite{lindell2018singlephoton} (2 runs) & Peng~\cite{peng2020photonefficient} (2 runs) & BU3D~\cite{koo2022bayesiana} (2 runs) & Proposed
    \end{tabular}
    \caption{Reconstructed point clouds on the Art scene with dual peaks. The first column shows the results by RT3D. The next three columns show the results by Lindell, Peng and BU3D, respectively, with two runs, assuming that the approximate positions of surfaces are known. The last column shows the results by the proposed method.}
    \label{fig:pc_art}
\end{figure*}
%
% ################
% Art Images  14
\begin{figure*}[ph]
    \centering
    \def\fh{60pt}
    \def\ppp{4.0} \def\sbr{4.0}
    \def\pppc{4.0} \def\sbrc{1.0}
    \def\scene{Art}

    \begin{tabular}{c@{\hspace{2pt}}c@{\hspace{2pt}}c@{\hspace{2pt}}c@{\hspace{2pt}}c@{\hspace{2pt}}c@{\hspace{2pt}}c@{\hspace{2pt}}c@{\hspace{2pt}}c}
    
    %\multicolumn{8}{c}{\centering{PPP = \ppp, SBR = \sbr}}\\[7pt]
    % \rotatebox[origin=r]{90}{\small\parbox{2cm}{SBR=1}} &
    \rotatebox[origin=l]{90}{\small\parbox{2cm}{PPP=4,SBR=4\\First peak}} &
    \includegraphics[totalheight=\fh]{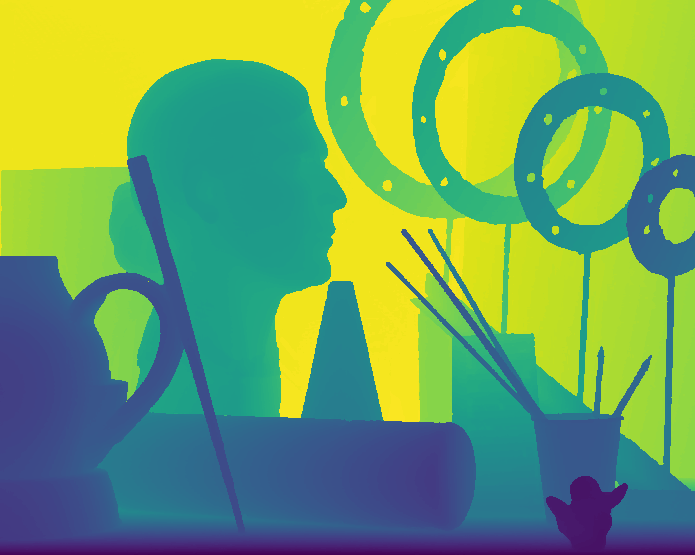} &
    \includegraphics[totalheight=\fh]{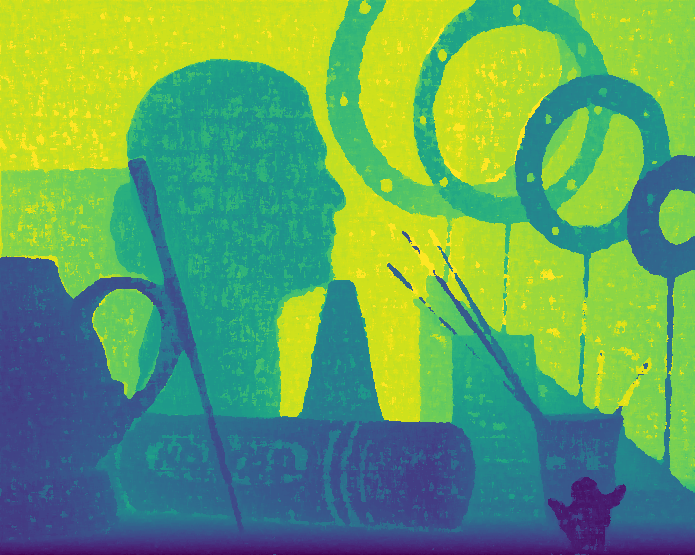} &
    \includegraphics[totalheight=\fh]{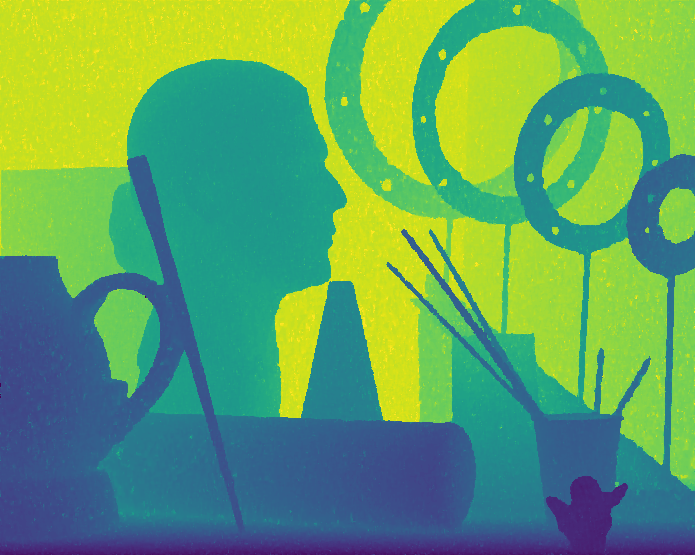} &
    \includegraphics[totalheight=\fh]{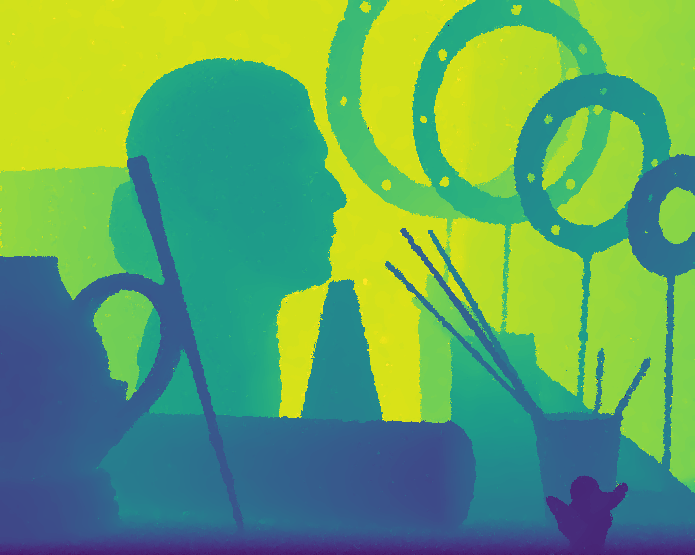} &
    \includegraphics[totalheight=\fh]{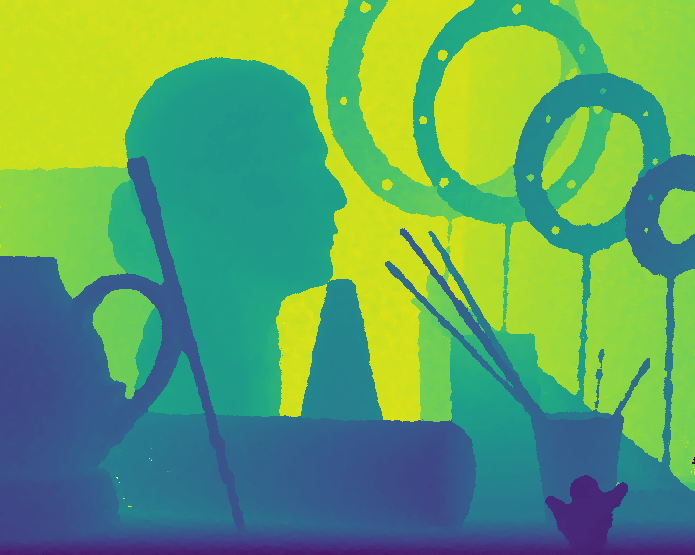} &
    \includegraphics[totalheight=\fh]{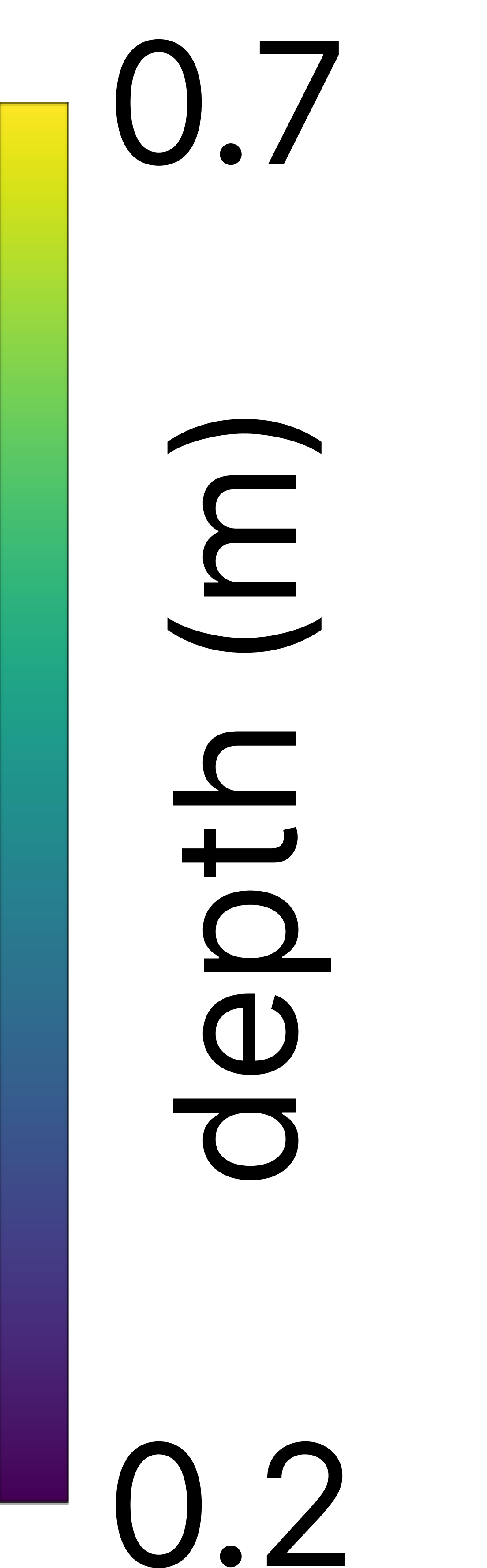} & 
    \includegraphics[totalheight=\fh]{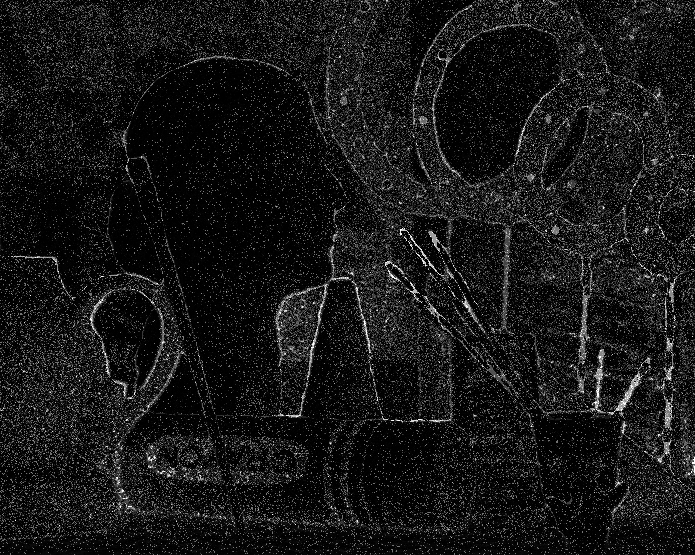} &
    \includegraphics[totalheight=\fh]{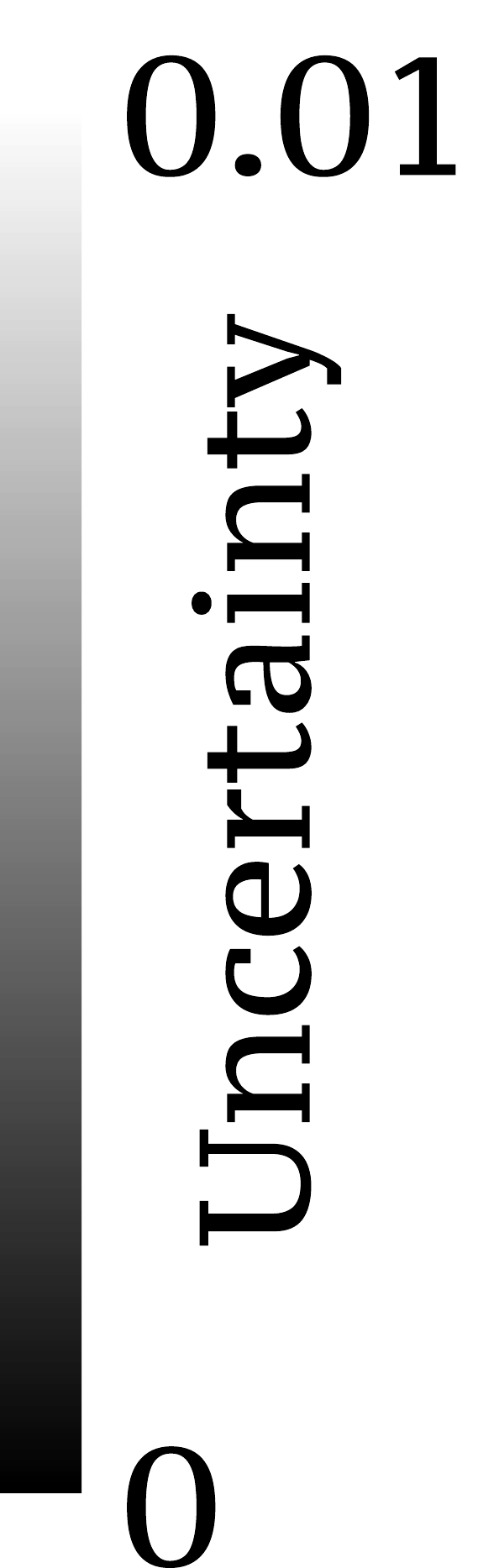} \\%[5pt]
    
    \rotatebox[origin=l]{90}{\small\parbox{2cm}{PPP=4,SBR=4\\Second peak}} &
    \includegraphics[totalheight=\fh]{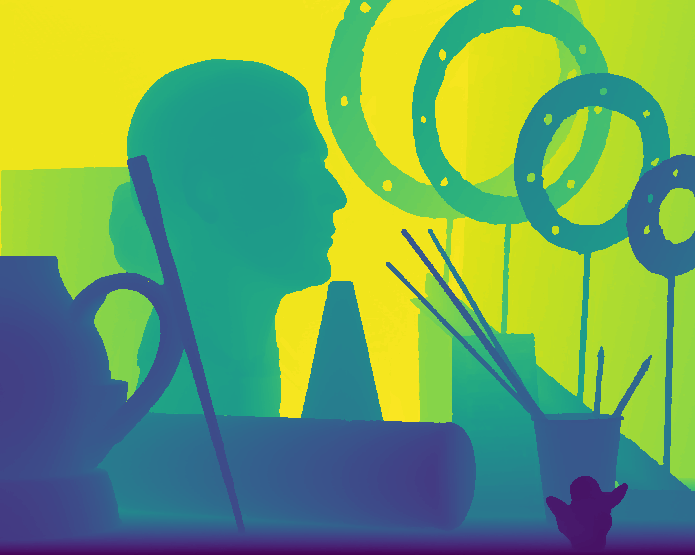} &
    \includegraphics[totalheight=\fh]{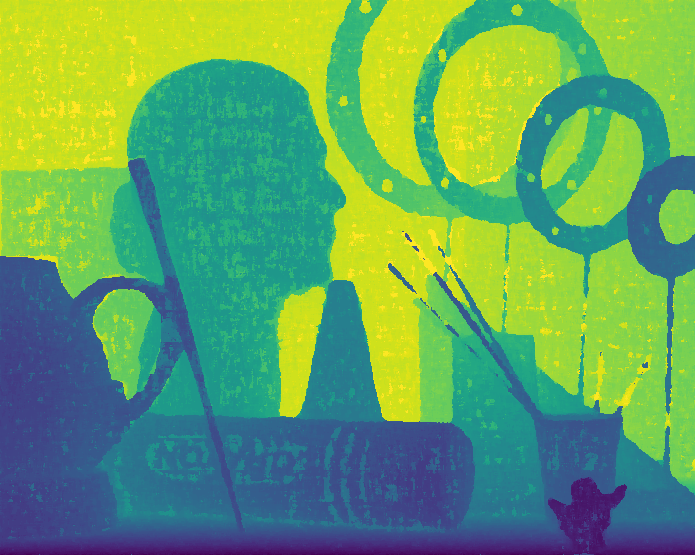} &
    \includegraphics[totalheight=\fh]{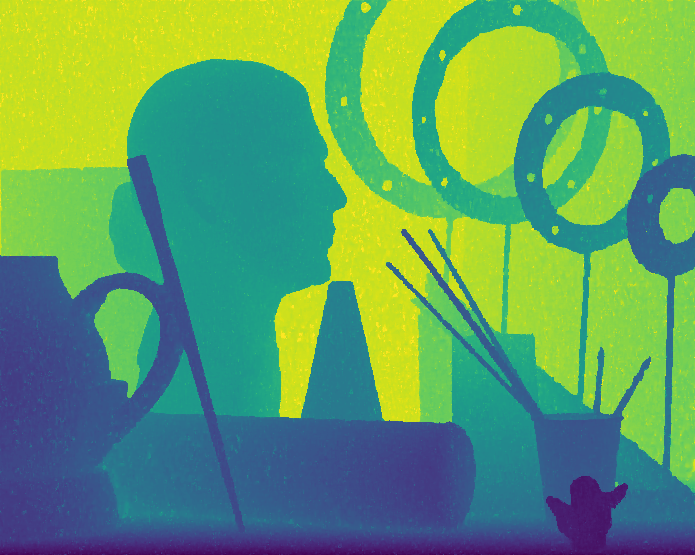} &
    \includegraphics[totalheight=\fh]{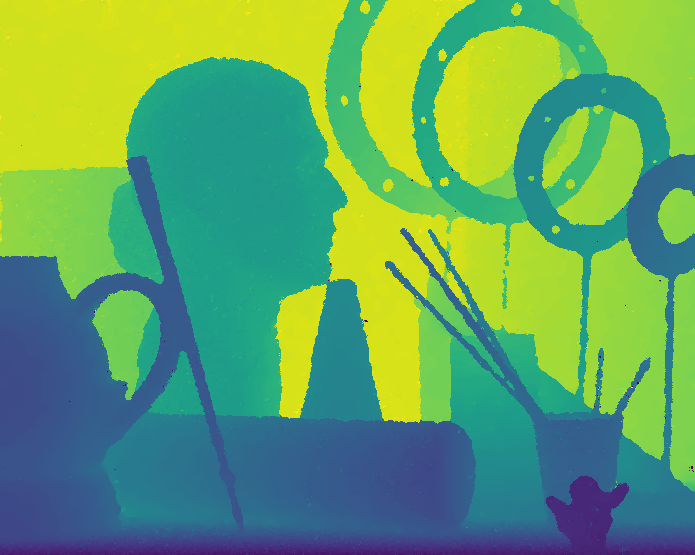} &
    \includegraphics[totalheight=\fh]{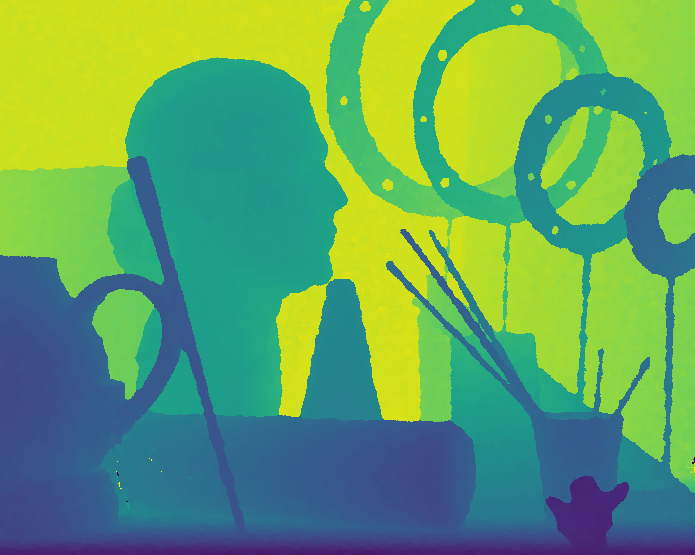} &
    \includegraphics[totalheight=\fh]{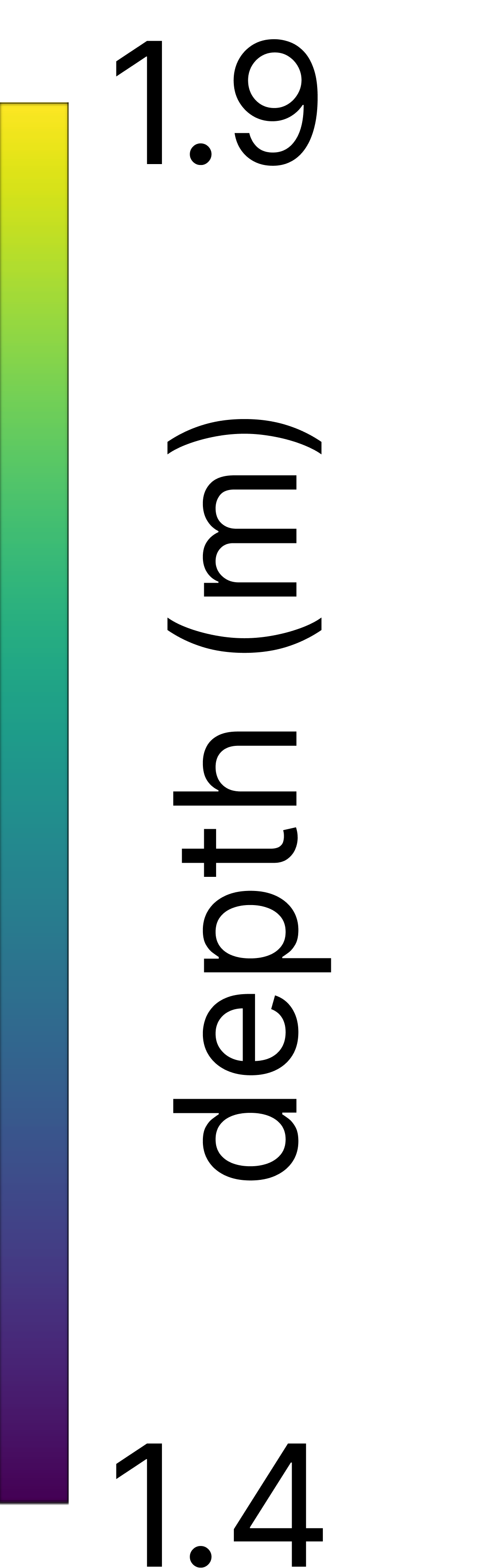} & 
    \includegraphics[totalheight=\fh]{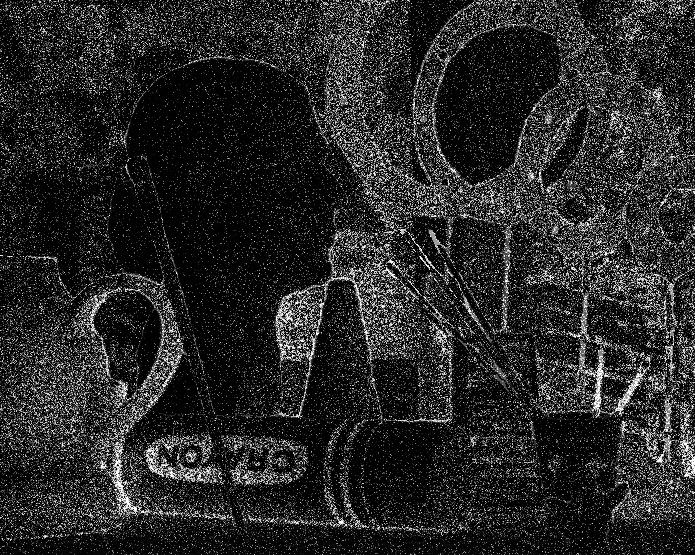}  &
    \includegraphics[totalheight=\fh]{fig/uncertainty/Uncertainty_colorbar.pdf} \\

    \rotatebox[origin=l]{90}{\small\parbox{2cm}{PPP=4,SBR=1\\First peak}} &
    \includegraphics[totalheight=\fh]{fig/img/\scene/Art_gt_img_f.png} &
    \includegraphics[totalheight=\fh]{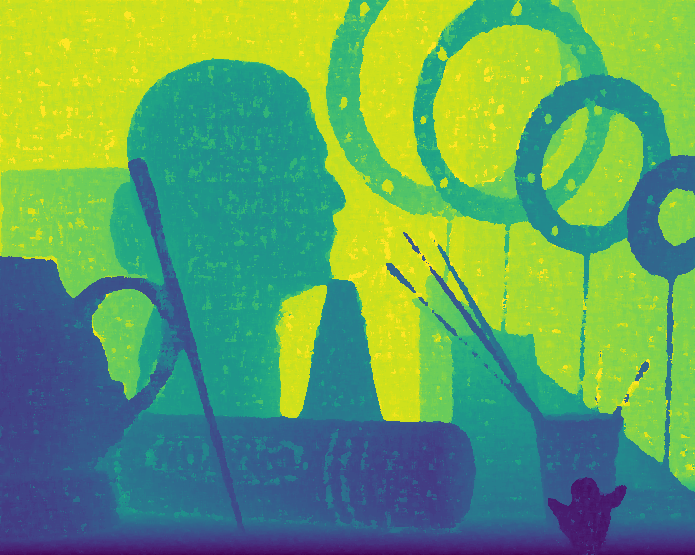} &
    \includegraphics[totalheight=\fh]{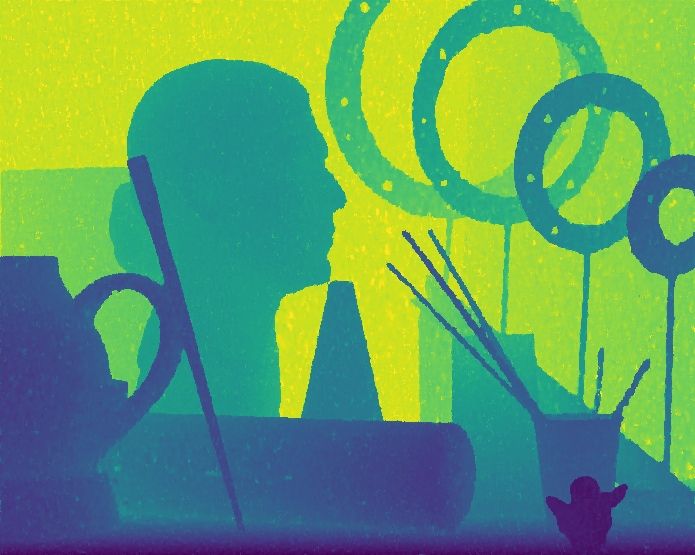} &
    \includegraphics[totalheight=\fh]{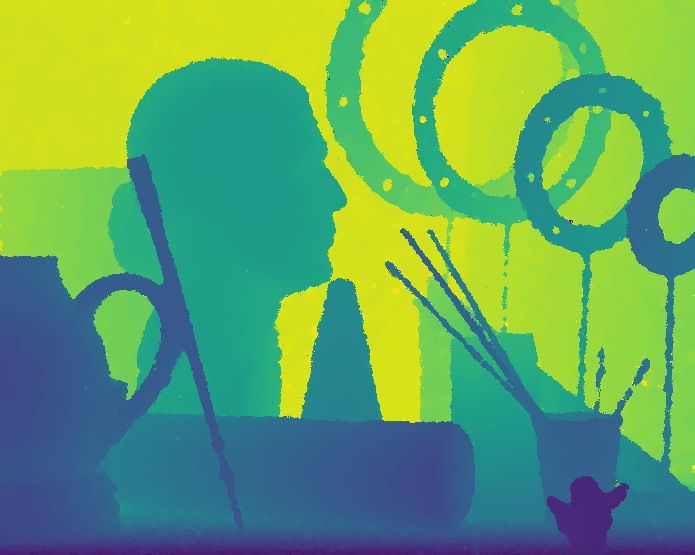} &
    \includegraphics[totalheight=\fh]{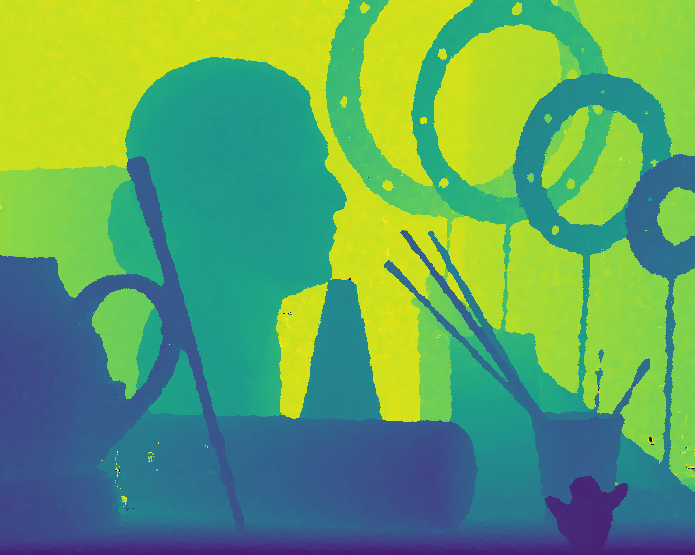} &
    \includegraphics[totalheight=\fh]{fig/img/colorbar_f.png} & 
    \includegraphics[totalheight=\fh]{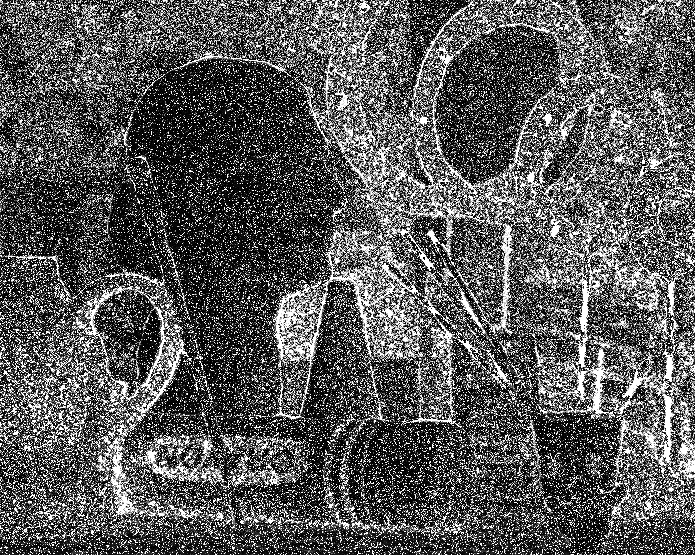} &
    \includegraphics[totalheight=\fh]{fig/uncertainty/Uncertainty_colorbar.pdf} \\

    \rotatebox[origin=l]{90}{\small\parbox{2cm}{PPP=4,SBR=1\\Second peak}} &
    \includegraphics[totalheight=\fh]{fig/img/\scene/Art_gt_img_f.png} &
    \includegraphics[totalheight=\fh]{fig/img/\scene/Lindell_\pppc_\sbrc_f.png} &
    \includegraphics[totalheight=\fh]{fig/img/\scene/Peng_\pppc_\sbrc_f.png} &
    \includegraphics[totalheight=\fh]{fig/img/\scene/BU3D_\pppc_\sbrc_f.png} &
    \includegraphics[totalheight=\fh]{fig/img/\scene/Proposed_\pppc_\sbrc_f.png} &
    \includegraphics[totalheight=\fh]{fig/img/colorbar_f.png} & 
    \includegraphics[totalheight=\fh]{fig/uncertainty/Art_\pppc_\sbrc_uncertainty_f.png} &
    \includegraphics[totalheight=\fh]{fig/uncertainty/Uncertainty_colorbar.pdf} \\
    
    & \small Ground Truth & \small Lindell (2 runs) & \small Peng (2 runs) & \small BU3D (2 runs) & \small Proposed & & Uncertainty
    \end{tabular}
    \caption{Reconstructed depth maps with different PPP and SBR levels on the Art scene. The first column shows the ground truth depth map with first peak (top), and second peak (bottom)}
    \label{fig:img_art}
\end{figure*}

% -------------------------------------
% Point clouds for Bowling1
\begin{figure*}[tp]
    \centering
    \def\fh{75pt} %90 without RT3D
    \def\ppp{16.0} \def\sbr{4.0}
    \def\pppa{4.0} \def\sbra{4.0}
    \def\pppb{4.0} \def\sbrb{1.0}
    \def\scene{Bowling1}
    
    \begin{tabular}{c@{ }c@{ }c@{ }c@{ }c@{ }c}
    
    \rotatebox[origin=l]{90}{\small\parbox{4cm}{$\,$PPP $=\ppp$, SBR$=\sbr$}} &
    \includegraphics[totalheight=\fh]{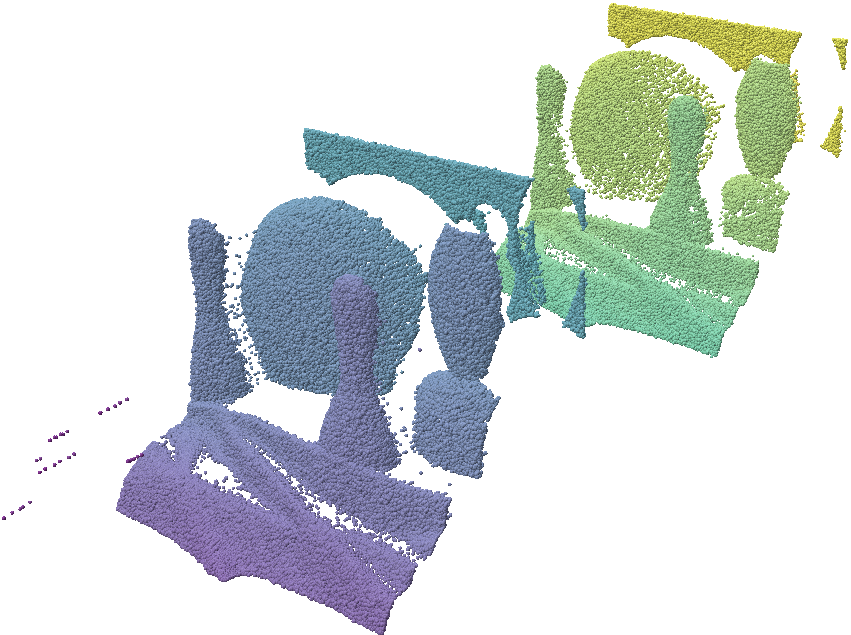} & %RT3D
    \includegraphics[totalheight=\fh]{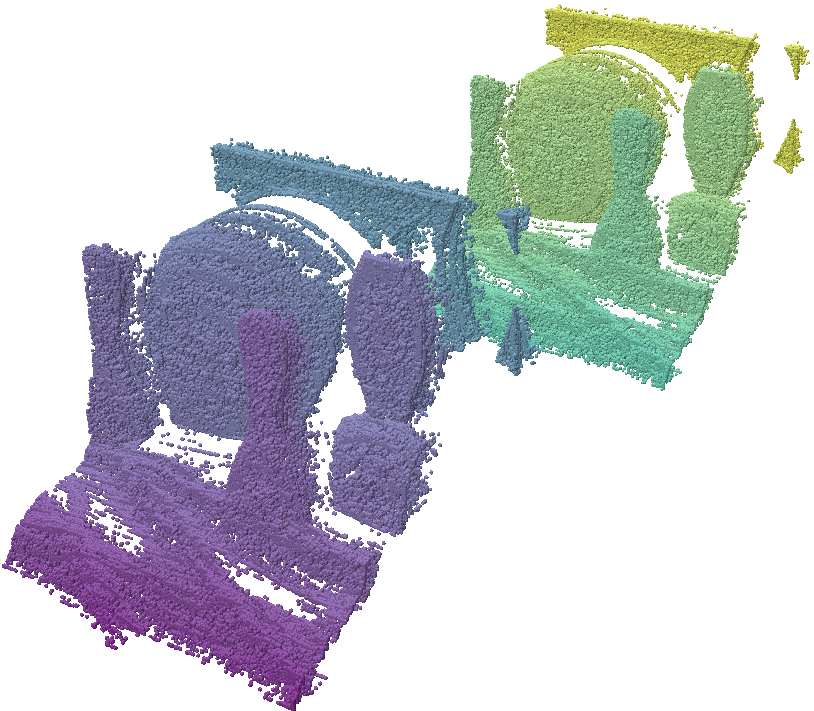} & %lindell
    \includegraphics[totalheight=\fh]{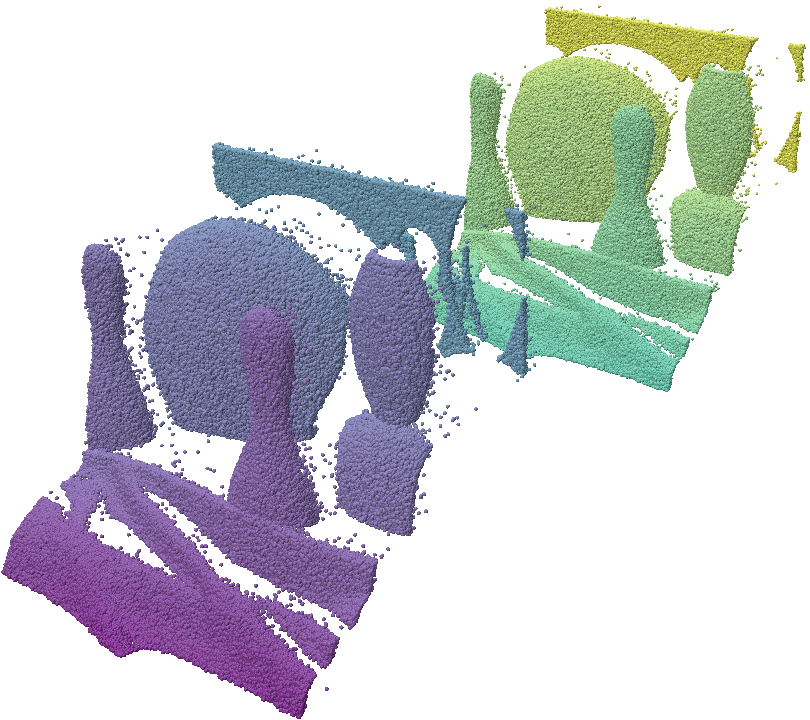} & %Peng
    \includegraphics[totalheight=\fh]{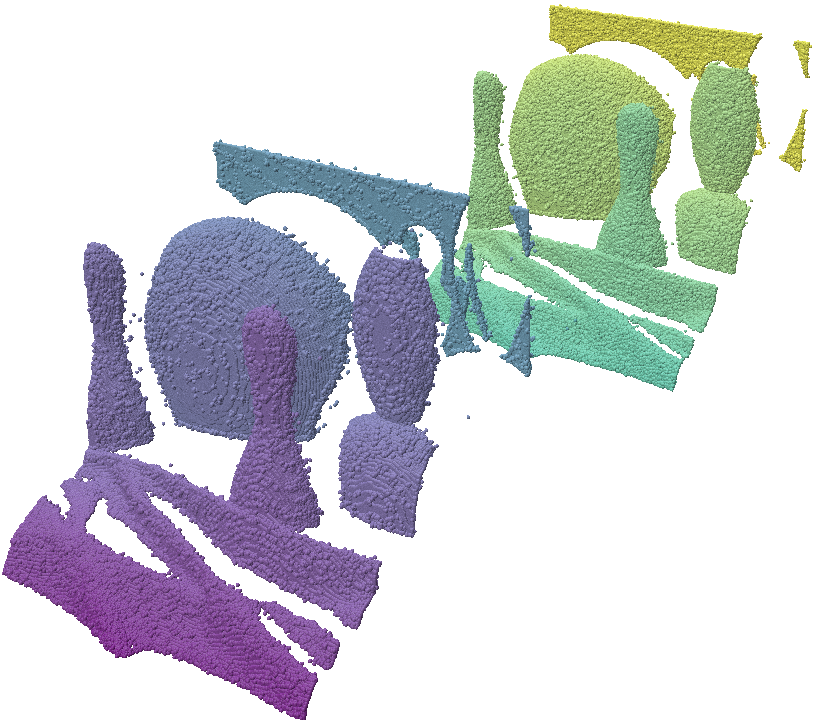} & % BU3D
    \includegraphics[totalheight=\fh]{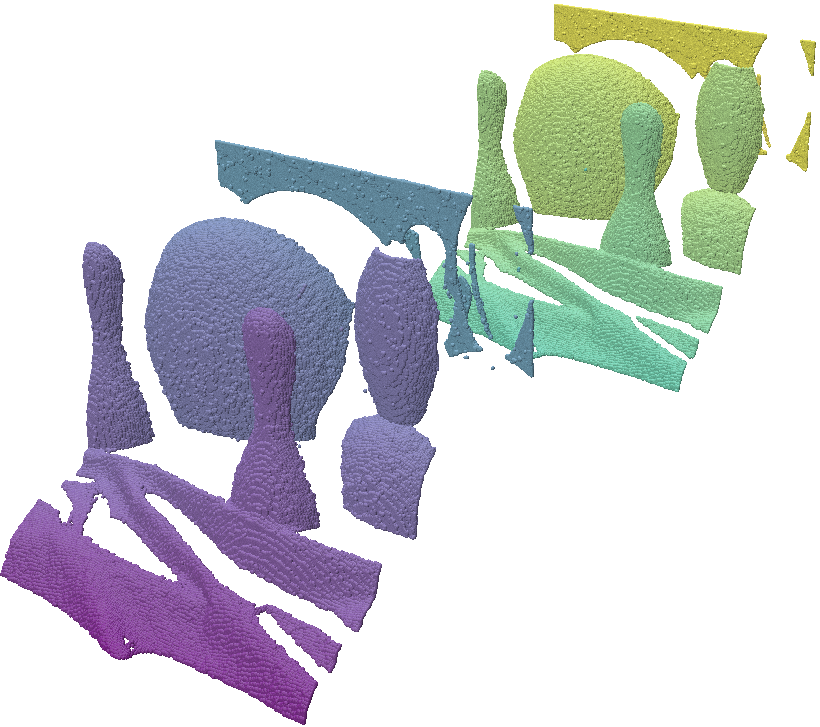} \\[-15pt]
    
    \rotatebox[origin=l]{90}{\small\parbox{4cm}{$\,$PPP $=\pppa$, SBR$=\sbra$}} &
    \includegraphics[totalheight=\fh]{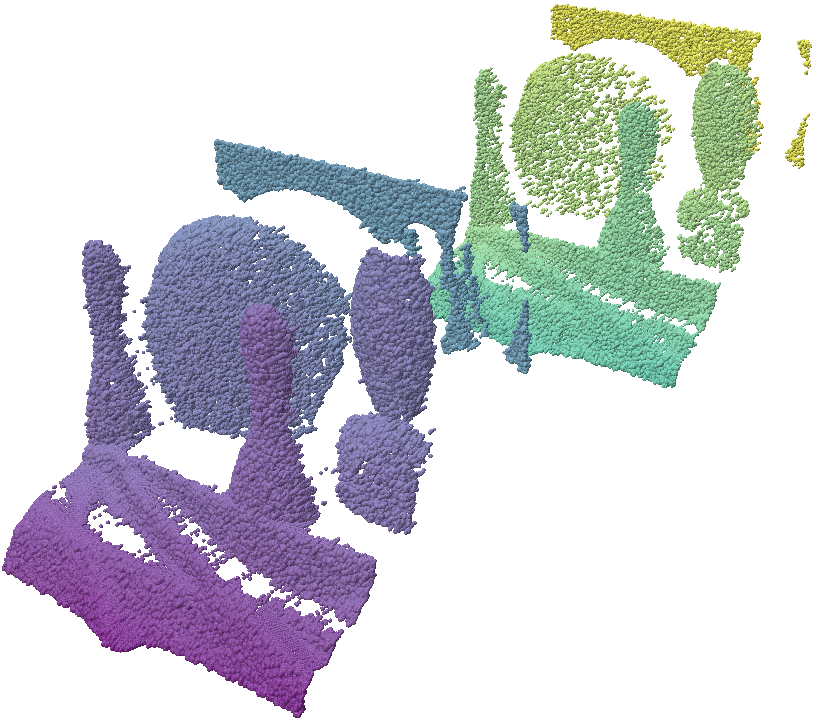} & %RT3D
    \includegraphics[totalheight=\fh]{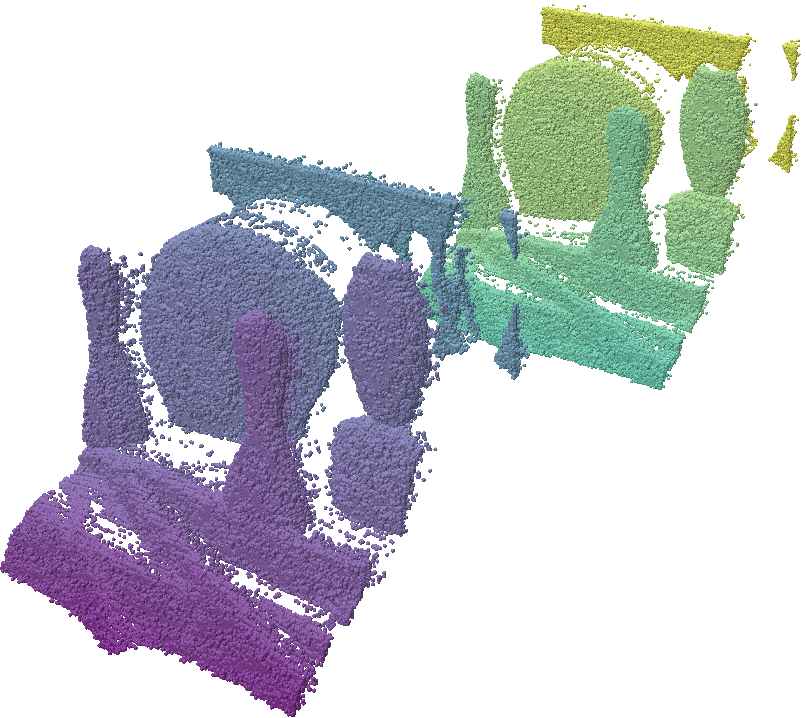} & %Lindell
    \includegraphics[totalheight=\fh]{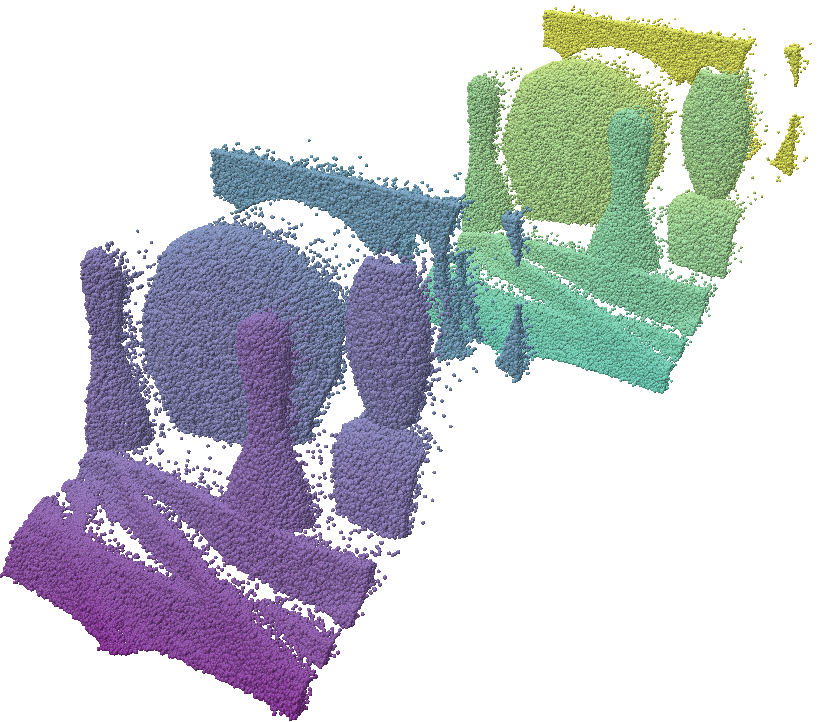} & %peng
    \includegraphics[totalheight=\fh]{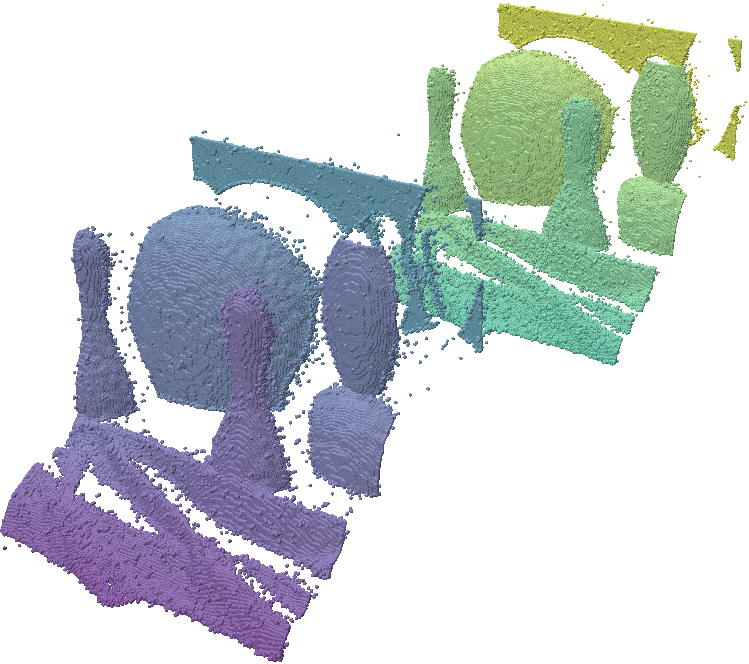} & %BU3D
    \includegraphics[totalheight=\fh] {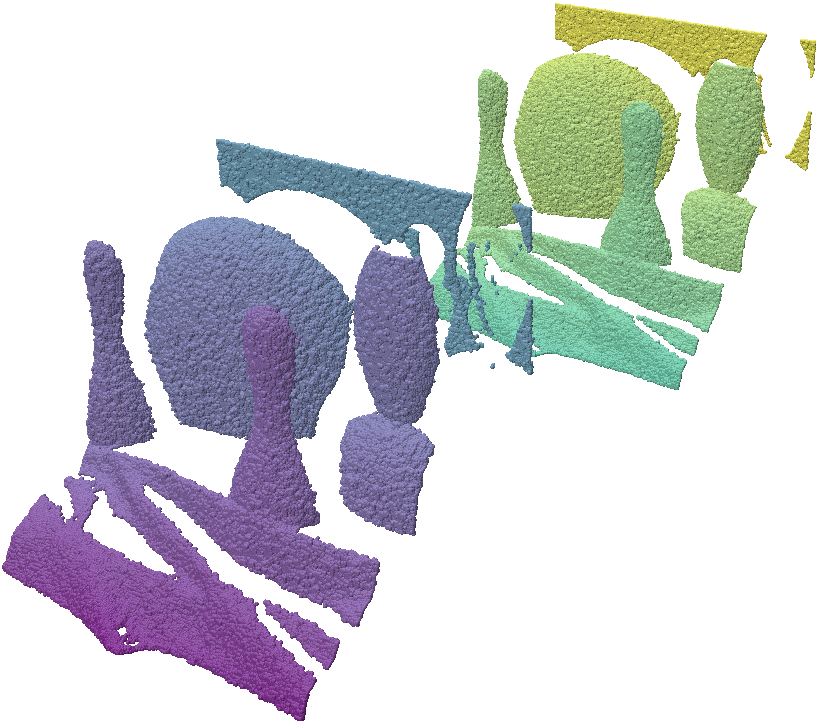}\\[-15pt]

    \rotatebox[origin=l]{90}{\small\parbox{4cm}{$\,$PPP $=\pppb$, SBR$=\sbrb$}} &
    \includegraphics[totalheight=\fh]{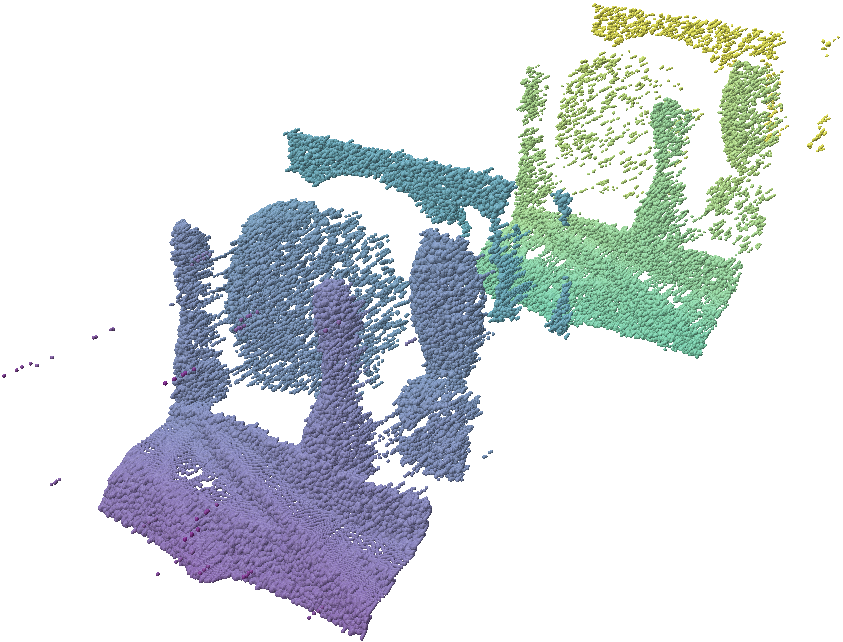} & %RT3D
    \includegraphics[totalheight=\fh]{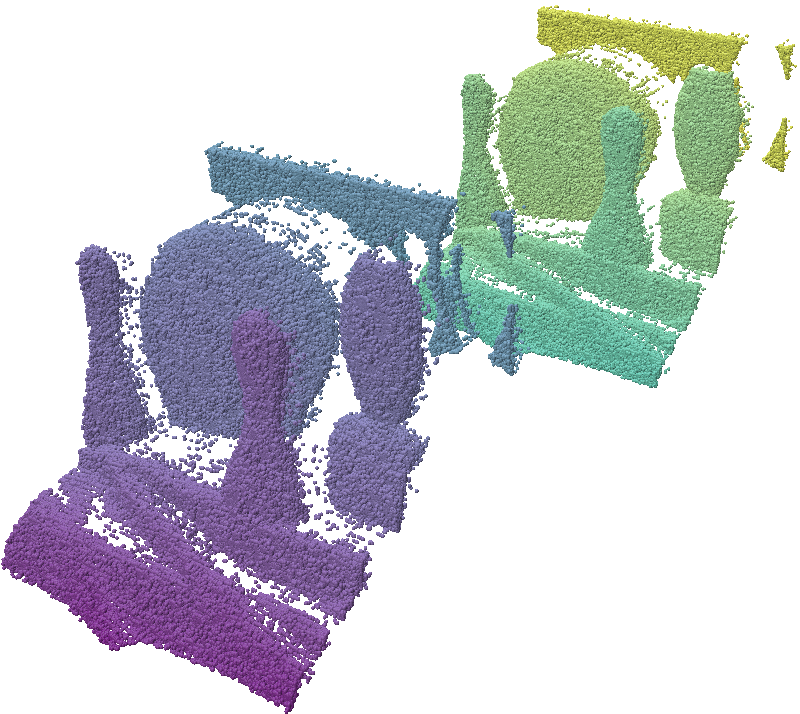} & %Lindell
    \includegraphics[totalheight=\fh]{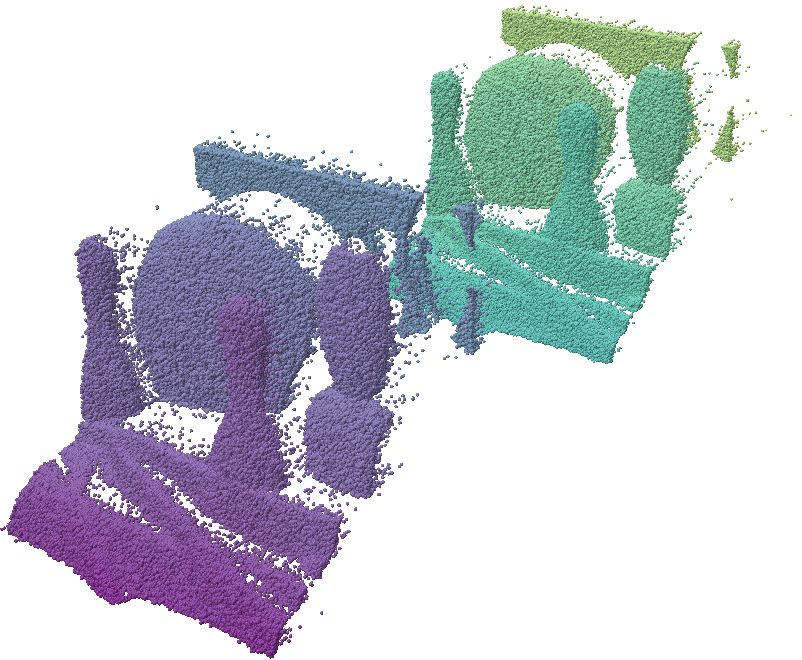} & %peng
    \includegraphics[totalheight=\fh]{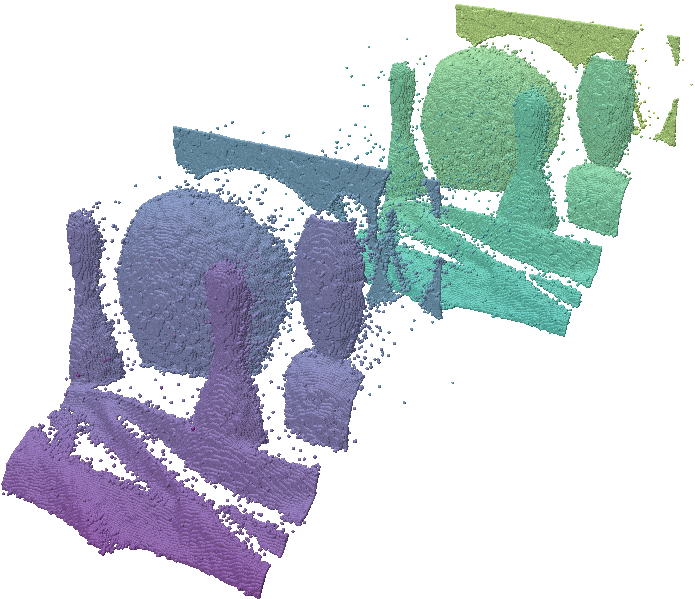} & %BU3D
    \includegraphics[totalheight=\fh] {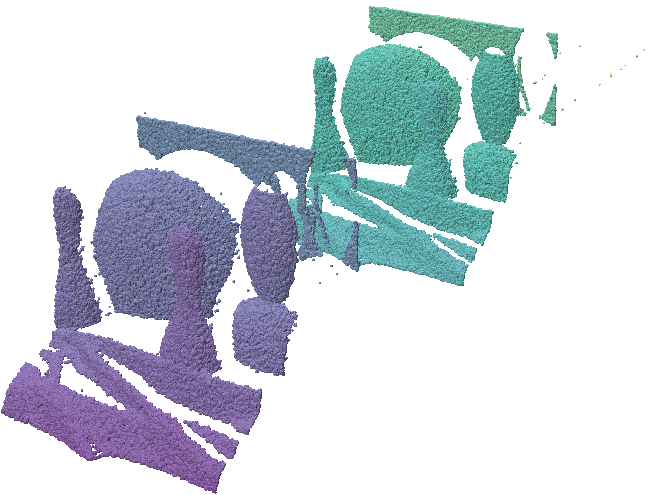}\\%noise 
    
    & {RT3D~\cite{tachella2019realtime}} & Lindell~\cite{lindell2018singlephoton} (2 runs) & Peng~\cite{peng2020photonefficient} (2 runs) & BU3D~\cite{koo2022bayesiana} (2 runs) & Proposed
\end{tabular}
\caption{Reconstructed point clouds on the Bowling scene with dual peaks. The first column shows the results by RT3D. The next three columns show the results by Lindell, Peng and BU3D, respectively, with two runs, assuming that the approximate positions of surfaces are known. The last column shows the results by the proposed method.}
    \label{fig:pc_bowling1}
\end{figure*}
%
% ################
% Bowling1 Images  14
\begin{figure*}[ph]
    \centering
    \def\fh{60pt}
    \def\ppp{4.0} \def\sbr{4.0}
    \def\pppc{4.0} \def\sbrc{1.0}
    \def\scene{Bowling1}

    \begin{tabular}{c@{\hspace{2pt}}c@{\hspace{2pt}}c@{\hspace{2pt}}c@{\hspace{2pt}}c@{\hspace{2pt}}c@{\hspace{2pt}}c@{\hspace{2pt}}c@{\hspace{2pt}}c}
    
    %\multicolumn{8}{c}{\centering{PPP = \ppp, SBR = \sbr}}\\[7pt]
    % \rotatebox[origin=r]{90}{\small\parbox{2cm}{SBR=1}} &
    \rotatebox[origin=l]{90}{\small\parbox{2cm}{PPP=4,SBR=4\\First peak}} &
    \includegraphics[totalheight=\fh]{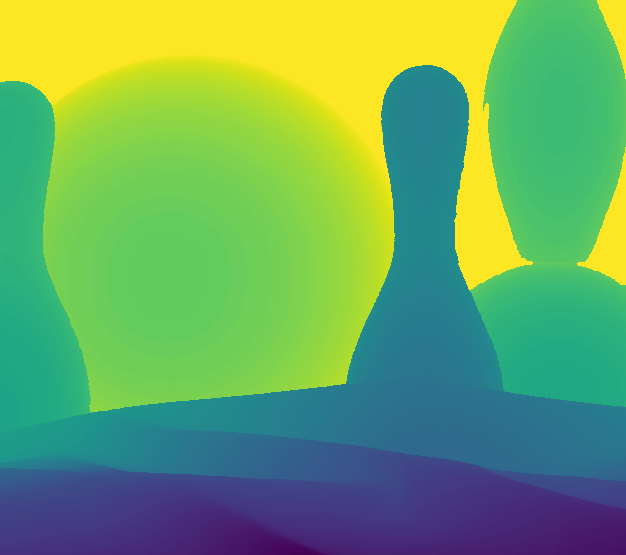} &
    \includegraphics[totalheight=\fh]{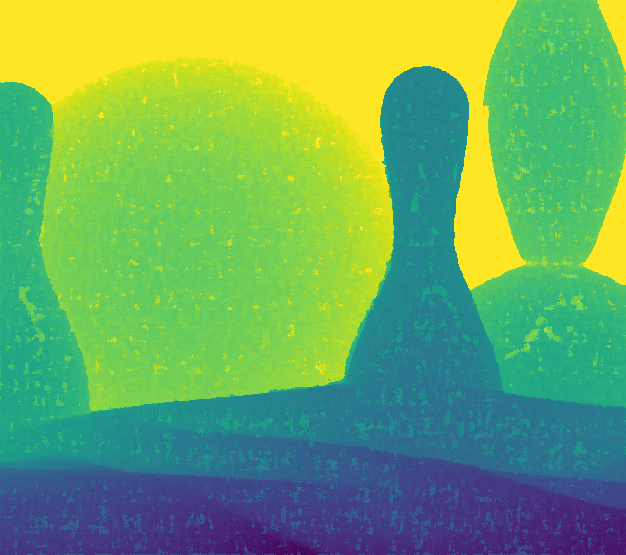} &
    \includegraphics[totalheight=\fh]{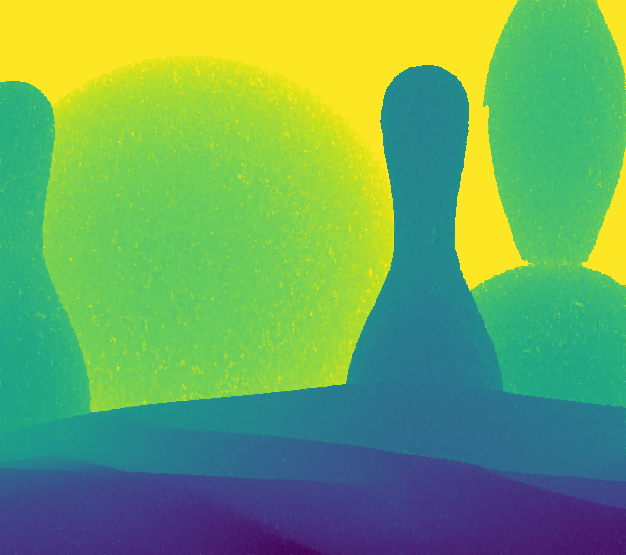} &
    \includegraphics[totalheight=\fh]{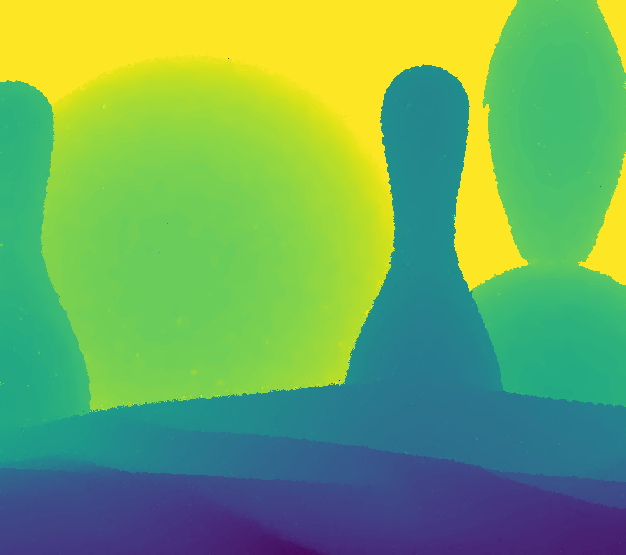} &
    \includegraphics[totalheight=\fh]{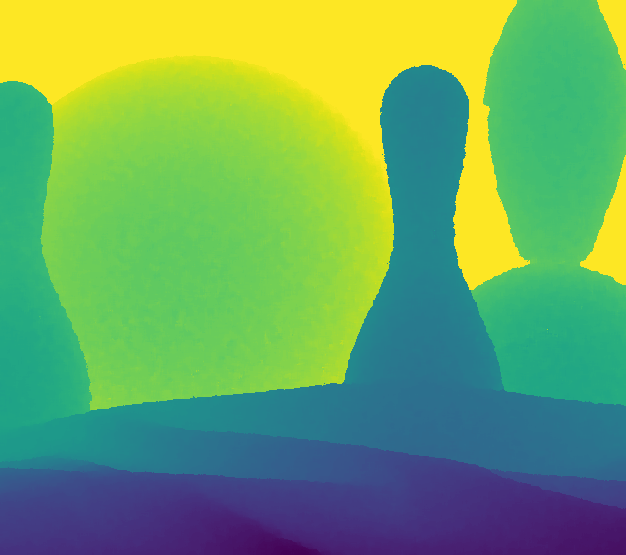} &
    \includegraphics[totalheight=\fh]{fig/img/colorbar_f.png} & 
    \includegraphics[totalheight=\fh]{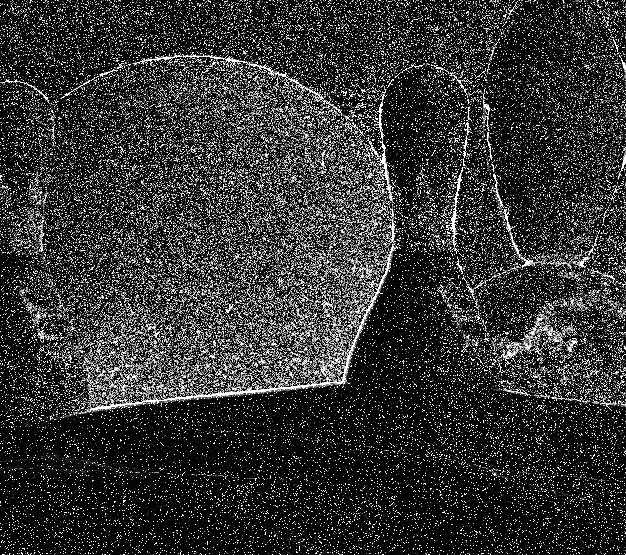} &
    \includegraphics[totalheight=\fh]{fig/uncertainty/Uncertainty_colorbar.pdf} \\%[5pt]
    
    \rotatebox[origin=l]{90}{\small\parbox{2cm}{PPP=4,SBR=4\\Second peak}} &
    \includegraphics[totalheight=\fh]{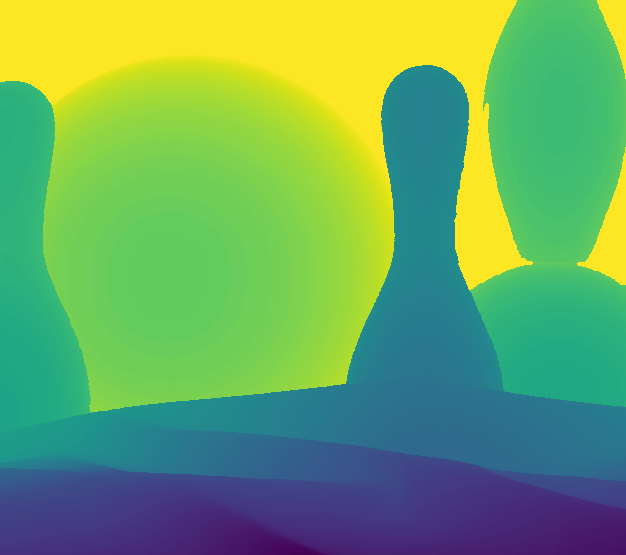} &
    \includegraphics[totalheight=\fh]{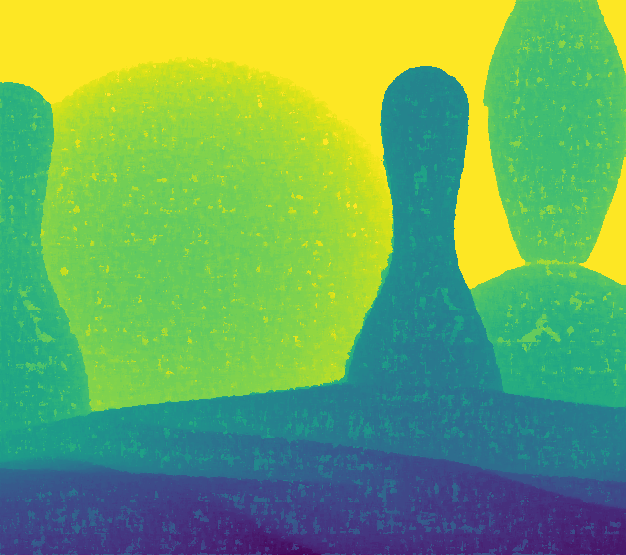} &
    \includegraphics[totalheight=\fh]{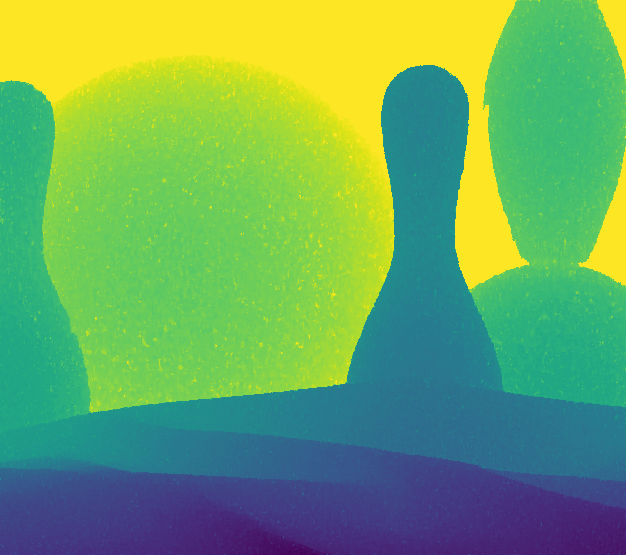} &
    \includegraphics[totalheight=\fh]{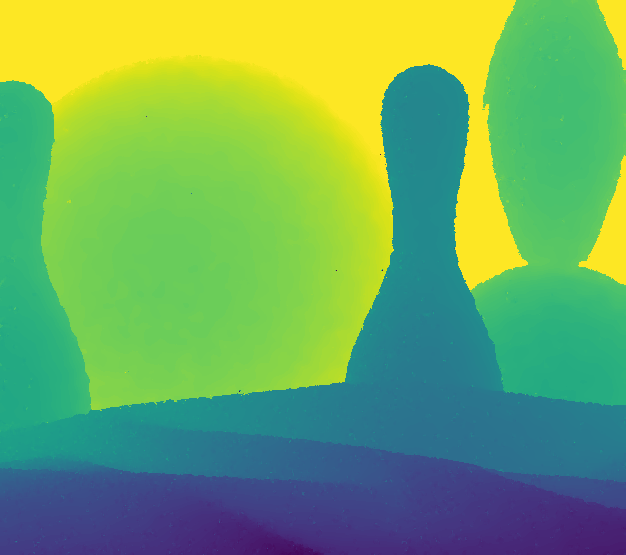} &
    \includegraphics[totalheight=\fh]{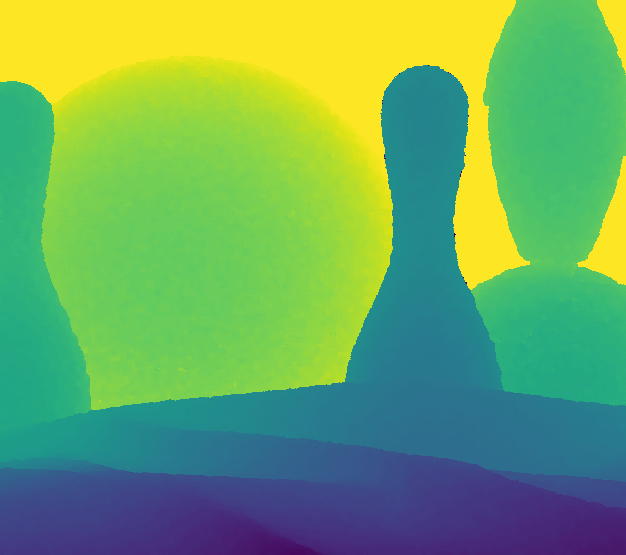} &
    \includegraphics[totalheight=\fh]{fig/img/colorbar_s.png} & 
    \includegraphics[totalheight=\fh]{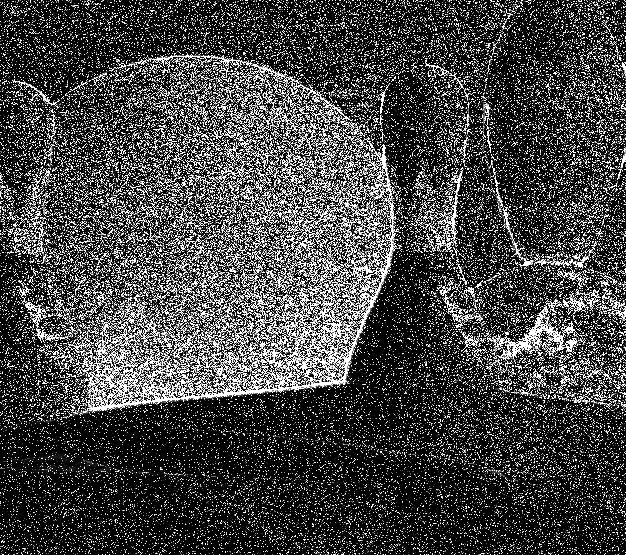}  &
    \includegraphics[totalheight=\fh]{fig/uncertainty/Uncertainty_colorbar.pdf} \\

    \rotatebox[origin=l]{90}{\small\parbox{2cm}{PPP=4,SBR=1\\First peak}} &
    \includegraphics[totalheight=\fh]{fig/img/\scene/\scene_gt_img_f.png} &
    \includegraphics[totalheight=\fh]{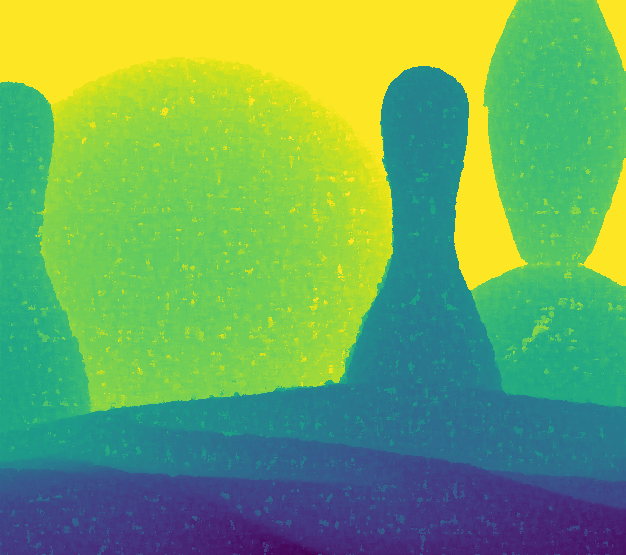} &
    \includegraphics[totalheight=\fh]{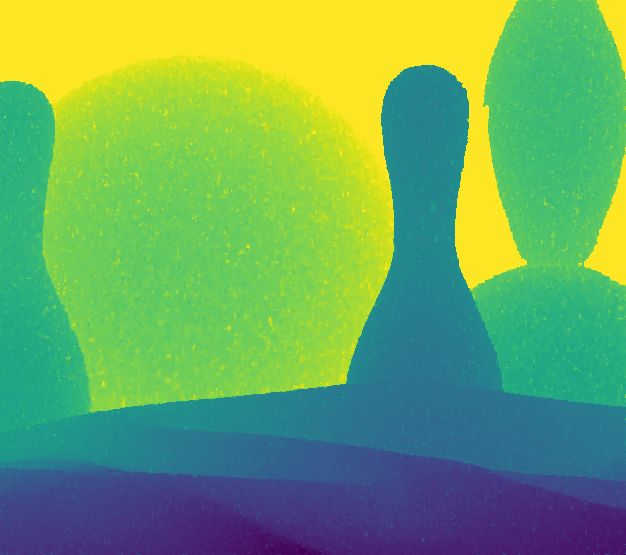} &
    \includegraphics[totalheight=\fh]{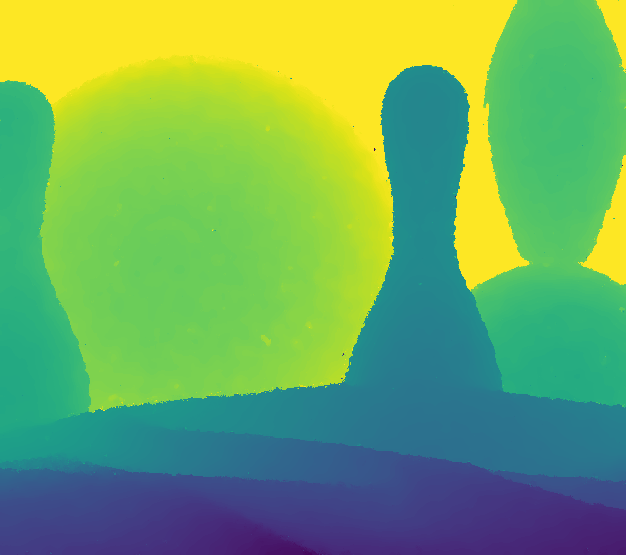} &
    \includegraphics[totalheight=\fh]{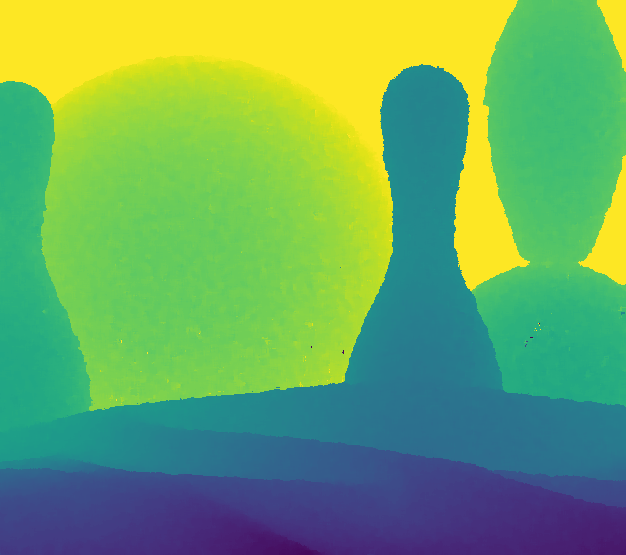} &
    \includegraphics[totalheight=\fh]{fig/img/colorbar_f.png} & 
    \includegraphics[totalheight=\fh]{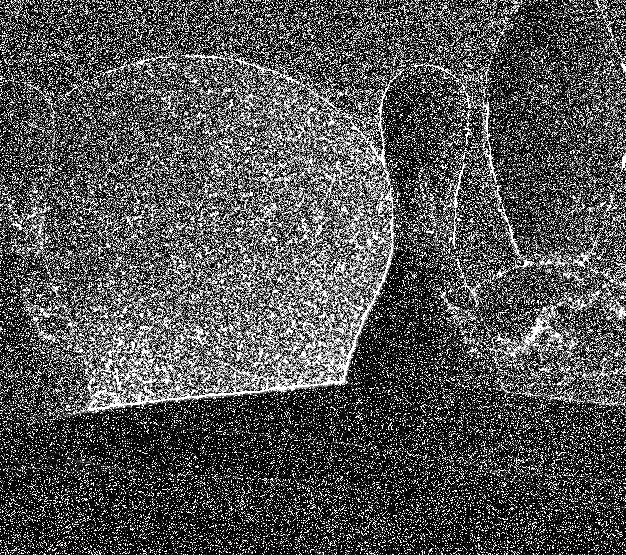} &
    \includegraphics[totalheight=\fh]{fig/uncertainty/Uncertainty_colorbar.pdf} \\

    \rotatebox[origin=l]{90}{\small\parbox{2cm}{PPP=4,SBR=1\\Second peak}} &
    \includegraphics[totalheight=\fh]{fig/img/\scene/\scene_gt_img_f.png} &
    \includegraphics[totalheight=\fh]{fig/img/\scene/Lindell_\pppc_\sbrc_f.png} &
    \includegraphics[totalheight=\fh]{fig/img/\scene/Peng_\pppc_\sbrc_f.png} &
    \includegraphics[totalheight=\fh]{fig/img/\scene/BU3D_\pppc_\sbrc_f.png} &
    \includegraphics[totalheight=\fh]{fig/img/\scene/Proposed_\pppc_\sbrc_f.png} &
    \includegraphics[totalheight=\fh]{fig/img/colorbar_f.png} & 
    \includegraphics[totalheight=\fh]{fig/uncertainty/\scene_\pppc_\sbrc_uncertainty_f.png} &
    \includegraphics[totalheight=\fh]{fig/uncertainty/Uncertainty_colorbar.pdf} \\
    
    & \small Ground Truth & \small Lindell (2 runs) & \small Peng (2 runs) & \small BU3D (2 runs) & \small Proposed & & Uncertainty
    \end{tabular}
    \caption{Reconstructed depth maps with different PPP and SBR levels on the Bowling scene. The first column shows the ground truth depth map with first peak (top), and second peak (bottom)}
    \label{fig:img_bowling1}
\end{figure*}

%
% ################

% ------------------------------
% Table Art Notation
\begin{table*}[t]
    \centering
    \caption{Quantitative comparison on the Art scene. Chamfer distance consists of two terms: O-To-G (distance from the estimated point cloud to the ground truth) and G-To-O (distance from the ground truth to the estimated point cloud). DAE, O-To-G, and G-To-O values are reported as mean $\pm$ standard deviation of pixelwise measures. All values are scaled by $10^{-2}$ for readibility.}

    \resizebox{\textwidth}{!}{%
    % \resizebox{\linewidth}{!}{%
    \begin{tabular}{l cccc c cccc c cccc}
    \hline
    \noalign{\vspace{2pt}}
     & \multicolumn{4}{c}{SBR = 1} & & \multicolumn{4}{c}{SBR = 4} & & \multicolumn{4}{c}{SBR = 16} \\[2pt]
    \cline{2-5} \cline{7-10} \cline{12-15} 
    \noalign{\vspace{2pt}}
     & DAE & Chamfer & O-To-G & G-To-O & & DAE & Chamfer & O-To-G & G-To-O & & DAE & Chamfer & O-To-G & G-To-O \\[2pt]
    \hline
    \noalign{\vspace{2pt}}
    
    \multicolumn{6}{l}{PPP = 1} \\
    RT3D & & $1.92$ & $0.88\pm0.91$ & $1.04\pm0.90$ & & & $1.17$ & $0.51\pm0.52$ & $0.66\pm0.61$ & & & $1.08$ & $0.54\pm0.51$ & $0.54\pm0.45$ \\
    Lindell (2 runs)  & $2.61\pm12.42$ & $1.55$ & $1.28\pm3.49 $ & $0.27\pm0.21$ & & $1.31\pm4.21$ & $0.98$ & $0.75\pm1.30 $ & $0.23\pm0.19$ & & $1.22\pm3.75$ & $0.91$ & $0.69\pm1.13 $ & $0.22\pm0.19$ \\
    Peng (2 runs) & $\textbf{0.99}\pm2.32$ & $\textbf{0.85}$ & $\textbf{0.66}\pm0.91 $ & $0.20\pm0.17$ & & $\textbf{0.82}\pm1.93$ & $0.74$ & $0.57\pm0.82 $ & $0.17\pm0.14$ & & $0.78\pm1.85$ & $0.71$ & $0.55\pm0.80 $ & $0.16\pm0.13$ \\
    BU3D (2 runs) & $2.74\pm14.83$ & $1.22$ & $0.88\pm3.72 $ & $0.35\pm0.25$ & & $1.16\pm6.91$ & $0.80$ & $0.46\pm1.59 $ & $0.33\pm0.23$ & & $0.98\pm5.49$ & $0.75$ & $0.41\pm1.15 $ & $0.33\pm0.23$ \\
    Proposed & $2.73\pm12.52$ & $1.92$ & $1.73\pm9.71$ & $\textbf{0.18}\pm0.18$ & & $0.93\pm4.76$ & $\textbf{0.58}$ & $\textbf{0.44}\pm3.04$ & $\textbf{0.15}\pm0.16$ & & $\textbf{0.74}\pm3.07$ & $\textbf{0.43}$ & $\textbf{0.29}\pm1.12$ & $\textbf{0.14}\pm0.15$  \\[2pt]
    \hline
    \noalign{\vspace{2pt}}
    \multicolumn{6}{l}{PPP = 4} \\
    RT3D & & $1.01$ & $0.40\pm0.53$ & $0.61\pm0.67$ & & & $0.60$ & $0.25\pm0.24$ & $0.35\pm0.36$ & & & $0.67$ & $0.40\pm0.40$ & $0.26\pm0.20$  \\
    Lindell (2 runs) & $1.20\pm3.07$ & $0.92$ & $0.71\pm1.20 $ & $0.21\pm0.18$ & & $1.40\pm3.29$ & $1.13$ & $0.89\pm1.46 $ & $0.24\pm0.20$ & & $1.53\pm3.41$ & $1.27$ & $1.01\pm1.57 $ & $0.26\pm0.23$ \\
    Peng (2 runs) & $0.63\pm1.50$ & $0.61$ & $0.47\pm0.72 $ & $0.14\pm0.11$ & & $0.53\pm1.23$ & $0.54$ & $0.41\pm0.62 $ & $0.13\pm0.11$ & & $0.49\pm1.16$ & $0.52$ & $0.39\pm0.56 $ & $0.13\pm0.10$ \\
    BU3D (2 runs) & $0.86\pm4.08$ & $0.74$ & $0.42\pm1.44 $ & $0.31\pm0.18$ & & $0.64\pm2.07$ & $0.64$ & $0.35\pm0.44 $ & $0.29\pm0.17$ & & $0.61\pm1.93$ & $0.62$ & $0.34\pm0.37 $ & $0.27\pm0.17$ \\
    Proposed & $\textbf{0.58}\pm3.11$ & $\textbf{0.30}$ & $\textbf{0.20}\pm1.71$ & $\textbf{0.10}\pm0.13$ & & $\textbf{0.47}\pm2.44$ & $\textbf{0.20}$ & $\textbf{0.12}\pm0.51$ & $\textbf{0.08}\pm0.13$ & & $\textbf{0.45}\pm2.28$ & $\textbf{0.18}$ & $\textbf{0.11}\pm0.15$ & $\textbf{0.07}\pm0.12$ \\[2pt]
    
    \hline
    \noalign{\vspace{2pt}}
    \multicolumn{6}{l}{PPP = 16} \\
    RT3D & & $0.71$ & $0.35\pm1.02$ & $0.36\pm0.34$ & & & $0.53$ & $0.29\pm0.32$ & $0.24\pm0.19$ & & & $0.50$ & $0.29\pm0.25$ & $0.22\pm0.15$ \\
    Lindell (2 runs) & $2.68\pm3.94$ & $2.51$ & $2.00\pm2.09 $ & $0.51\pm0.53$ & & $2.99\pm3.86$ & $2.97$ & $2.28\pm1.98 $ & $0.69\pm0.67$ & & $3.05\pm3.79$ & $3.10$ & $2.33\pm1.90 $ & $0.76\pm0.72$ \\
    Peng (2 runs) & $\textbf{0.39}\pm0.97$ & $0.42$ & $0.30\pm0.43 $ & $0.12\pm0.10$ & & $\textbf{0.33}\pm0.78$ & $0.38$ & $0.26\pm0.33 $ & $0.11\pm0.10$ & & $\textbf{0.31}\pm0.72$ & $0.37$ & $0.25\pm0.30 $ & $0.11\pm0.10$ \\
    BU3D (2 runs) & $0.54\pm2.34$ & $0.56$ & $0.33\pm0.80 $ & $0.23\pm0.16$ & & $0.44\pm1.44$ & $0.47$ & $0.28\pm0.29 $ & $0.20\pm0.15$ & & $0.42\pm1.35$ & $0.45$ & $0.26\pm0.23 $ & $0.18\pm0.14$ \\
    Proposed & $\textbf{0.39}\pm2.28$ & $\textbf{0.12}$ & $\textbf{0.07}\pm0.24$ & $\textbf{0.05}\pm0.11$ & & $0.37\pm2.24$ & $\textbf{0.09}$ & $\textbf{0.05}\pm0.10$ & $\textbf{0.04}\pm0.11$ & & $0.36\pm2.24$ & $\textbf{0.08}$ & $\textbf{0.04}\pm0.09$ & $\textbf{0.04}\pm0.10$ \\[2pt]
    \hline
    \end{tabular}%
    }
    \label{tab:art}
\end{table*}
%
% ------------------------------

% ------------------------------

% table Bowling1 Notation
\begin{table*}[!]
    \centering
    \caption{Quantitative comparison on the Bowling scene. Chamfer distance consists of two terms: O-To-G (distance from the estimated point cloud to the ground truth) and G-To-O (distance from the ground truth to the estimated point cloud). DAE, O-To-G, and G-To-O values are reported as mean $\pm$ standard deviation of pixelwise measures. All values are scaled by $10^{-2}$ for readibility.}

    \resizebox{\textwidth}{!}{%
    % \resizebox{\linewidth}{!}{%
    \begin{tabular}{l cccc c cccc c cccc}
    \hline
    \noalign{\vspace{2pt}}
        & \multicolumn{4}{c}{SBR = 1} & & \multicolumn{4}{c}{SBR = 4} & & \multicolumn{4}{c}{SBR = 16} \\[2pt]
        \cline{2-5} \cline{7-10} \cline{12-15} 
    
    \noalign{\vspace{2pt}}
        & DAE & Chamfer & O-To-G & G-To-O & & DAE & Chamfer & O-To-G & G-To-O & & DAE & Chamfer & O-To-G & G-To-O \\[2pt]
    \hline
    \noalign{\vspace{2pt}}
    
    \multicolumn{6}{l}{PPP = 1} \\
    RT3D & & $1.71$ & $0.70\pm0.68$ & $1.01\pm0.89$ & & & $1.02$ & $0.42\pm0.44$ & $0.60\pm0.52$ & & & $1.07$ & $0.57\pm0.53$ & $0.51\pm0.39$ \\
    Lindell (2 runs) & $1.83\pm9.44$ & $1.44$ & $1.17\pm3.01 $ & $0.27\pm0.20$ & & $0.95\pm2.57$ & $0.95$ & $0.72\pm1.19 $ & $0.23\pm0.18$ & & $0.86\pm2.17$ & $0.87$ & $0.65\pm1.04 $ & $0.22\pm0.17$ \\
    Peng (2 runs) & $\textbf{0.75}\pm1.70$ & $\textbf{0.77}$ & $\textbf{0.59}\pm0.82 $ & $0.18\pm0.15$ & & $0.66\pm1.50$ & $0.69$ & $0.54\pm0.78 $ & $0.16\pm0.12$ & & $0.65\pm1.47$ & $0.68$ & $0.53\pm0.78 $ & $0.15\pm0.12$\\
    BU3D (2 runs) & $1.61\pm11.03$ & $1.00$ & $0.67\pm2.86 $ & $0.33\pm0.22$ & & $0.67\pm3.88$ & $0.72$ & $0.39\pm0.93 $ & $0.33\pm0.20$ & & $0.61\pm2.92$ & $0.70$ & $0.37\pm0.63 $ & $0.33\pm0.19$ \\
    Proposed & $1.88\pm9.93$ & $1.39$ & $1.22\pm7.12$ & $\textbf{0.16}\pm0.16$ & & $\textbf{0.62}\pm3.77$ & $\textbf{0.48}$ & $\textbf{0.35}\pm2.08$ & $\textbf{0.13}\pm0.13$ & & $\textbf{0.51}\pm2.82$ & $\textbf{0.39}$ & $\textbf{0.26}\pm0.77$ & $\textbf{0.12}\pm0.13$ \\ [2pt]
    \hline
    \noalign{\vspace{2pt}}
    \multicolumn{6}{l}{PPP = 4} \\
    RT3D & & $0.98$ & $0.41\pm0.69$ & $0.56\pm0.60$ & & & $0.54$ & $0.26\pm0.26$ & $0.28\pm0.26$ & & & $0.63$ & $0.38\pm0.36$ & $0.25\pm0.19$ \\
    Lindell (2 runs)& $0.81\pm2.05$ & $0.81$ & $0.61\pm1.01 $ & $0.21\pm0.16$ & & $0.97\pm2.20$ & $0.98$ & $0.76\pm1.26 $ & $0.22\pm0.16$ & & $1.09\pm2.31$ & $1.10$ & $0.86\pm1.38 $ & $0.24\pm0.17$ \\
    Peng (2 runs) & $0.53\pm1.26$ & $0.57$ & $0.43\pm0.66 $ & $0.13\pm0.11$ & & $0.46\pm1.08$ & $0.50$ & $0.37\pm0.55 $ & $0.13\pm0.10$ & & $0.43\pm0.99$ & $0.48$ & $0.35\pm0.51 $ & $0.12\pm0.10$ \\
    BU3D (2 runs) & $0.55\pm2.13$ & $0.69$ & $0.36\pm0.77 $ & $0.33\pm0.16$ & & $0.49\pm1.51$ & $0.64$ & $0.34\pm0.30 $ & $0.30\pm0.15$ & & $0.49\pm1.43$ & $0.63$ & $0.34\pm0.29 $ & $0.29\pm0.15$ \\
    Proposed & $\textbf{0.35}\pm2.51$ & $\textbf{0.24}$ & $\textbf{0.16}\pm0.86$ & $\textbf{0.09}\pm0.11$ & & $\textbf{0.28}\pm1.99$ & $\textbf{0.17}$ & $\textbf{0.10}\pm0.19$ & $\textbf{0.07}\pm0.10$ & & $\textbf{0.27}\pm2.07$ & $\textbf{0.16}$ & $\textbf{0.09}\pm0.15$ & $\textbf{0.06}\pm0.10$ \\ [2pt]
    
    \hline
    \noalign{\vspace{2pt}}
    \multicolumn{6}{l}{PPP = 16} \\
    RT3D & & $0.73$ & $0.35\pm1.40$ & $0.38\pm0.39$ & & & $0.51$ & $0.28\pm0.45$ & $0.23\pm0.17$ & & & $0.46$ & $0.26\pm0.23$ & $0.20\pm0.12$ \\
    % Peng \cite{peng} &  & 0.0053 & 0.0024 & 0.0029 & &  & 0.0051 & 0.0023 & 0.0028 & &  & 0.0051 & 0.0023 & 0.0028 \\
    Lindell (2 runs) & $2.38\pm2.82$ & $2.50$ & $2.05\pm2.05 $ & $0.45\pm0.31$ & & $2.73\pm2.60$ & $3.03$ & $2.40\pm1.86 $ & $0.63\pm0.39$ & & $2.83\pm2.50$ & $3.22$ & $2.50\pm1.75 $ & $0.72\pm0.43$ \\
    Peng (2 runs) & $0.34\pm0.83$ & $0.39$ & $0.28\pm0.36 $ & $0.12\pm0.10$ & & $0.30\pm0.69$ & $0.36$ & $0.24\pm0.28 $ & $0.11\pm0.10$ & & $0.29\pm0.65$ & $0.35$ & $0.24\pm0.26 $ & $0.12\pm0.10$ \\
    BU3D (2 runs) & $0.43\pm1.34$ & $0.56$ & $0.31\pm0.46 $ & $0.24\pm0.14$ & & $0.37\pm1.07$ & $0.47$ & $0.27\pm0.24 $ & $0.20\pm0.13$ & & $0.35\pm0.97$ & $0.44$ & $0.26\pm0.22 $ & $0.18\pm0.13$ \\
    Proposed & $\textbf{0.21}\pm1.90$ & $\textbf{0.10}$ & $\textbf{0.06}\pm0.14$ & $\textbf{0.04}\pm0.08$ & & $\textbf{0.19}\pm1.81$ & $\textbf{0.07}$ & $\textbf{0.04}\pm0.12$ & $\textbf{0.03}\pm0.08$ & & $\textbf{0.18}\pm1.78$ & $\textbf{0.06}$ & $\textbf{0.03}\pm0.09$ & $\textbf{0.03}\pm0.07$ \\ [2pt]
    \hline
    \end{tabular}%
    }
    \label{tab:bowling}
\end{table*}

% ------------------------------

\textbf{Evaluation metrics.} We employ two popular evaluation metrics to measure the accuracy of the predicted depth maps. The first metric is the Depth Absolute Error (DAE), $L_1$ error norm, defined as $\operatorname{DAE}\,(\boldsymbol x, \boldsymbol {x^*}) = \frac{1}{N} \| \boldsymbol x - \boldsymbol {x}^* \|_1$ where $N$ is the number of pixels. Another metric used is the $L_1$-Chamfer distance between two point clouds $S_1$ and $S_2$, defined as
\begin{equation}
    d_{\operatorname{CD}}(S_1, S_2) = \frac{1}{N}\left(\sum_{\boldsymbol a \in S_1} \min _{\boldsymbol b \in S_2}\|\boldsymbol a-\boldsymbol b\|_1+\sum_{\boldsymbol b \in S_2} \min _{\boldsymbol a \in S_1}\|\boldsymbol a-\boldsymbol b\|_1\right)
\end{equation}
where $|S_1|=|S_2|=N$ are the number of points in the point clouds, the first term measures the average distance from each point in $S_1$ to the nearest point in $S_2$ and the second term measures the average distance from each point in $S_2$ to the nearest point in $S_1$. To analyze the results, we will report both terms in the Chamfer distance.

% \textbf{Qualitative comparison.} Fig.~\ref{fig:pc_art} reports the reconstructed point clouds for the Art scene. Note that existing deep learning methods of Lindell, Peng and BU3D are run twice to estimate the depth of the first and the second surfaces separately. The results by Peng observe bleeding artifacts, while the result by BU3D show less such artifacts, but observe some outlier points in the third row. Compared to other methods, the proposed method reconstructs the point clouds with less bleeding artifacts with less outliers, but often has outliers in the far region in the third row. Fig.~\ref{fig:img_art} shows the estimated depth maps for the same scene. Overall, the results are consistent with the point cloud results. The last column shows the uncertainty information by the proposed method. The uncertainty map shows higher values in the regions with outliers, indicating the uncertainty of the depth estimation.
%^ original TCI
\textbf{Qualitative comparison.} 
Fig.~\ref{fig:pc_art} reports the reconstructed point clouds for the Art scene. Note that existing deep learning methods of Lindell, Peng and BU3D are run twice to estimate the depth of the first and the second surfaces separately. 
The results by Peng observe bleeding artifacts, while the results by BU3D show fewer such artifacts but observe some outlier points in the third row. Meanwhile, RT3D shows fewer bleeding artifacts as well, but yields a sparser reconstruction in the third row with fewer points.
Compared to other methods, the proposed method reconstructs the point clouds with less bleeding artifacts with less outliers, but often has outliers in the far region in the third row. Fig.~\ref{fig:img_art} shows the estimated depth maps for the same scene. Overall, the results are consistent with the point cloud results. The last column shows the uncertainty information by the proposed method. The uncertainty map shows higher values in the regions with outliers, indicating the uncertainty of the depth estimation.

Fig.~\ref{fig:pc_bowling1} shows the estimated point clouds for the Bowling scenes. Similar to the Art scene, the proposed method yields point clouds with more clear boundaries than Peng's result and less outliers than BU3D's result. One disadvantage of the proposed method is that it has some outliers in the far region. Fig.~\ref{fig:img_bowling1} shows the corresponding depth maps with uncertainty information by the proposed method in the last column.
% BU3D: 11.49(initial), 0.48(infer)
% Proposed: 282.74 sec(initial), 11.49(initial) 2.08 sec(infer) , new initial(14.78450) -> 16.9
% running time
\begin{table}[ht]  \centering
    \caption{Comparison of running time in GPU on the Art scene.}
   \begin{tabular}{l c r r r }
   \toprule
   \multicolumn{1}{c}{Method} &   Runtime (sec) & Train time & Parameters \\ \midrule
   Lindell~\cite{lindell2018singlephoton} (2 runs)  & 427.6 & 24 hours & 1,728,996 \\
   Peng~\cite{peng2020photonefficient} (2 runs) & 74.6 & 35 hours & 568,298 \\
   BU3D~\cite{peng2020photonefficient} (2 runs) & 11.9 & 9 hours & 53,136 \\
   Proposed  & 13.5 & 9 hours & 21,024 \\
   \bottomrule
   \end{tabular}
   \label{tab:time}
\end{table}

\textbf{Quantitative comparison.} Table~\ref{tab:art} and~\ref{tab:bowling} report the quantitative results for the Art scene and the Bowling scene, respectively. To analyze the results, we report the Chamfer distance with two terms separately: O-To-G (distance from the estimated point cloud to the ground truth) and G-To-O (distance from the ground truth to the estimated point cloud). In all scenarios, the proposed method shows the lowest error for G-To-O, which indicates that the estimated point cloud is closer to the ground truth, albeit with some outliers. When PPP and SBR are both 1, Peng's method shows the lowest errors except for G-To-O, but as PPP and SBR increase, the proposed method shows better performance overall. Fig.~\ref{fig:contour_art} visualizes the errors with different levels of PPP and SBR for the Art scene. When PPP and SBR are both 1, Peng's method shows the lowest error. As PPP and SBR increase, the proposed method shows the lowest error overall. Table~\ref{tab:time} compares the running time of the proposed method with other methods. While other methods run twice, the proposed method only once. The runtime (13.5 seconds) of the proposed method consists of two parts: estimating the initial multiscale (11.49 seconds) and the inference time (2.8 seconds). The number of parameters of the proposed method is shown to be the smallest among the methods, which indicates the efficiency of the proposed method. The proposed method requires 20.3 GB of GPU memory for training. For testing on the Art scene, it requires 21.5 GB for initial mulstiscale estimation and 1.7 GB for inference.

% We also compare the proposed method with the previous conference paper~\cite{koo2024bayesian} on the Art scene in Table~\ref{tab:art_conference}. We change TODO
% The proposed method shows better performance than the previous method, especially when SBR is 16.

% \begin{table}[H]
%     \centering
%     \caption{Comparison with previous method~\cite{koo2024bayesian} on the Art scene.}
%     \resizebox{\columnwidth}{!}{%
%     % \resizebox{\linewidth}{!}{%
%     \begin{tabular}{l c c c c c}
%     \hline
%     \noalign{\vspace{2pt}}
%      & SBR = 1 & & SBR = 4 & & SBR = 16 \\[2pt]
%     \noalign{\vspace{2pt}}
%     \hline
    
%     \noalign{\vspace{2pt}}
%     \multicolumn{6}{l}{PPP = 4} \\

%     Conference & $0.0069\pm0.0346$  & & $0.0058\pm0.0282$ & & $0.0055\pm0.0267$\\
%     Proposed & $\textbf{0.0058}\pm0.0313$ & & $\textbf{0.0047}\pm0.0245$ & & $\textbf{0.0045}\pm0.0228$ \\[2pt]
    
%     \hline

%     \end{tabular}%
%     }
%     \label{tab:art_conference}
% \end{table}

\textbf{Ablation study.} To analyze the network architecture, we perform an ablation study in terms of the number $L$ of multiple scales and the number of graph attention layers. As shown in Table~\ref{tab:ablation}, the number of multiple scales affects the performance. We choose $L=4$, because increasing $L$ beyond this value does not significantly improve the results. 
% This number includes all the graph attention layers both in the squeeze parts and the expansion parts.
We also report the effect of the number of graph attention layers. 
This number includes all the graph attention layers both in the squeeze parts and the expansion parts.
Since the number of graph attention layers affects the number of parameters and the computational cost, we choose 36 graph attention layers, as a trade-off between performance and computational cost.
The reported uncertainty corresponds to the average uncertainty of the depth absolute error (DAE) across all pixels. We also studied an effect on two hyperparameters: the number of neighbors in the k-NN and the Gumbel-Softmax temperature used for hard attention, where we observed no significant performance changes. Throughout the experiments, we set k in kNN to 6 and the temperature to 10.

%We have also checked the effect of $k$ in the k-NN algorithm, but the effect is not significant.

% -----------------------------------------
% Contour Art
\def\fh{310pt}
\begin{figure}[t]\center
\includegraphics[totalheight=\fh,trim={0 0.1cm 0 0},clip]{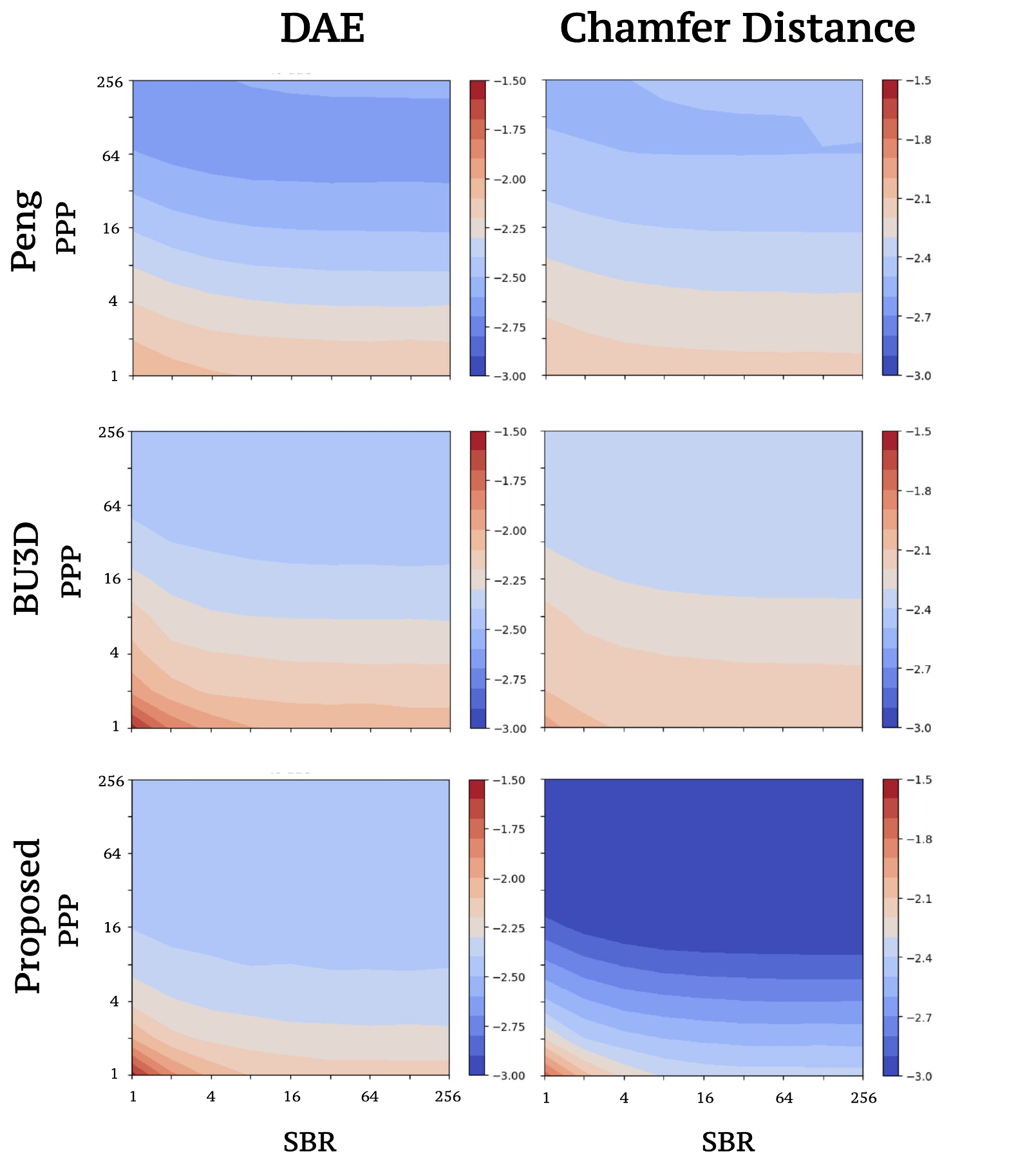}
\caption{Errors in terms of different levels of SBR and PPP on the Art scene
by three methods: Peng, BU3D and the proposed (Top-to-Bottom). Two
evaluation metrics of DAE and Chamfer Distance are used (Left-to-Right) and the
error values are presented in a base-10 log scale.}
\label{fig:contour_art}
\end{figure}
   
% Ablation study
% Art, 1, 1 & 2,2 & 4,4 
\begin{table}
    \centering
    \caption{Ablation study on Art scene with PPP and SBR = 2.0}
    \resizebox{0.49\textwidth}{!}{
    \begin{tabular}{ c c c c c c c c c }
    \toprule
    %\noalign{\vspace{2pt}}
    L & \#Conv. & k in kNN & Temperature & DAE &  Uncertainty & \#Params \\[2pt]
    \midrule
    %\noalign{\vspace{2pt}}
    2 & 36 & 6 & 10 & 0.0595 & 0.0258  & 17808 \\[2pt]
    % \hline
    %\noalign{\vspace{2pt}}
    3 &  & & & 0.0069 & 0.0319  & 19416 \\[2pt]
    % \hline
    %\noalign{\vspace{2pt}}
    4 & & & & 0.0067  & 0.0078  & 21024 \\[2pt]
    % \hline
    %\noalign{\vspace{2pt}}
    5 & & & & 0.0072 & 0.0151  & 22632 \\[2pt]
    \midrule
    %\noalign{\vspace{2pt}}
    4 & 12 & 6 & 10 & 0.0069 & 0.0104  & 1056 \\[2pt]
    %\noalign{\vspace{2pt}}
    & 24 & & &  0.0069 & 0.0087  & 7584 \\[2pt]
    %\noalign{\vspace{2pt}}
    & 36 & & &  0.0067 & 0.0078  & 21024 \\[2pt]
    \midrule
    %\noalign{\vspace{2pt}}
    4 & 36 & 4 & 10 & 0.0066 & 0.0078  & 21024 \\[2pt]
     &  & 6, 8 &  & 0.0067 & 0.0078  & 21024 \\[2pt]
    \midrule
    4 & 36 & 6 & 0.1-100&  0.0067 & 0.0078  & 21024 \\
    \bottomrule
    \end{tabular}
    }
    \label{tab:ablation}
\end{table}

\begin{figure*}[t]
    \centering
    \def\fh{75pt}
    \def\sbra{8.57} \def\sbrb{4.29} \def\sbrc{2.14}
    \def\sbrd{1.07} \def\sbre{0.54} \def\sbrf{0.27}
    \begin{tabular}{c@{ }c@{ }c@{ }c@{ }c@{ }c@{\hspace{5pt}}c}
    
    \rotatebox[origin=l]{90}{\small\parbox{4cm}{$\,$SBR$=\sbra$\\(original)}} &
    \includegraphics[totalheight=\fh]{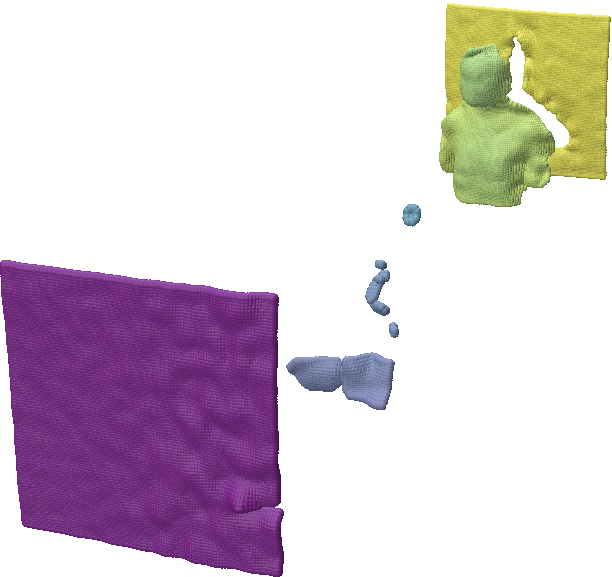} & 
    \includegraphics[totalheight=\fh]{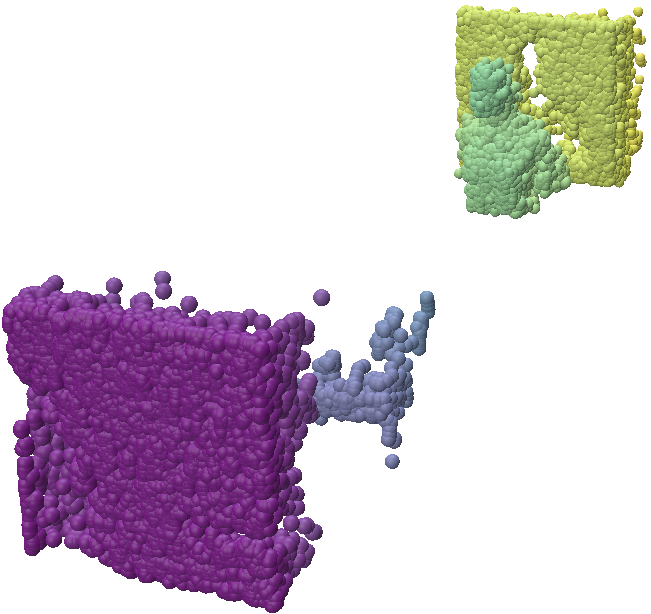} &
    \includegraphics[totalheight=\fh]{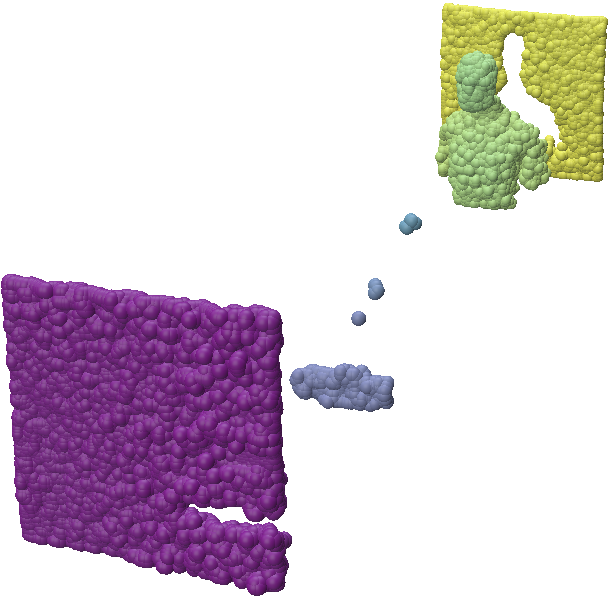} & 
    \includegraphics[totalheight=\fh]{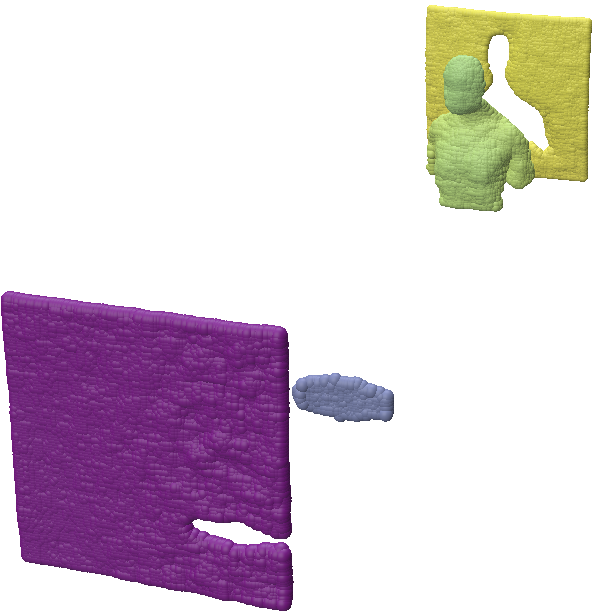}&
    \includegraphics[totalheight=\fh]{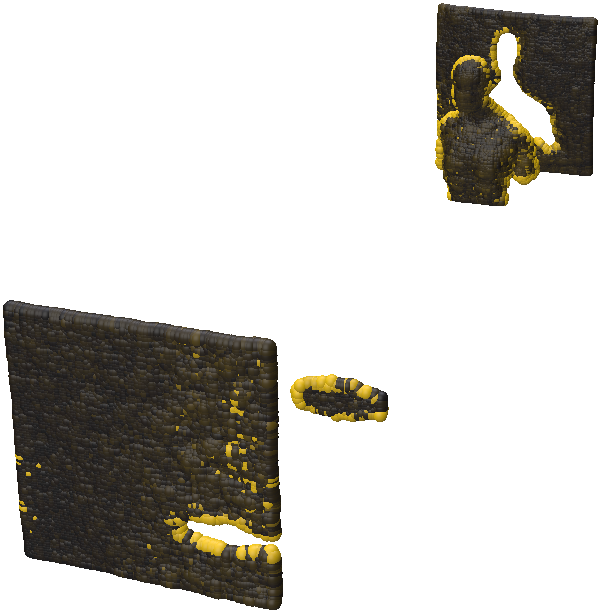} &
    \includegraphics[totalheight=\fh]{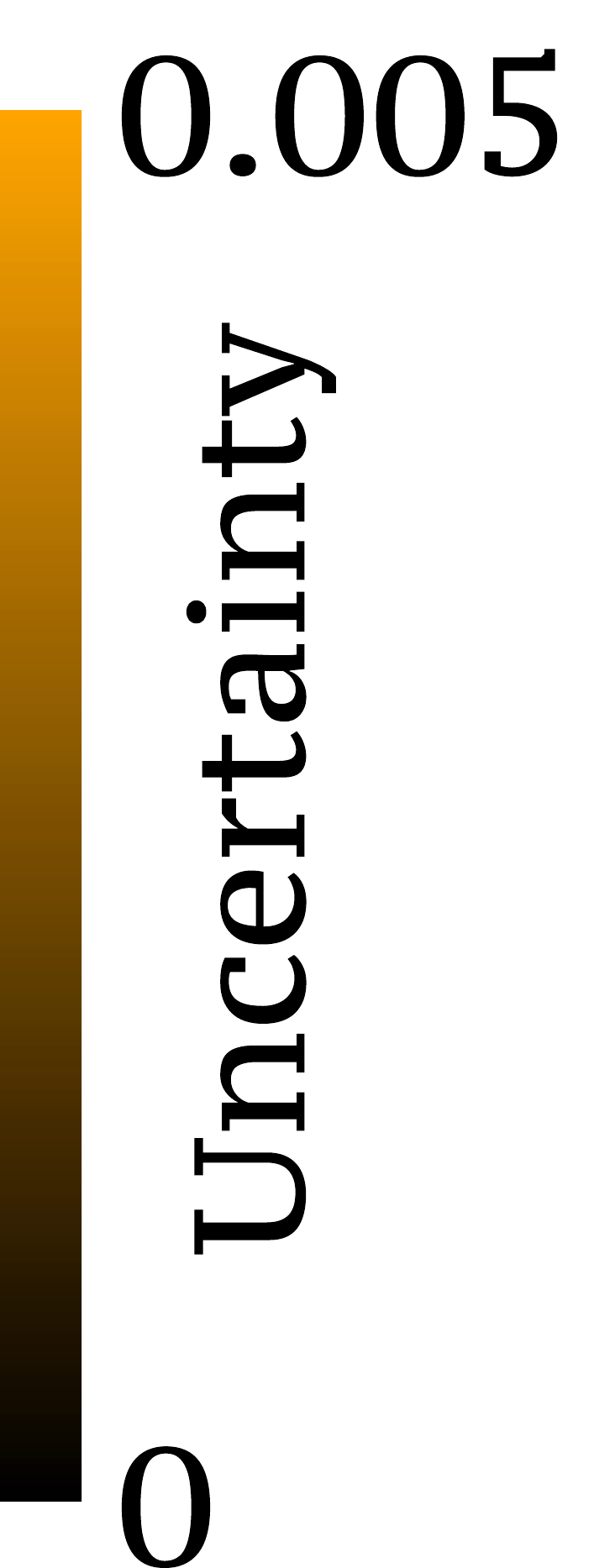} \\[-35pt]
    
    \rotatebox[origin=l]{90}{\small\parbox{4cm}{$\,$SBR$=\sbrd$\\\phantom{a}}} &
    \includegraphics[totalheight=\fh]{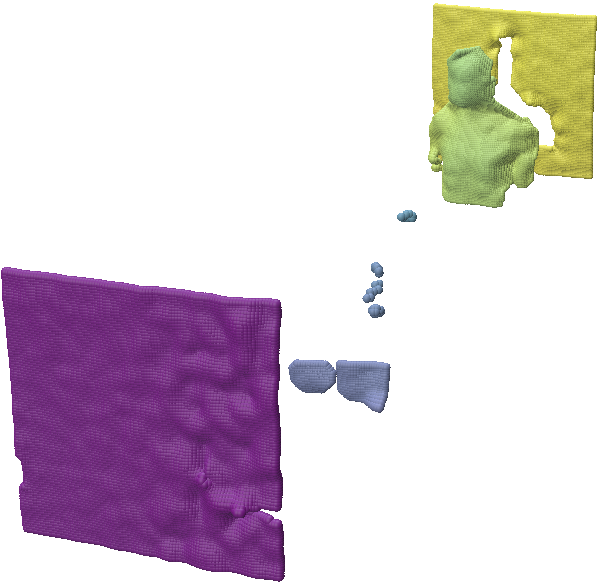} &
    \includegraphics[totalheight=\fh]{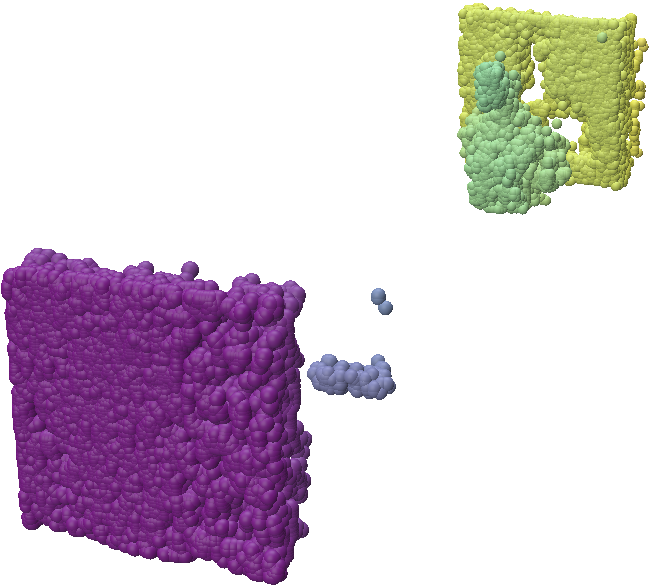} &
    \includegraphics[totalheight=\fh]{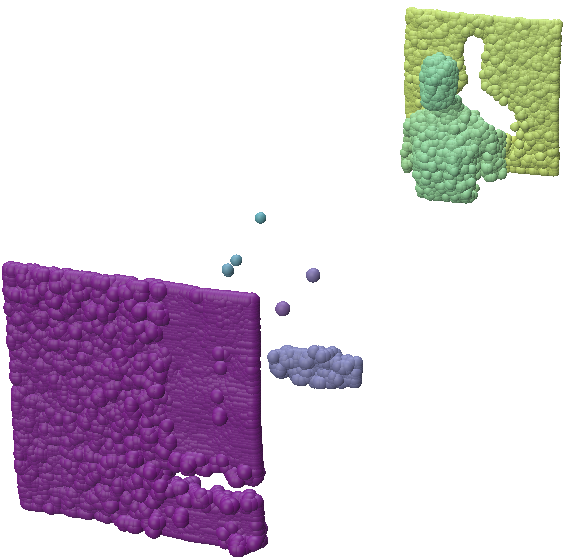} & 
    \includegraphics[totalheight=\fh]{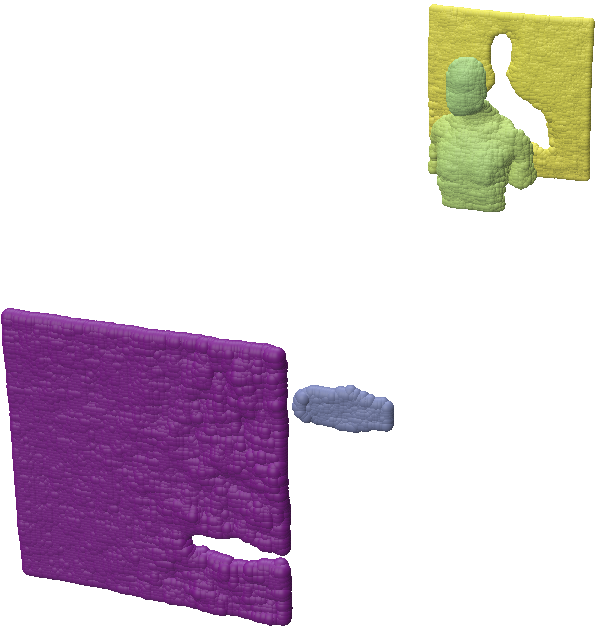}&
    \includegraphics[totalheight=\fh]{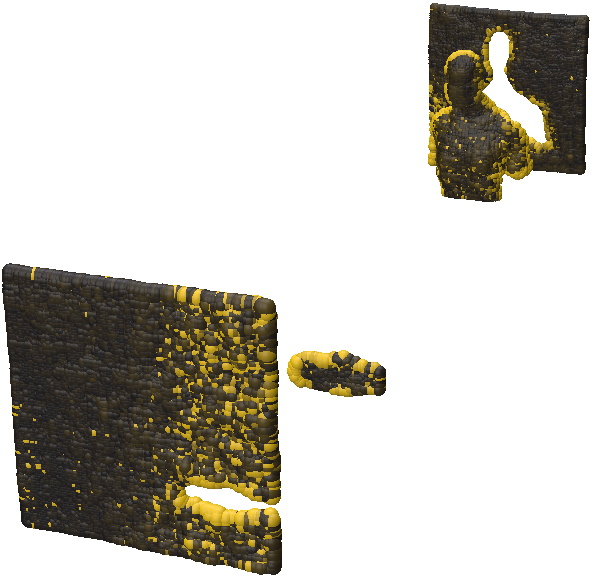}&
    \includegraphics[totalheight=\fh]{fig/real_data/Uncertainty_pc_colorbar.pdf} \\[-35pt]
    
    % \rotatebox[origin=l]{90}{\small\parbox{4cm}{$\,$SBR$=\sbre$}} &
    % \includegraphics[totalheight=\fh]{fig/real_data/Peng_\sbre.png} &
    % \includegraphics[totalheight=\fh]{fig/real_data/RT3D_\sbre.png} &
    % \includegraphics[totalheight=\fh]{fig/real_data/BU3D_\sbre.png} & 
    % \includegraphics[totalheight=\fh]{fig/real_data/Proposed_\sbre.png}&
    % \includegraphics[totalheight=\fh]{fig/real_data/Uncertainty_clip_\sbre.png}&
    % \includegraphics[totalheight=\fh]{fig/real_data/Uncertainty_pc_colorbar.pdf} \\[-10pt]
    
    \rotatebox[origin=l]{90}{\small\parbox{4cm}{$\,$SBR$=\sbrf$\\\phantom{a}}} &
    \includegraphics[totalheight=\fh]{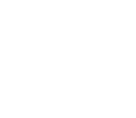} &
    \includegraphics[totalheight=\fh]{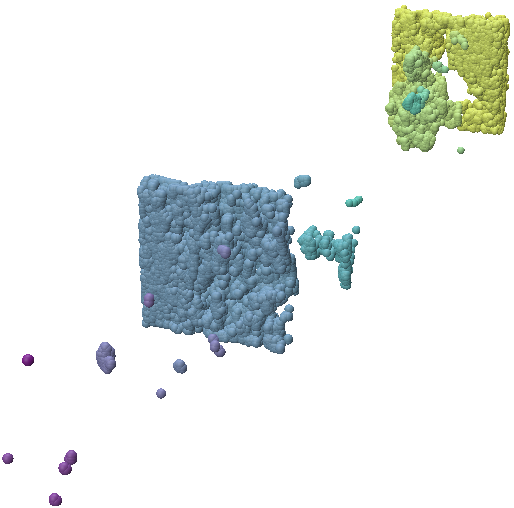} &
    \includegraphics[totalheight=\fh]{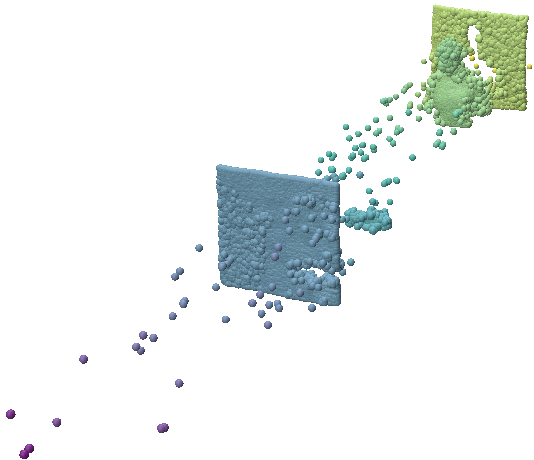} & 
    \includegraphics[totalheight=\fh]{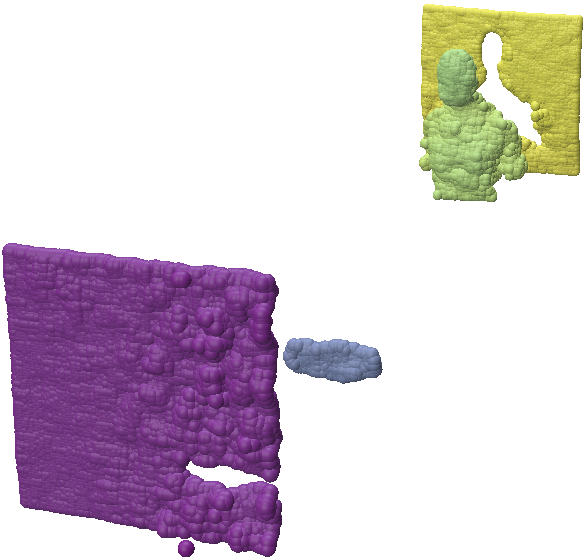}&
    \includegraphics[totalheight=\fh]{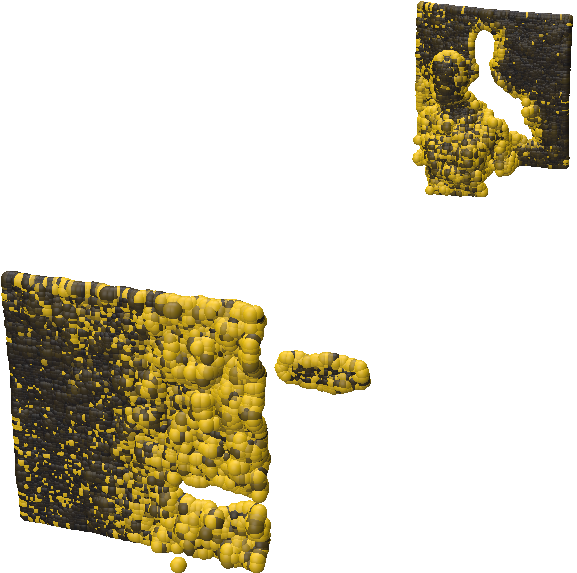}&
    \includegraphics[totalheight=\fh]{fig/real_data/Uncertainty_pc_colorbar.pdf} \\[2pt]
    & RT3D & Peng (2 runs) & BU3D (2 runs) & Proposed & Uncertainty &
    \end{tabular}
    \caption{Reconstruction results on the real data of Mannequin behind scattering object~\cite{shin2016computational} with different SBR levels.}
    \label{fig:real}
\end{figure*}

% -------------------------------------
% Point clouds for Lindell Real data

\begin{figure*}[tp]
    \centering
    \def\fw{0.15\textwidth}  %0.18

    %checkerboard size
    \def\fwrc{0.155\textwidth} % For RT3D width for checkerboard
    \def\fwlc{0.145\textwidth} % For Lindell width for checkerboard
    \def\fwpc{0.145\textwidth} % For peng width for checkerboard
    \def\fwbc{0.145\textwidth} % For BU3D width for checkerboard

    %elephant size
    \def\fwpe{0.145\textwidth} % For PEng width for Elephant
    \def\fwbe{0.165\textwidth} % For BU3D width for Elephant

    %roll size
    \def\fwlr{0.155\textwidth} % For Lindell width for Roll
    \def\fwpr{0.145\textwidth} % For Peng width for Roll
    \def\fwbr{0.165\textwidth} % For BU3D width for Roll

    \begin{tabular}{c@{}c@{}c@{}c@{}c@{}c@{\hspace{5pt}}c@{\hspace{5pt}}c}

    \includegraphics[width=\fwrc]{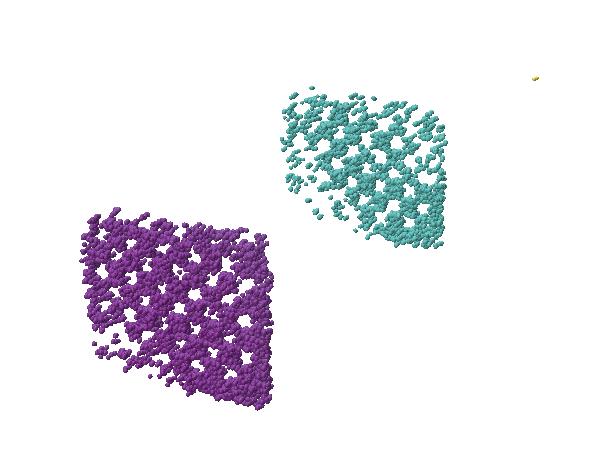} & %RT3D
    \includegraphics[width=\fwlc]{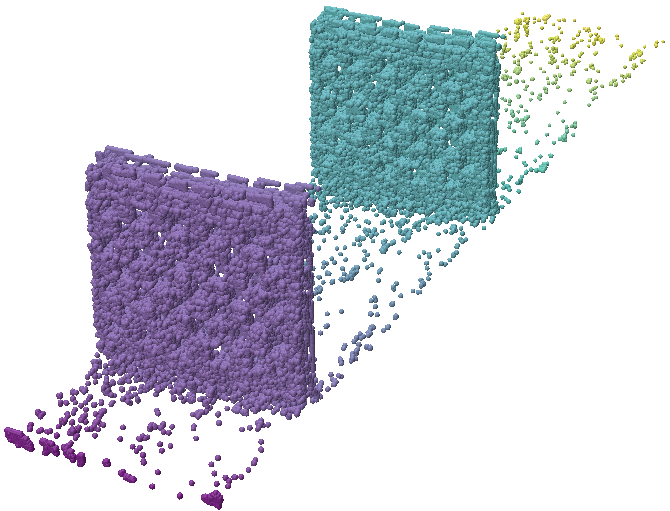} & %lindell
    \includegraphics[width=\fwpc]{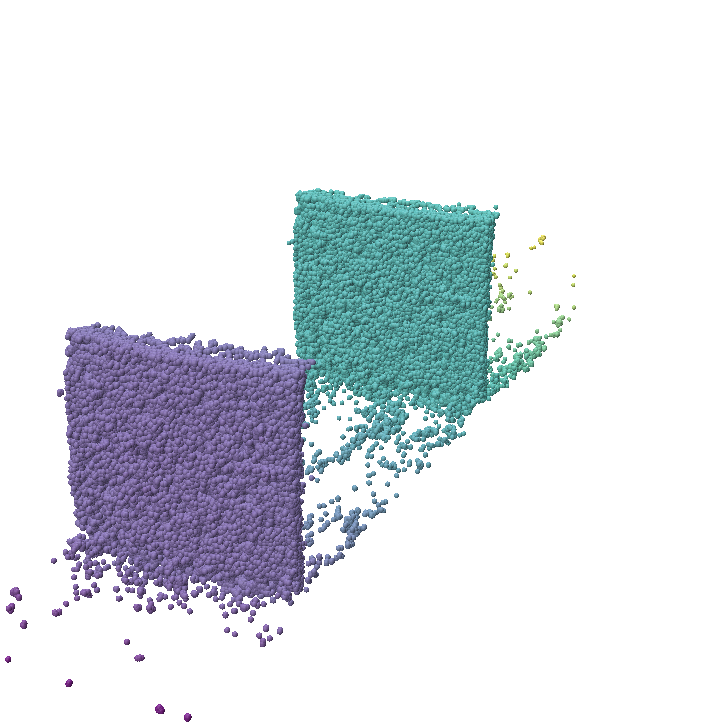} & %Peng
    \includegraphics[width=\fwbc]{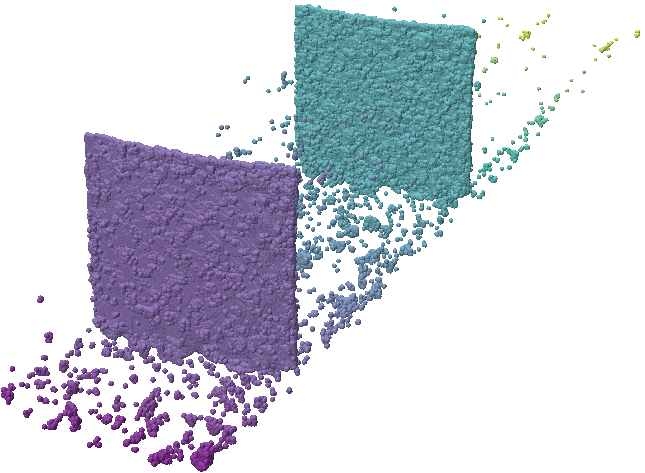} & % BU3D
    \includegraphics[width=\fw]{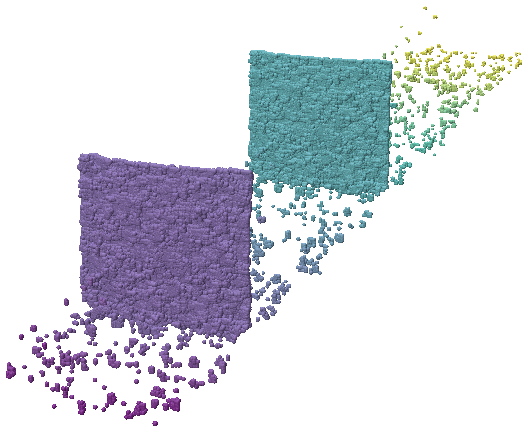} & %Proposed
    \includegraphics[width=\fw]{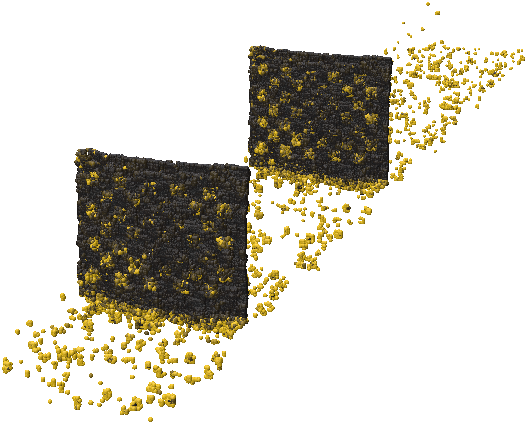} &
    \includegraphics[totalheight=70pt]{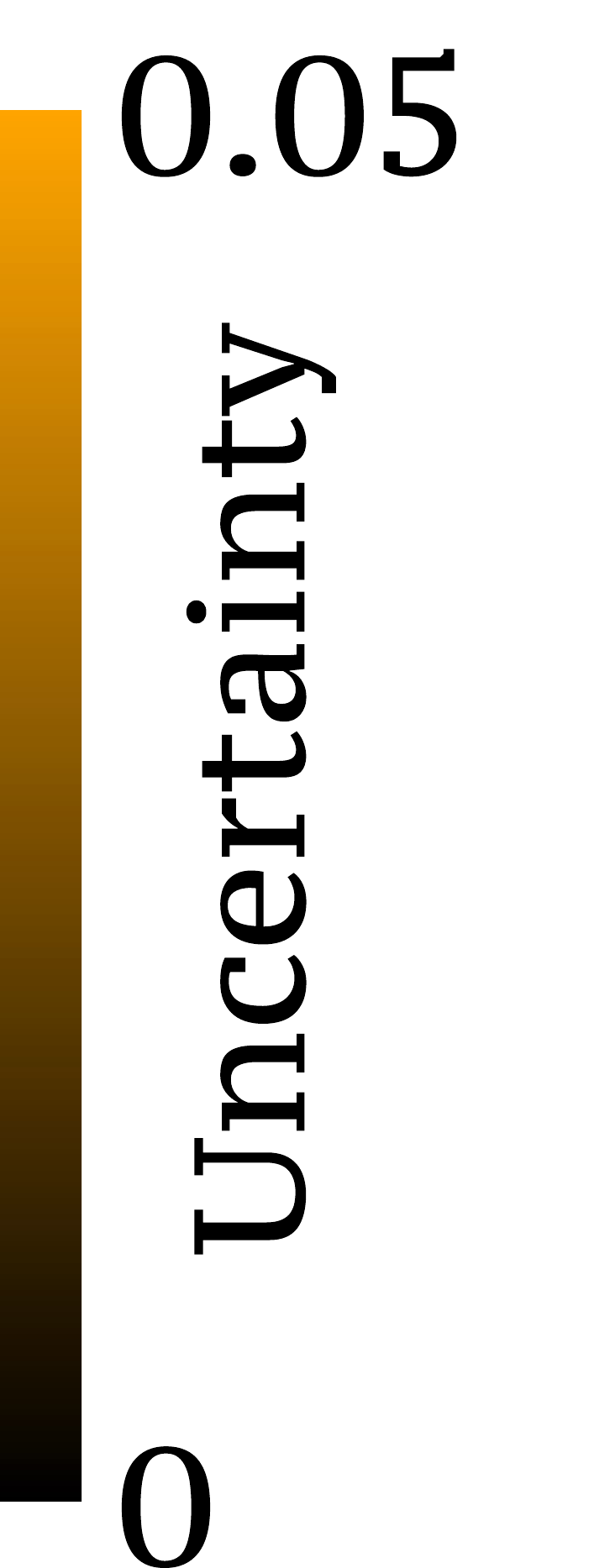} \\[0pt]

    \includegraphics[width=\fw]{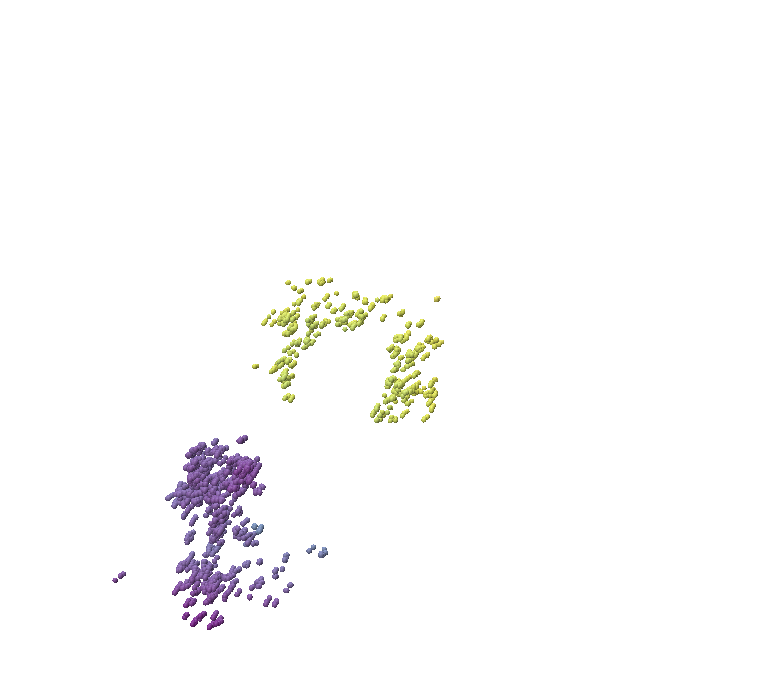} & %RT3D
    \includegraphics[width=\fw]{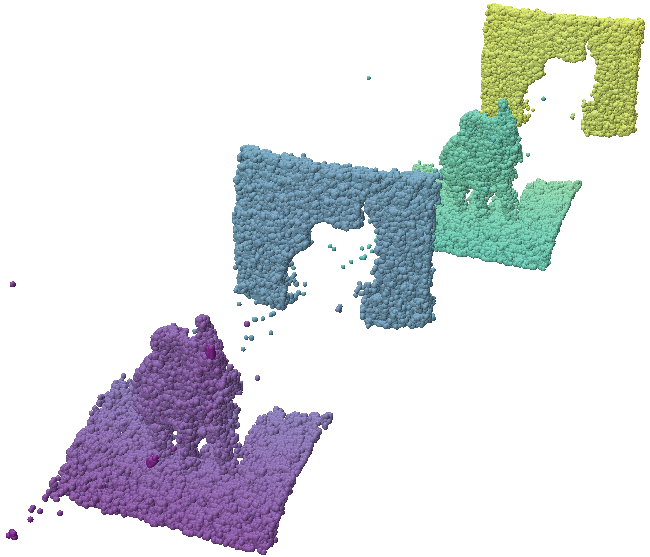} & %lindell
    \includegraphics[width=\fwpe]{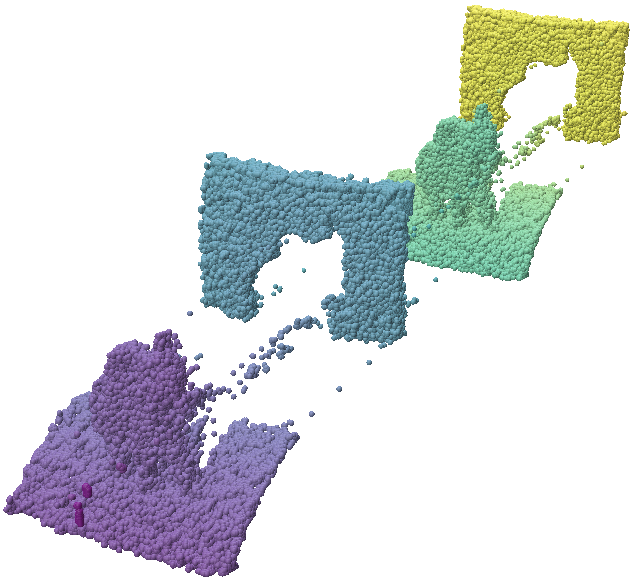} & %Peng
    \includegraphics[width=\fwbe]{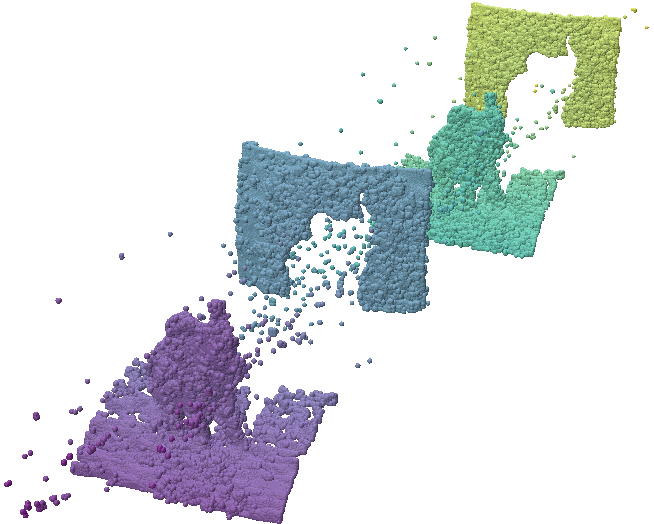} & % BU3D
    \includegraphics[width=\fw]{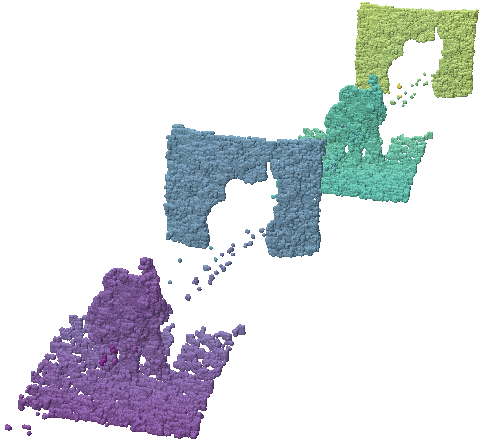} & %Proposed
    \includegraphics[width=\fw]{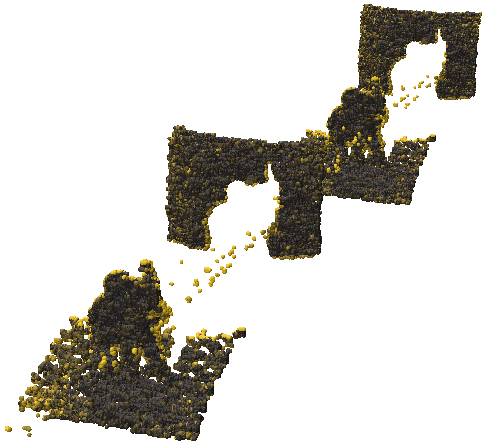} &
    \includegraphics[totalheight=70pt]{fig/real_data/Lindell_real/Uncertainty_lindell_real_colorbar.pdf} \\[0pt]

    \includegraphics[width=\fw]{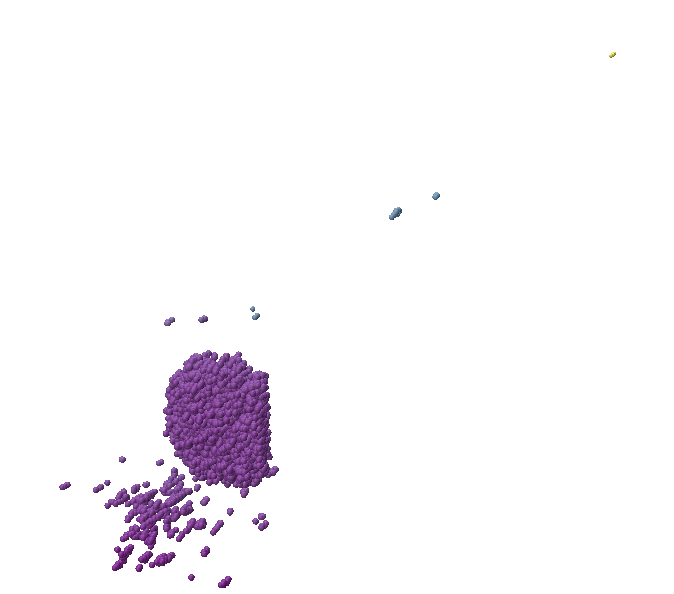} & %RT3D
    \includegraphics[width=\fwlr]{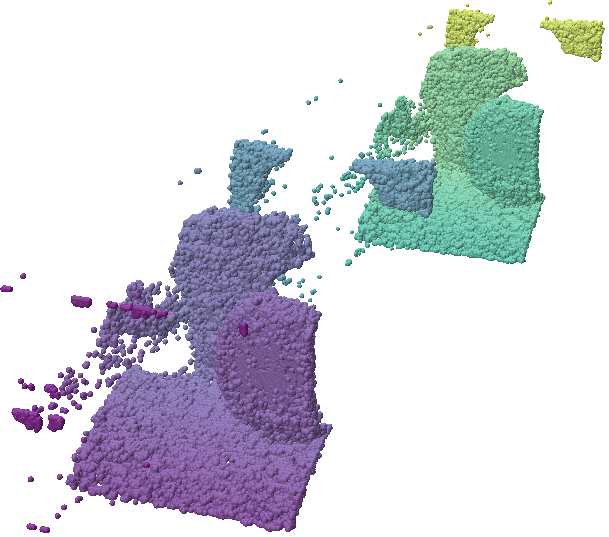} & %lindell
    \includegraphics[width=\fwpr]{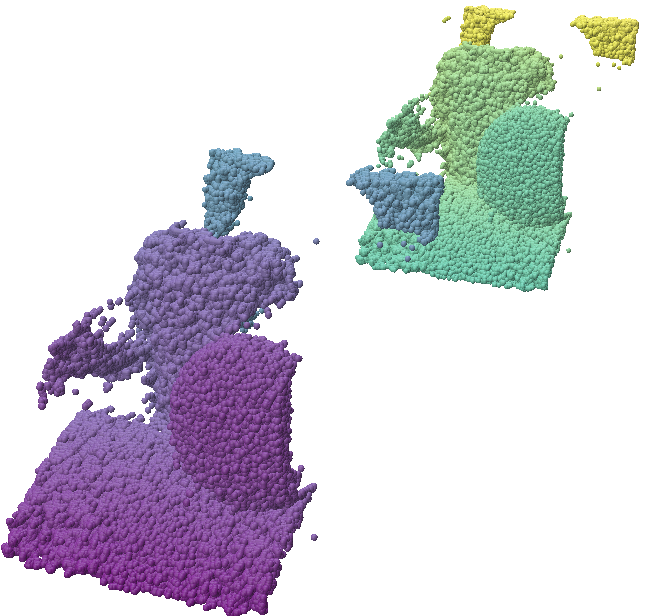} & %Peng
    \includegraphics[width=\fwbr]{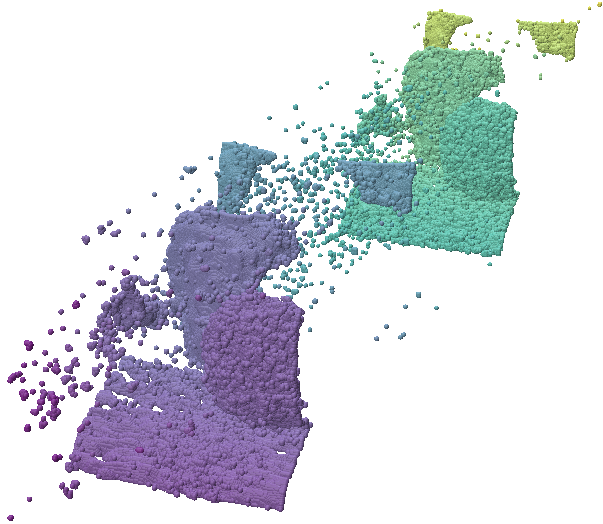} & % BU3D
    \includegraphics[width=\fw]{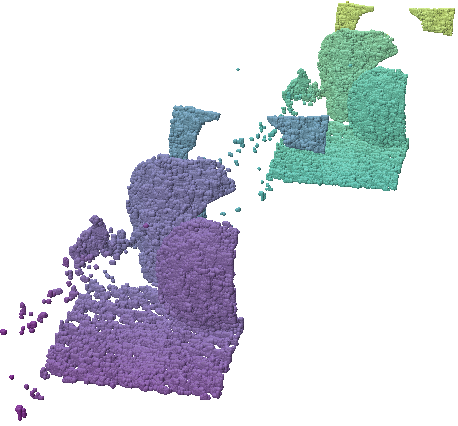} & %Proposed
    \includegraphics[width=\fw]{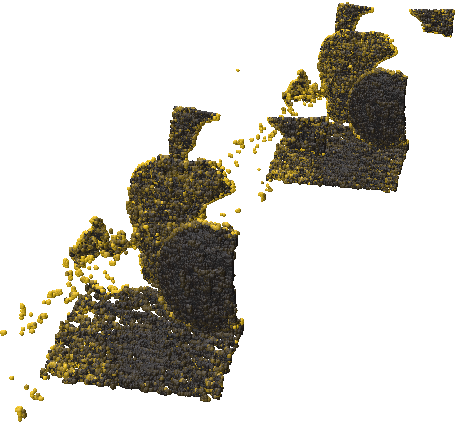} & %Uncertainty
    \includegraphics[totalheight=70pt]{fig/real_data/Lindell_real/Uncertainty_lindell_real_colorbar.pdf} \\[0pt]

    \scriptsize RT3D~\cite{tachella2019realtime} &\scriptsize Lindell~\cite{lindell2018singlephoton} (2 runs) &\scriptsize  Peng~\cite{peng2020photonefficient} (2 runs) &\scriptsize  BU3D~\cite{koo2022bayesiana} (2 runs) &\scriptsize  Proposed & \scriptsize Uncertainty
    
    \end{tabular}
    
    \caption{Reconstructed point clouds from real dataset~\cite{lindell2018singlephoton} modified to contain dual peaks. Lindell, Peng and BU3D are run twice to estimate the first and second surfaces separately. RT3D and the proposed method.}
    \label{fig:lindell_real}
\end{figure*}
% -------------------------------------

\subsection{Results on real data}

% This dataset is widely used for evaluating the performance of single-photon depth imaging methods.
We evaluate the proposed method on the real dual-peak data of Mannequin behind scattering object~\cite{shin2016computational} with different SBR levels. The histogram dataset has a spatial-time resolution of $99 \times 99 \times 4001$ with PPP=45 and SBR=8.57. To generate different SBR levels, we add background photons, so that the new SBR levels are 8.57, 2.14, 1.07, 0.54, and 0.27. For the comparison, we run Peng~\cite{peng2020photonefficient} and BU3D~\cite{koo2022bayesiana} two times in the region of scattering medium and the Mannequin, separately. In this regard, they have the advantage of knowing the approximate positions of the surfaces. We also compare our results to the multi-surface reconstruction method RT3D~\cite{tachella2019realtime}.

The reconstruction results are shown in Fig.~\ref{fig:real}. As the SBR level decreases, the quality of the reconstructed point clouds decreases for other methods. When SBR=0.27, the RT3D fails to reconstruct meaningful point clouds. The proposed method outperforms the other methods in terms of the quality of the reconstruction. As the SBR decreases, we observe the higher uncertainty values. This experiment shows the effectiveness of the proposed method in the real dataset with an accurate reconstruction having uncertainty estimation. 
The proposed method takes 1.71 seconds in total --- 1.53 seconds for initial multiscale estimation and 0.18 seconds for inference. The memory usage is 2.5GB for initial multiscale estimation and 0.5GB for inference. Compared to the Art scene, the memory usage is much lower due to the smaller spatial resolution.

% MIT time
% Total: 1.70645 Initial:1.52545 , inference:0.181s - 0.160s, 0.013s, 0.008s (for each stage)

% Lindell real

We conduct an additional experiment on another real dataset provided in~\cite{lindell2018singlephoton}.
The histograms in this dataset have a spatial-time resolution of $256 \times 256 \times 1536$, with scenes recorded in challenging environments.
To make dual peaks in each pixel's histogram, we modify the dataset by duplicating the portion of the histogram corresponding to the object.
% Similar to previous experiments, we run Lindell~\cite{lindell2018singlephoton}, Peng~\cite{peng2020photonefficient} and BU3D~\cite{koo2022bayesiana} twice, while RT3D~\cite{tachella2019realtime} and our method run only once without knowing prior knowledge.
Fig.~\ref{fig:lindell_real} shows the reconstructed pointclouds. In the checkerboard scene (1st row), both BU3D and our method show noisy results but produce flatter surfaces than Peng. In contrast, RT3D yields inaccurate sparse point clouds due to irregular reflectivity. 
In the elephant scene (2nd row), Peng and our method show fewer artifacts than BU3D. In the last row, the proposed method observes some outliers, but produces less noisy results than BU3D. Considering the fact that the proposed method runs only once, it shows a competitive performance compared to Peng's results.

% Fig.~\ref{fig:lindell_real} shows the reconstructed pointclouds. In the checkerboard scene (1st row), both BU3D and our method show noisy results but produce flatter surfaces than Peng. In contrast, RT3D yields inaccurate sparse point clouds due to irregular reflectivity. 
% In the elephant scene (2nd row), Peng and our method show fewer artifacts than BU3D. Furthermore, our method reconstructs the elephant with a slimmer shape, while Peng's result appears bulkier.
% In the last row, RT3D fails to reconstruct most of the scene. In contrast, Peng yields a notably clean reconstruction. Both BU3D and our method suffer from outliers, but our method produces less noisy results.}

\section{Conclusion} \label{sec:6}

We have proposed a novel deep learning method for dual-peak reconstruction in single-photon Lidar imaging. 
Unlike existing deep learning-based methods, the proposed method supports the dual-peak reconstruction in a single run with uncertainty information. 
The experimental results show that the proposed method presents less artifacts such as bleeding effects with accurate results, although the proposed method often observes outliers in low photon cases. 
The proposed method can be extended to operate on unstructured or arbitrarily sampled spatial coordinates, such as measurements obtained from arbitrary directions.
Adapting the method to such cases would involve modifying the loss computation from grid-based to point-based such as Chamfer distance.
A limitation of the current method is to assume a fixed number of peaks per pixel, constrained to a dual-peak model. Future research directions include extending the framework from dual-peak to multi-peak reconstruction scenarios and incorporating reflectivity estimation capabilities. Another future work can be to develop adaptive mechanisms to automatically detect and handle varying numbers of targets per pixel.

%\clearpage

\bibliographystyle{IEEEtran}
\bibliography{IEEEabrv,punroll}

\begin{thebibliography}{10}
\providecommand{\url}[1]{#1}
\csname url@samestyle\endcsname
\providecommand{\newblock}{\relax}
\providecommand{\bibinfo}[2]{#2}
\providecommand{\BIBentrySTDinterwordspacing}{\spaceskip=0pt\relax}
\providecommand{\BIBentryALTinterwordstretchfactor}{4}
\providecommand{\BIBentryALTinterwordspacing}{\spaceskip=\fontdimen2\font plus
\BIBentryALTinterwordstretchfactor\fontdimen3\font minus \fontdimen4\font\relax}
\providecommand{\BIBforeignlanguage}[2]{{%
\expandafter\ifx\csname l@#1\endcsname\relax
\typeout{** WARNING: IEEEtran.bst: No hyphenation pattern has been}%
\typeout{** loaded for the language `#1'. Using the pattern for}%
\typeout{** the default language instead.}%
\else
\language=\csname l@#1\endcsname
\fi
#2}}
\providecommand{\BIBdecl}{\relax}
\BIBdecl

\bibitem{wallace2020full}
A.~Wallace, A.~Halimi, and G.~S. Buller, ``Full waveform {{LiDAR}} for adverse weather conditions,'' \emph{IEEE Trans. Veh. Technol.}, vol.~69, no.~7, pp. 7064--7077, 2020.

\bibitem{rapp2020advances}
J.~Rapp, J.~Tachella, Y.~Altmann, S.~McLaughlin, and V.~K. Goyal, ``Advances in single-photon lidar for autonomous vehicles: {{Working}} principles, challenges, and recent advances,'' \emph{IEEE Signal Process. Mag.}, vol.~37, no.~4, pp. 62--71, 2020.

\bibitem{buller2007ranging}
G.~Buller and A.~Wallace, ``Ranging and three-dimensional imaging using time-correlated single-photon counting and point-by-point acquisition,'' \emph{IEEE J. Sel. Top. Quantum Electron.}, vol.~13, no.~4, pp. 1006--1015, 2007.

\bibitem{kirmani2014first}
A.~Kirmani, D.~Venkatraman, D.~Shin, A.~Cola{\c c}o, F.~N. Wong, J.~H. Shapiro, and V.~K. Goyal, ``First-photon imaging,'' \emph{Science}, vol. 343, no. 6166, pp. 58--61, 2014.

\bibitem{shin2015photonefficient}
D.~Shin, A.~Kirmani, V.~K. Goyal, and J.~H. Shapiro, ``Photon-{{Efficient Computational}} 3-{{D}} and {{Reflectivity Imaging With Single-Photon Detectors}},'' \emph{IEEE Trans. Comput. Imaging}, vol.~1, no.~2, pp. 112--125, Jun. 2015.

\bibitem{halimi2016restoration}
A.~Halimi, Y.~Altmann, A.~McCarthy, X.~Ren, R.~Tobin, G.~S. Buller, and S.~McLaughlin, ``Restoration of intensity and depth images constructed using sparse single-photon data,'' in \emph{Eur. {{Signal Process}}. {{Conf}}. {{EUSIPCO}}}.\hskip 1em plus 0.5em minus 0.4em\relax Budapest, Hungary: IEEE, 2016, pp. 86--90.

\bibitem{chen2020learning}
S.~Chen, A.~Halimi, X.~Ren, A.~McCarthy, X.~Su, S.~McLaughlin, and G.~Buller, ``\BIBforeignlanguage{English}{Learning non-local spatial correlations to restore sparse 3d single-photon data},'' \emph{\BIBforeignlanguage{English}{IEEE Trans. Image Process.}}, vol.~29, pp. 3119--3131, 2020.

\bibitem{tachella2019bayesian}
J.~Tachella, Y.~Altmann, X.~Ren, A.~McCarthy, G.~S. Buller, S.~McLaughlin, and J.-Y. Tourneret, ``Bayesian {{3D Reconstruction}} of {{Complex Scenes}} from {{Single-Photon Lidar Data}},'' \emph{SIAM J. Imaging Sci.}, vol.~12, no.~1, pp. 521--550, Jan. 2019.

\bibitem{hernandez-marin2008multilayered}
S.~{Hernandez-Marin}, A.~M. Wallace, and G.~J. Gibson, ``Multilayered {{3D LiDAR Image Construction Using Spatial Models}} in a {{Bayesian Framework}},'' \emph{IEEE Trans. Pattern Anal. Mach. Intell.}, vol.~30, no.~6, pp. 1028--1040, Jun. 2008.

\bibitem{halimi2017object}
A.~Halimi, A.~Maccarone, A.~McCarthy, S.~McLaughlin, and G.~S. Buller, ``Object depth profile and reflectivity restoration from sparse single-photon data acquired in underwater environments,'' \emph{IEEE Trans. Comput. Imaging}, vol.~3, no.~3, pp. 472--484, 2017.

\bibitem{Pawlikowska_OE2017}
A.~M. Pawlikowska, A.~Halimi, R.~A. Lamb, and G.~S. Buller, ``Single-photon three-dimensional imaging at up to 10 kilometers range,'' \emph{Opt. Express}, vol.~25, no.~10, pp. 11\,919--11\,931, May 2017.

\bibitem{rapp2017few}
J.~Rapp and V.~K. Goyal, ``A {{Few Photons Among Many}}: {{Unmixing Signal}} and {{Noise}} for {{Photon-Efficient Active Imaging}},'' \emph{IEEE Trans. Comput. Imaging}, vol.~3, no.~3, pp. 445--459, Sep. 2017.

\bibitem{halimi2020robust}
A.~Halimi, R.~Tobin, A.~McCarthy, J.~{Bioucas-Dias}, S.~McLaughlin, and G.~S. Buller, ``Robust {{Restoration}} of {{Sparse Multidimensional Single-Photon LiDAR Images}},'' \emph{IEEE Trans. Comput. Imaging}, vol.~6, pp. 138--152, 2020.

\bibitem{Tobin_SR21}
R.~Tobin, A.~Halimi, A.~McCarthy, P.~Soan, and G.~Buller, ``Robust real-time 3d imaging of moving scenes through atmospheric obscurants using single-photon lidar,'' \emph{Sci. Rep.}, 2021.

\bibitem{altmann2018range}
Y.~Altmann and S.~McLaughlin, ``Range {{Estimation}} from {{Single-Photon Lidar Data Using}} a {{Stochastic Em Approach}},'' in \emph{Eur. {{Signal Process}}. {{Conf}}. {{EUSIPCO}}}, 2018.

\bibitem{legros2020expectationmaximization}
Q.~Legros, S.~Meignen, S.~McLaughlin, and Y.~Altmann, ``Expectation-{{Maximization Based Approach}} to {{3D Reconstruction From Single-Waveform Multispectral Lidar Data}},'' \emph{IEEE Trans. Comput. Imaging}, vol.~6, pp. 1033--1043, 2020.

\bibitem{venkatakrishnan2013plugandplay}
S.~V. Venkatakrishnan, C.~A. Bouman, and B.~Wohlberg, ``Plug-and-{{Play}} priors for model based reconstruction,'' in \emph{{{IEEE Glob}}. {{Conf}}. {{Signal Inf}}. {{Process}}.}\hskip 1em plus 0.5em minus 0.4em\relax IEEE, 2013.

\bibitem{tachella2019realtime}
J.~Tachella, Y.~Altmann, N.~Mellado, A.~McCarthy, R.~Tobin, G.~S. Buller, J.-Y. Tourneret, and S.~McLaughlin, ``Real-time {{3D}} reconstruction from single-photon lidar data using plug-and-play point cloud denoisers,'' \emph{Nat. Commun.}, vol.~10, no.~1, Dec. 2019.

\bibitem{ronneberger2015unet}
O.~Ronneberger, P.~Fischer, and T.~Brox, ``U-net: Convolutional networks for biomedical image segmentation,'' in \emph{Proc. Int. Conf. Med. Image Comput. Comput.-Assist. Intervent.}\hskip 1em plus 0.5em minus 0.4em\relax Cham, Switzerland: Springer International Publishing, 2015, pp. 234--241.

\bibitem{zhou2020unet++}
Z.~Zhou, M.~M.~R. Siddiquee, N.~Tajbakhsh, and J.~Liang, ``Unet++: Redesigning skip connections to exploit multiscale features in image segmentation,'' \emph{IEEE Trans. Med. Imag.}, vol.~39, no.~6, pp. 1856--1867, 2020.

\bibitem{Chen_2018_ECCV}
C.~Chen, Z.~Xiong, X.~Tian, and F.~Wu, ``Deep boosting for image denoising,'' in \emph{Proc. Eur. Conf. Comput. Vis. ECCV}, September 2018, pp. 3--18.

\bibitem{chen2020realworld}
C.~Chen, Z.~Xiong, X.~Tian, Z.-J. Zha, and F.~Wu, ``Real-{{World Image Denoising}} with {{Deep Boosting}},'' \emph{IEEE Trans. Pattern Anal. Mach. Intell.}, vol.~42, no.~12, pp. 3071--3087, Dec. 2020.

\bibitem{wang2018nonlocal}
X.~Wang, R.~Girshick, A.~Gupta, and K.~He, ``Non-local neural networks,'' in \emph{Proc. IEEE/CVF Conf. Comput. Vis. Pattern Recognit.}, 2018, pp. 7794--7803.

\bibitem{Yue2018CompactGN}
K.~Yue, M.~Sun, Y.~Yuan, F.~Zhou, E.~Ding, and F.~Xu, ``Compact generalized non-local network,'' in \emph{Neural Information Processing Systems}, 2018.

\bibitem{lindell2018singlephoton}
D.~B. Lindell, M.~O'Toole, and G.~Wetzstein, ``Single-photon {{3D}} imaging with deep sensor fusion,'' \emph{ACM Trans. Graph.}, vol.~37, no.~4, Aug. 2018.

\bibitem{sun2020spadnet}
Z.~Sun, D.~B. Lindell, O.~Solgaard, and G.~Wetzstein, ``{{SPADnet}}: {{Deep RGB-SPAD}} sensor fusion assisted by monocular depth estimation,'' \emph{Opt. Express}, vol.~28, no.~10, p. 14948, May 2020.

\bibitem{peng2020photonefficient}
J.~Peng, Z.~Xiong, X.~Huang, Z.-P. Li, D.~Liu, and F.~Xu, ``Photon-{{Efficient 3D Imaging}} with {{A Non-local Neural Network}},'' in \emph{Eur. {{Conf}}. {{Comput}}. {{Vis}}. {{ECCV}}}, 2020.

\bibitem{Zang:21}
Z.~Zang, D.~Xiao, and D.~D.-U. Li, ``Non-fusion time-resolved depth image reconstruction using a highly efficient neural network architecture,'' \emph{Opt. Express}, vol.~29, no.~13, pp. 19\,278--19\,291, Jun 2021.

\bibitem{Zhao:22}
X.~Zhao, X.~Jiang, A.~Han, T.~Mao, W.~He, and Q.~Chen, ``Photon-efficient 3d reconstruction employing a edge enhancement method,'' \emph{Opt. Express}, vol.~30, no.~2, pp. 1555--1569, Jan 2022.

\bibitem{peng2023penonlocal}
J.~Peng, Z.~Xiong, H.~Tan, X.~Huang, Z.-P. Li, and F.~Xu, ``Boosting photon-efficient image reconstruction with a unified deep neural network,'' \emph{IEEE Trans. Pattern Anal. Mach. Intell.}, vol.~45, no.~4, pp. 4180--4197, 2023.

\bibitem{koo2022bayesiana}
J.~Koo, A.~Halimi, and S.~Mclaughlin, ``A bayesian based deep unrolling algorithm for single-photon lidar systems,'' \emph{IEEE J. Sel. Top. Signal Process.}, 2022.

\bibitem{halimi2021robust}
A.~Halimi, A.~Maccarone, R.~A.~Lamb, G.~S.~Buller, and S.~McLaughlin, ``Robust and {{Guided Bayesian Reconstruction}} of {{Single-Photon 3D Lidar Data}}: {{Application}} to {{Multispectral}} and {{Underwater Imaging}},'' \emph{IEEE Trans. Comput. Imaging}, vol.~7, pp. 961--974, 2021.

\bibitem{gregor2010learning}
K.~Gregor and Y.~LeCun, ``Learning fast approximations of sparse coding,'' in \emph{Int. {{Conf}}. {{Mach}}. {{Learn}}. {{ICML}}}, 2010.

\bibitem{monga2021algorithm}
V.~Monga, Y.~Li, and Y.~C. Eldar, ``Algorithm {{Unrolling}}: {{Interpretable}}, {{Efficient Deep Learning}} for {{Signal}} and {{Image Processing}},'' \emph{IEEE Signal Process. Mag.}, vol.~38, no.~2, pp. 18--44, Mar. 2021.

\bibitem{yang2016deep}
y.~{yang}, J.~Sun, H.~Li, and Z.~Xu, ``Deep {{ADMM-Net}} for {{Compressive Sensing MRI}},'' in \emph{Adv. {{Neural Inf}}. {{Process}}. {{Syst}}.}, 2016.

\bibitem{zhang2020deep}
K.~Zhang, L.~Van~Gool, and R.~Timofte, ``Deep {{Unfolding Network}} for {{Image Super-Resolution}},'' in \emph{2020 {{IEEECVF Conf}}. {{Comput}}. {{Vis}}. {{Pattern Recognit}}. {{CVPR}}}.\hskip 1em plus 0.5em minus 0.4em\relax Seattle, WA, USA: IEEE, Jun. 2020, pp. 3214--3223.

\bibitem{koo2024bayesian}
J.~Koo, A.~Halimi, and S.~McLaughlin, ``Bayesian {{Deep Unfolding}} with {{Graph Attention}} for {{Dual-Peak Single-Photon Lidar Imaging}},'' in \emph{Eur. {{Signal Process}}. {{Conf}}. {{EUSIPCO}}}, 2024, pp. 646--650.

\bibitem{sheehan2021sketchinga}
M.~P. Sheehan, J.~Tachella, and M.~E. Davies, ``A {{Sketching Framework}} for {{Reduced Data Transfer}} in {{Photon Counting Lidar}},'' \emph{IEEE Trans. Comput. Imaging}, vol.~7, pp. 989--1004, 2021.

\bibitem{gutierrez-barragan2023learned}
F.~{Gutierrez-Barragan}, F.~Mu, A.~Ardelean, A.~Ingle, C.~Bruschini, E.~Charbon, Y.~Li, M.~Gupta, and A.~Velten, ``Learned {{Compressive Representations}} for {{Single-Photon 3D Imaging}},'' in \emph{Int. {{Conf}}. {{Comput}}. {{Vis}}. {{ICCV}}}, 2023.

\bibitem{parikh2014proximal}
N.~Parikh and S.~Boyd, ``Proximal algorithms,'' \emph{Foundations and Trends in optimization}, vol.~1, no.~3, pp. 127--239, 2014.

\bibitem{velickovic2018graph}
P.~Veli{\v c}kovi{\'c}, G.~Cucurull, A.~Casanova, A.~Romero, P.~Li{\`o}, and Y.~Bengio, ``Graph {{Attention Networks}},'' \emph{Int. Conf. Learn. Represent. ICLR}, 2018.

\bibitem{maddison2017concrete}
C.~J. Maddison, A.~Mnih, and Y.~W. Teh, ``The {{Concrete Distribution}}: {{A Continuous Relaxation}} of {{Discrete Random Variables}},'' in \emph{Int. {{Conf}}. {{Learn}}. {{Represent}}. {{ICLR}}}, Mar. 2017.

\bibitem{jang2017categorical}
E.~Jang, S.~Gu, and B.~Poole, ``Categorical {{Reparameterization}} with {{Gumbel-Softmax}},'' in \emph{Int. {{Conf}}. {{Learn}}. {{Represent}}. {{ICLR}}}, Aug. 2017.

\bibitem{butler2012naturalistic}
D.~J. Butler, J.~Wulff, G.~B. Stanley, and M.~J. Black, ``A naturalistic open source movie for optical flow evaluation,'' in \emph{Eur. {{Conf}}. {{Comput}}. {{Vis}}. {{ECCV}}}, 2012.

\bibitem{hirschmuller2007evaluation}
H.~Hirschmuller and D.~Scharstein, ``Evaluation of cost functions for stereo matching,'' in \emph{{{IEEE Conf}}. {{Comput}}. {{Vis}}. {{Pattern Recognit}}. {{CVPR}}}, 2007.

\bibitem{shin2016computational}
D.~Shin, F.~Xu, F.~N.~C. Wong, J.~H. Shapiro, and V.~K. Goyal, ``Computational multi-depth single-photon imaging,'' \emph{Opt. Express}, vol.~24, no.~3, p. 1873, Feb. 2016.

\end{thebibliography}

\section{Appendix} \label{sec:7}

\subsection{Comparison with Conference Version}
The network architecture has been improved over our previous conference version~\cite{koo2024bayesian} in two main ways.

In this paper, we improved the graph attention layer design in the expansion step. The conference paper~\cite{koo2024bayesian} concatenated the $d_{feat}$ and $|d-x|_{feat}$ features and applied GAT operations. In the proposed network, we apply GAT operations to each feature separately and compute attention weights. This allows us to compute attention weights for each feature independently, which is more flexible and effective. 
% On the next page, we provide two figures to see the difference between the conference paper and the proposed network.

Second, we make the hard attention mechanism explicit. In the conference paper, hard attention was implicitly applied twice during the squeeze step, where the smaller of the two estimated depths was taken as the first depth and the larger as the second. In the proposed network, we explicitly estimate the first and second depths by applying hard attention to the 4 scales, respectively. This explicit separation improves the robustness of the network’s depth estimation.
The benefit of having the new observation models is the interpretability through uncertainty information. The proposed algorithm in this paper has the ability to estimate uncertainty, a key benefit compared to the conference paper. Moreover, the observation model is not limited to the proposed algorithm and the model can lead to different algorithms in future work.

% -------

To demonstrate the improvements over our previous conference version~\cite{koo2024bayesian}, we provide quantitative comparisons on two scenes: \textit{Art} and \textit{Bowling}.
% o highlight the improvements made since the previous conference version of our work, we present a quantitative comparison on two representative scenes: \textit{Art} and \textit{Bowling}. 
Tables~\ref{tab:art_conference} and~\ref{tab:bowling_conference} report the DAE (Depth Absolute Error) under various signal-to-background ratios (SBRs), with photons-per-pixel (PPP) fixed at 4.

Our proposed model consistently outperforms the conference version across all tested SBR levels. 
Notably, even under low-SBR conditions (SBR = 1), our method demonstrates improved robustness and accuracy.
% , highlighting the benefits of architectural revisions and enhanced uncertainty modeling. 
These results empirically validate the effectiveness of the proposed changes over the earlier version.

% ------------------------------
% Table Art with conference 
\begin{table}[H]
    \centering
    \caption{Quantitative comparison with conference model on the Art scene.}
    \resizebox{\columnwidth}{!}{%
    % \resizebox{\linewidth}{!}{%
    \begin{tabular}{l c c c c c}
    \hline
    \noalign{\vspace{2pt}}
     & SBR = 1 & & SBR = 4 & & SBR = 16 \\[2pt]
    %\cline{2-3} \cline{4-5} \cline{6-7} 
    \noalign{\vspace{2pt}}
     %& DAE & & DAE & & DAE \\[2pt]
    \hline
    
    \noalign{\vspace{2pt}}
    \multicolumn{6}{l}{PPP = 4} \\

    Conference & $0.0069\pm0.0346$  & & $0.0058\pm0.0282$ & & $0.0055\pm0.0267$\\
    Proposed & $\textbf{0.0058}\pm0.0313$ & & $\textbf{0.0047}\pm0.0245$ & & $\textbf{0.0045}\pm0.0228$ \\[2pt]
    
    \hline

    \end{tabular}%
    }
    \label{tab:art_conference}
\end{table}
%
% ------------------------------

% % table Bowling1 with conference
\begin{table}[H]
    \centering
    \caption{Quantitative comparison with conference model on the Bowling scene.}
    \resizebox{\columnwidth}{!}{%
    \begin{tabular}{l c c c c c}
    \hline
    \noalign{\vspace{2pt}}
        & SBR = 1 & & SBR = 4 & & SBR = 16 \\[2pt]
    \noalign{\vspace{2pt}}
        & DAE & & DAE & & DAE \\[2pt]
    \hline
    \noalign{\vspace{2pt}}

    \multicolumn{6}{l}{PPP = 4} \\

    Conference & $0.0041\pm0.0301$ & & $0.0032\pm0.0236$ & & $0.0030\pm0.0237$  \\
    Proposed & $\textbf{0.0035}\pm0.0252$ & & $\textbf{0.0028}\pm0.0199$ & & $\textbf{0.0027}\pm0.0206$ \\ [2pt]
    
    \hline

    \end{tabular}%
    }
    \label{tab:bowling_conference}
\end{table}

%%%%%%%%%%%%%%%%%%%%%%%%%%%%%%%%%%%%%%%%%%%%%%%%%%%%%%%%%%%%%

\end{document}